\documentclass[preprint,12pt,authoryear]{elsarticle}

\usepackage{amsmath,amssymb,bm}
\usepackage{graphicx,booktabs,multirow}
\usepackage{geometry}
\usepackage[authoryear]{natbib}
\usepackage{hyperref}
\usepackage{tikz}
\usepackage{comment}
\usepackage{subfig}
\usepackage{float}
\usepackage{xcolor}
\usepackage[ruled,vlined,linesnumbered]{algorithm2e}
\usepackage{wrapfig}
\usepackage{array}

\usetikzlibrary{matrix,positioning,arrows.meta,calc,backgrounds}

\DontPrintSemicolon
\SetKwInput{KwIn}{Input}
\SetKwInput{KwOut}{Output}

\begin{document}

\begin{frontmatter}
\title{Beyond Accuracy: Reliability and Uncertainty Estimation in Convolutional Neural Networks}

\author[1]{Sanne Ruijs}
\ead{Sanneruys2@gmail.com}

\author[1]{Alina Kosiakova}
\ead{alina.kosiakova@student.lu.se}

\author[2]{Farrukh Javed\corref{cor1}}
\ead{farrukh.javed@stat.lu.se}

\address[1]{Department of Economics, Lund University, Sweden}
\address[2]{Department of Statistics, Lund University, Sweden}

\cortext[cor1]{Corresponding author}


\begin{abstract}




Deep neural networks (DNNs) have become integral to a wide range of scientific and practical applications due to their flexibility and strong predictive performance. Despite their accuracy, however, DNNs frequently exhibit poor calibration, often assigning overly confident probabilities to incorrect predictions. This limitation underscores the growing need for integrated mechanisms that provide reliable uncertainty estimation. In this article, we compare two prominent approaches for uncertainty quantification: a Bayesian approximation via Monte Carlo Dropout and the nonparametric Conformal Prediction framework. Both methods are assessed using two convolutional neural network architectures; H-CNN VGG16 and GoogLeNet, trained on the Fashion-MNIST dataset. The empirical results show that although H-CNN VGG16 attains higher predictive accuracy, it tends to exhibit pronounced overconfidence, whereas GoogLeNet yields better-calibrated uncertainty estimates. Conformal Prediction additionally demonstrates consistent validity by producing statistically guaranteed prediction sets, highlighting its practical value in high-stakes decision-making contexts.
Overall, the findings emphasize the importance of evaluating model performance beyond accuracy alone and contribute to the development of more reliable and trustworthy deep learning systems.
\end{abstract}




\end{frontmatter}



\section{Introduction}
\label{sec:introduction}
Deep neural networks (DNNs) have become a cornerstone of modern machine learning, owing to their ability to model complex data structures and their broad applicability across scientific domains such as medical imaging, robotics, and earth observation \citep{Gawlikowski2023}. Despite their impressive performance, DNNs are often regarded as ``black boxes'' due to their limited interpretability and the difficulty of aligning their internal representations with human reasoning \citep{Roth2024}. This lack of transparency is particularly concerning in safety-critical applications. Furthermore, research has shown that neural networks frequently exhibit overconfidence, assigning high-probability predictions even when they are incorrect \citep{Nguyen2015}. A key limitation of conventional DNNs is their reliance on point estimates without providing any quantification of uncertainty. As a result, estimating predictive uncertainty becomes essential for assessing the reliability and robustness of model outputs, particularly in high-risk or decision-sensitive scenarios.

To address the lack of built-in uncertainty estimation, several methods have been proposed. One widely used technique is Bayesian inference with Monte Carlo (MC) Dropout, which approximates a posterior distribution while maintaining computational efficiency (see, for example, \citep{Gal2016} and \citep{SON2026131927}). Alternatively, Conformal Prediction offers statistically valid prediction sets without requiring assumptions about the underlying data distribution \citep{angelopoulos2022gentleintroductionconformalprediction}.These approaches represent two fundamentally different paradigms: Bayesian methods are probabilistic and integrated during model training and inference, whereas Conformal Prediction is a post-hoc method that can be applied to any pre-trained model. These two methods have been selected due to their inherent differences, which offer valuable comparisons, as well as their growing popularity in uncertainty estimation.

As deep learning research grows, uncertainty quantification (UQ) has become increasingly relevant. Between 2010 and 2020, over 2,500 papers were published on UQ in various fields (see, for example, \citep{Abdar2021} and \citep{FERCHICHI2025130242} for review). Yet, few studies offer a systematic comparison of Bayesian and Conformal approaches, particularly across diverse neural network architectures. Additionally, the relationship between accuracy and uncertainty remains ambiguous \citep{Roth2024}. High classification accuracy does not necessarily imply trustworthy predictions, as models often remain overconfident even when incorrect. Therefore, UQ should play a more central role in model evaluation.

The CNN architectures were chosen based on their strong empirical performance in prior studies and complementary design characteristics. H-CNN VGG16 was selected for its demonstrated effectiveness on the Fashion-MNIST dataset, particularly in distinguishing between visually similar clothing categories. Its hierarchical architecture helps reduce misclassification of ambiguous classes and enhances model interpretability. GoogLeNet, by contrast, adopts an inception-based design that processes features through parallel convolutional paths. This architecture achieves high accuracy while being more parameter-efficient than H-CNN VGG16, making it a computationally attractive alternative without compromising performance.

This study aims to fill this gap by conducting a comparative analysis of two uncertainty estimation methods: Bayesian inference via MC Dropout and Conformal Prediction across two convolutional neural network architectures: H-CNN VGG16 and GoogLeNet. Beyond their overall performance, the analysis investigates the behavior of predictive uncertainty at multiple levels, including the decomposition of uncertainty and its manifestation in ambiguous class predictions. The study contributes to a more comprehensive understanding of model reliability and advances the interpretability and trustworthiness of deep learning models. The Fashion-MNIST dataset is used throughout, offering a standardized benchmark in image classification for a precise comparison of the selected methods across model architectures without unexplained effects from data inconsistency.

\subsection*{Main Contributions}

\begin{itemize}
    \item Comparative study of Bayesian MC Dropout and Conformal Prediction in neural networks
    \item Uses uncertainty to expose class ambiguity to enhance decision making
    \item Shows MC Dropout limits in overfitting-prone deep hierarchical architectures
    \item Demonstrates complementary strengths of Bayesian MC Dropout and Conformal Prediction
    \item Highlights potential of uncertainty estimation with minimal computational cost

\end{itemize}

In summary, this article contributes to a more comprehensive understanding of predictive reliability in deep learning.  By examining both Bayesian and Conformal frameworks across distinct architectures (VGG16 and GoogLeNet), the study advances the development of models that are not only accurate but also transparent and trustworthy.

\section{Literature Review}
\label{sec1}

This section reviews the importance of uncertainty quantification (UQ) in deep learning and examines two leading methods: Bayesian approximation via Monte Carlo Dropout (MC Dropout) and Conformal Prediction (CP). It also explores prior research on the Fashion-MNIST dataset as a benchmarking tool for neural network models and describes the rationale behind selecting the CNN architectures used in the article.


\subsection{Uncertainty Estimation in Deep Neural Networks}
\label{sec:uncertainty}
Deep learning models, particularly CNNs, have achieved significant success in image classification tasks, including high-risk domains such as medical diagnostics and autonomous systems \citep{angelopoulos2022gentleintroductionconformalprediction}. Despite their high predictive accuracy, these models are often overconfident, producing unreliable probability estimates \citep{Guo2017,Gawlikowski2023}. This miscalibration undermines trust in critical applications, where understanding a model’s uncertainty is essential.  As noted by \citet{Poceviit2023}, deep neural networks typically rely on the softmax output to estimate “confidence,” which reflects the conditional probability assigned to each class. However, as \citet{Guo2017} demonstrate, these softmax-derived probabilities are often poorly calibrated and do not match the true likelihood of correctness. As a result, relying on softmax outputs as a measure of model confidence can be misleading.

Thus, there is a growing demand for methods that estimate predictive uncertainty in a statistically sound and interpretable manner. Uncertainty quantification serves several essential roles:  First, it allows practitioners to defer uncertain cases to human experts \citep{Papadopoulos08}, enhances interpretability by highlighting ambiguous samples \citep{lu2022fairconformalpredictorsapplications}, and supports robust decision-making in high-risk contexts.  Second, it offers additional insight into the model’s behaviour and aids in the interpretation of deep learning methods; for instance, analysing the size and composition of prediction sets can help identify ambiguous inputs or systematically difficult classes \citep{lu2022fairconformalpredictorsapplications}. These capabilities make uncertainty estimation an essential component in deploying machine learning models responsibly in high-risk environments. 

 \subsection{Uncertainty Estimation Methods}
\label{sec:uncertainty1}

In this section, we provide a review of two uncertainty estimation methods and describe previous research regarding their applicability to neural networks.

\subsubsection{Conformal Predictions}
\label{sec:uncertainty3}
CP is a model-agnostic technique that provides prediction sets with a guaranteed error rate under minimal assumptions \citep{Shafer2008,Fontana2023}. It only requires data exchangeability and can be applied post hoc to any point prediction model. This includes classification and regression methods, such as support vector machines, decision trees and neural networks \citep{Shafer2008}. This versatility directly allows the method to be easily implemented on large datasets and deep, complex models without altering the structure of the underlying architectures. CP has also been integrated with various CNNs, including VGG16 and ResNet, and successfully used in applications ranging from facial recognition \citep{Matiz2019} to skin lesion diagnosis \citep{lu2022fairconformalpredictorsapplications}. Its simplicity and statistical guarantees make it an appealing option for uncertainty quantification, especially when Bayesian priors are difficult to specify. 

\subsubsection{Bayesian Approximation Using Monte Carlo Dropout}
\label{sec:uncertainty2}

Bayesian methods can effectively capture two types of uncertainties in data modelling, namely the aleatoric and epistemic uncertainties \citep{kendall_2019}. The first, aleatoric uncertainty, captures the natural noise present in observations, such as measurement variability, that remains constant regardless of how much data we collect. The second, epistemic uncertainty, represents our incomplete understanding of the model itself. Unlike aleatoric uncertainty, this model uncertainty diminishes as we gather more training data, reflecting our growing confidence in the learned parameters. While the former is irreducible, the latter can be reduced by improving the model learned by the neural network and introducing more data \citep{Gawlikowski2023,Essbai2024}.

While conceptually straightforward, neural networks are often non-linear and high-dimensional, making the process of inference computationally infeasible and the resulting posterior distribution intractable \citep{kendall_2019,sun2022conformalmethodsquantifyinguncertainty}. MC Dropout, introduced by \citet{Gal2016}, addressed this limitation and approximates Bayesian inference by applying dropout at inference time, thereby sampling from an approximate posterior distribution.  Instead of learning fixed weights, the method samples from an approximated posterior distribution during inference, introducing model uncertainty. \citet{Gal2016} showed that the model is effectively regularised by averaging multiple sampled weight configurations to reduce variance and discouraging over-dependence on specific parameters. The training process remains unchanged, but becomes scalable compared to other Bayesian inference methods.

MC Dropout have been used in various domains confirming its scalability and robustness. For example, in  medical imaging, \citet{Eaton-Rosen2018} performed uncertainty quantification of brain tumour image segmentation on the ResNet architecture. \citet{Gal2017} demonstrated the effectiveness of MC Dropout by evaluating it on both MNIST and dermoscopic lesion image datasets using the VGG16 CNN architecture. Their approach incorporated MC Dropout to approximate predictive distributions and quantify predictive uncertainty during inference. Similarly, in soil analysis, \citet{Padarian2022} applied a 2D convolutional neural network with MC Dropout to soil spectral data for predicting it's properties while assessing prediction uncertainty. Their research specifically focused on evaluating how uncertainty quantification methods perform when testing data differs significantly from training distributions. The results revealed that MC Dropout provides substantially wider and more reliable prediction intervals for out-of-domain data compared to alternative methods such as bootstrapping. These findings demonstrate the broad applicability and robustness of MC Dropout, establishing it as a suitable choice for Bayesian uncertainty quantification in neural network applications.

\subsubsection{Comparison of Bayesian Inference and Conformal Predictions}
\label{sec:uncertainty4}
The two methods, Bayesian inference and Conformal Prediction, adopt fundamentally different strategies for uncertainty estimation. Bayesian inference is a probabilistic framework that relies on prior assumptions about the data distribution and model parameters. While Monte Carlo (MC) Dropout offers an efficient approximation of the posterior distribution \citep{KendallGal2017}, achieving proper calibration of this distribution remains a challenge. In contrast, Conformal Predictions are nonparametric and distribution-free, providing finite-sample guarantees on coverage without requiring strong model assumptions \citep{sun2022conformalmethodsquantifyinguncertainty}. One key advantage of Conformal Prediction is its flexibility: it can be applied post hoc to any trained model, including deep neural networks, without modifying the underlying architecture. Moreover, it is computationally more efficient and easier to implement than Bayesian methods \citep{angelopoulos2022gentleintroductionconformalprediction}. However, its primary limitation lies in being overly conservative, often resulting in unnecessarily wide prediction sets or intervals \citep{fan2024utopiauniversallytrainableoptimal}.

Despite the widespread application of both methods across deep learning and convolutional neural network (CNN) architectures, few studies have performed direct comparisons of their performance on shared benchmarks. Furthermore, little attention has been paid to how architectural design affects the behavior of uncertainty estimation techniques. For instance, \citet{fontes2023bayesian} examined uncertainty quantification in binary classification using logistic regression models on small-scale datasets. Their evaluation, based on F1 scores at different uncertainty thresholds (top 25\%, 50\%, and 75\% most confident predictions), revealed that while Conformal Predictions required reducing the training set size, they provided valid prediction sets. Meanwhile, the Bayesian approach offered greater flexibility by capturing both epistemic and aleatoric uncertainty, albeit at the cost of higher computational complexity.

Similarly, \citet{liang2024conformalpredictionquantifyinguncertainty} compared MC Dropout, CP, and ensemble methods for quantifying uncertainty in neural operators, specialized networks used for solving complex partial differential equations. Their results demonstrated that Conformal Prediction yielded more reliable confidence intervals with theoretical coverage guarantees. In another study, \citet{Khurjekar2023} evaluated the statistical validity of uncertainty intervals produced by Conformal Prediction and MC Dropout in the context of direction-of-arrival estimation. While Conformal Prediction consistently met the required coverage levels, MC Dropout failed to do so, underscoring its limitations in providing calibrated uncertainty estimates. Furthermore, \citet{shiman_li} investigated pixel-level, sample-level and overall uncertainty evaluation for medical image segmentation. One of the methods employed was MC Dropout, demonstrating a good balance between reliability and accuracy, which is easy to implement.

Overall, these findings highlight the complementary strengths and weaknesses of both approaches and emphasize the need for systematic evaluations, particularly in relation to neural network architecture and application domain. This article addresses this gap in the current literature by evaluating both uncertainty estimation methods on a common benchmark. The review revealed no prior systematic comparison of the two techniques, nor an assessment of how CNN architectural differences influence uncertainty quantification outcomes. The novelty of this article lies in its comparative perspective and the interpretation of how model design affects uncertainty behavior in deep neural networks.

\subsection{Fashion-MNIST and Prior Research}
\label{sec:dataset}

This section outlines the rationale for selecting Fashion-MNIST as the benchmarking dataset for this study and highlights its widespread use in deep learning research. It also reviews prior work that applied the dataset in image classification tasks, particularly with convolutional neural networks (CNNs).

\subsubsection{Benchmarking Fashion-MNIST with Deep Learning Models}
\label{sec:dataset1}

Fashion-MNIST is a publicly available dataset introduced by \citet{xiao2017fashionmnistnovelimagedataset} as a more challenging alternative to the classic MNIST digit dataset. Designed for benchmarking machine learning algorithms, it has become a standard case study for image classification tasks. Upon release, Fashion-MNIST was benchmarked using several traditional classifiers, such as Gradient Boosting (88.0\% accuracy), K-Nearest Neighbours (85.4\%), and Support Vector Classifiers (89.7\%). Since then, it has been widely adopted in the deep learning community, especially for evaluating convolutional neural networks, which leverage layered architectures to learn spatial hierarchies from edges to complex textures \citep{bbouzidi2024convolutionalneuralnetworksvision}.

Numerous CNN architectures have been applied to Fashion-MNIST with consistently high performance. LeNet, one of the earliest CNN models, consists of seven layers and achieved 90.14\% accuracy on this dataset \citep{Vives-Boix2021}. AlexNet, with its deeper architecture and use of larger filters and non-linear activations \citep{Krizhevsky2012}, improved performance to 91.19\% \citep{Vives-Boix2021}. VGG-type architectures introduced by \citet{Simonyan2015}, notably VGG16 and VGG19, with 16 and 19 layers respectively, enabled extraction of more complex features and reached accuracies of 92.89\% and 92.90\% on Fashion-MNIST \citep{Seo2019}. These results were further supported by \citet{Vives-Boix2021}, who reported 92.45\% accuracy using VGG16. Due to this strong performance, VGG-type models are now commonly used as baselines for evaluating newer CNN architectures on Fashion-MNIST.

Over time, CNN research has shifted toward increasingly deeper and more complex architectures \citep{Krichen2023}. As noted by \citet{Alzubaidi2021}, shallow networks often struggle to capture hierarchical patterns in high-dimensional data, limiting their effectiveness. This trend has motivated a growing focus on balancing computational cost with model accuracy, which is a trade-off that remains central to architecture selection in modern CNN research.

\subsubsection{Rationale for Architecture Selection}
\label{sec:dataset2}

This study employs two high-performing convolutional architectures: H-CNN VGG16 \citep{Seo2019} and GoogLeNet \citep{Szegedy2015,Samia2022,Vives-Boix2021}, chosen for their strong empirical results and complementary design characteristics.

 \citet{Seo2019} proposed a Hierarchical Convolutional Neural Network (H-CNN) to address the challenge of misclassifying visually similar apparel categories such as shirts, T-shirts, and coats. The model leverages a hierarchical structure, first classifying broad categories (e.g., tops and bottoms) and subsequently refining predictions to more specific labels. To validate the approach, the H-CNN structure was applied to VGG16 and VGG19, resulting in a notable accuracy gain. Specifically, the VGG16-based H-CNN achieved 93.52\% accuracy, approximately 1\% higher than standard VGG16. The architecture also incorporated dropout regularization and loss weight scheduling, encouraging the model to focus gradually from general to finer-grained categories. This helped mitigate early overfitting and improved the model’s ability to resolve class ambiguities.

GoogLeNet (also known as Inception V1), introduced by \citet{Szegedy2015}, is a prominent deep learning architecture that differs from traditional sequential designs by using parallel convolutional paths. Instead of stacking layers linearly, GoogLeNet employs Inception modules, which apply multiple convolutional filters of varying sizes within the same layer.  The outputs of these parallel operations are then concatenated to form the module’s final output \citep{bbouzidi2024convolutionalneuralnetworksvision}. This design improves computational efficiency and helps mitigate issues such as vanishing gradients, contributing to more stable training \citep{Janjua2023}. GoogLeNet also uses dropout prior to the final fully connected layer to reduce overfitting \citep{Szegedy2015}. When applied to Fashion-MNIST, GoogLeNet achieved high accuracy; 93.75\% according to \citet{Samia2022}, and 91.89\% in \citet{Seo2019}. Despite being shallower than VGG-type networks, GoogLeNet offers comparable accuracy with significantly fewer parameters, making it a computationally efficient alternative.

The contrasting design philosophies of GoogLeNet and H-CNN VGG16 allow for an insightful investigation into how architecture influences the performance and behavior of uncertainty quantification methods. While CNNs have traditionally emphasized accuracy, their tendency toward poor calibration and overconfidence remains a key challenge \citep{Guo2017}. These limitations highlight the need for robust uncertainty estimation strategies. This study addresses a notable gap in the literature by conducting a side-by-side evaluation of uncertainty quantification methods across structurally distinct CNN architectures using a standardized benchmark dataset. Unlike most previous research that centers on classification accuracy, this work places greater emphasis on uncertainty calibration and interpretability, thereby contributing to the advancement of more transparent and trustworthy deep learning systems.

\section{Dataset}
\label{sec:data3}
The dataset used in this study is Fashion-MNIST, a publicly available benchmark introduced by \citet{xiao2017fashionmnistnovelimagedataset}. It comprises 70{,}000 grayscale images of fashion products sourced from Zalando’s online catalog, encompassing a diverse selection of men’s, women’s, kids’, and unisex clothing. Each image, sized at 28$\times$28 pixels, depicts a single item and is annotated with one of ten predefined class labels: \textit{T-shirt/top}, \textit{Trouser}, \textit{Pullover}, \textit{Dress}, \textit{Coat}, \textit{Sandal}, \textit{Shirt}, \textit{Sneaker}, \textit{Bag}, and \textit{Ankle boot}. These labels correspond to Zalando’s silhouette codes and were manually verified to ensure annotation accuracy and consistency.

To maintain compatibility with MNIST-based models, the dataset underwent standardized preprocessing, including whitespace trimming, aspect-ratio-preserving resizing, Gaussian sharpening, grayscale conversion, intensity inversion, and centering based on the object’s center of mass. Images with low contrast or unsuitable backgrounds were excluded. The final dataset is divided into a training set of 60{,}000 images and a test set of 10{,}000 images, and it retains the same format and file structure as MNIST, facilitating straightforward adoption in existing machine learning workflows.

\section{Methodology}
\label{sec:empirical1}

This section outlines the methodological framework used in the study, covering the selected  CNN architectures, uncertainty estimation techniques, and evaluation metrics.

\subsection{Model Architectures}
\label{sec:models}
The study evaluates two convolutional architectures, H-CNN VGG16 and GoogLeNet, selected for their strong empirical performance on Fashion-MNIST and their complementary structural designs.

The H-CNN VGG16 architecture, introduced by \citet{Seo2019}, builds on the VGG16 model by incorporating a hierarchical classification strategy to address confusion among visually similar classes. The model applies dropout regularization and loss-weight scheduling to shift learning gradually from general to fine-grained categories, helping reduce overfitting and improve interpretability.
GoogLeNet (Inception v1), proposed by \citet{Szegedy2015}, utilizes Inception modules to process input through multiple convolutional paths in parallel. This architecture efficiently captures both local and global features while maintaining a lower parameter count compared to VGG-based models. To ensure a fair comparison, training hyperparameters such as batch size, learning rate, and number of epochs were kept consistent across both models. All experiments were conducted on a workstation equipped with an NVIDIA RTX 3080 GPU, 32 GB RAM, and an Intel\textsuperscript{\textregistered} Core\texttrademark~i9-11900KF processor. Training times reported in Section~\ref{sec:empirical2} reflect this setup.

\subsection{Uncertainty Estimation Methods}
\label{sec:uq_methods}

Two uncertainty estimation frameworks are evaluated: Inductive Conformal Prediction (ICP) and Monte Carlo (MC) Dropout. This section focuses on the theoretical foundations and implementation of ICP.

\subsubsection{Conformal Predictions}
\label{sec:CP}

Conformal Prediction (CP) is a distribution-free framework that quantifies uncertainty in machine learning models by producing prediction sets instead of single-point estimates. These sets offer finite-sample coverage guarantees and do not rely on the data distribution, making CP particularly attractive for high-dimensional tasks such as image classification \citep{angelopoulos2022gentleintroductionconformalprediction}. CP is based on the assumption of exchangeability, a weaker condition than i.i.d., which ensures that the joint probability of the data remains invariant under permutations. This allows CP to maintain its validity without additional assumptions on the model or data-generating process \citep{Shafer2008, zhou2025conformalpredictiondataperspective}.

The primary goal of CP is to ensure that, with a user-defined significance level $\alpha$, the prediction set $\mathcal{C}(x_{\text{new}})$ for a new input $x_{\text{new}}$ contains the true label $y_{\text{new}}$ with probability at least \(1 - \alpha\):
\begin{equation}
    P\left(y_{\text{new}} \in \mathcal{C}(x_{\text{new}})\right) \geq 1 - \alpha.
    \label{eq:conformal_validity}
\end{equation}

To implement CP, the data is divided into training, calibration, and test sets. The calibration set is used to compute nonconformity scores, which assess how atypical a prediction is. While a larger calibration set improves the precision of prediction sets, it may reduce model performance by shrinking the training set \citep{barber2023conformal}. To balance this trade-off, this study uses 2,000 samples (approximately 2.86\% of the dataset) for calibration, preserving sufficient training data while maintaining reliable coverage, consistent with the suggestions of \citet{angelopoulos2022gentleintroductionconformalprediction}.

In classification tasks, the nonconformity score is often defined as:
\begin{equation}
    s_i = 1 - \hat{f}(X_i)_{y_i},
    \label{eq:nonconformity_score}
\end{equation}
where $\hat{f}(X_i)_{y_i}$ is the softmax probability assigned to the true class label. Higher scores indicate lower confidence in the true class, and thus higher uncertainty. Once nonconformity scores are calculated for the calibration set, a quantile threshold $\hat{q}$ is determined:
\begin{equation}
    \hat{q} = \text{the } \left\lceil \frac{(n+1)(1 - \alpha)}{n} \right\rceil\text{-th smallest score},
    \label{eq:conformal_quantile}
\end{equation}
where $n$ is the size of the calibration set. The prediction set for a new input includes all labels for which the predicted nonconformity score is less than or equal to $\hat{q}$.

CP aims to balance two properties: \textit{validity}, or the statistical guarantee that the true label lies within the prediction set, and \textit{efficiency}, which reflects the set’s compactness and informativeness. While CP guarantees validity under the exchangeability assumption, efficiency is not assured and depends on the quality of the underlying model and nonconformity function \citep{Shafer2008}. This is particularly relevant in multiclass problems, where inefficient prediction sets may include several irrelevant classes, reducing interpretability.

CP is categorized into two main types: Transductive Conformal Prediction (TCP) and Inductive Conformal Prediction (ICP). TCP assigns each possible label to the new input, computes the corresponding nonconformity score, and compares it against calibration scores to form the prediction set. While accurate, this process is computationally intensive. In contrast, ICP trains a single model on the training set and applies it to both calibration and test data. The calibration set is used to compute a quantile threshold, and predictions for new inputs are made using a fixed rule derived from the training phase \citep{Papadopoulos08, Fontana2023}.

This study adopts Inductive Conformal Prediction (ICP) for its computational efficiency and scalability, particularly suited for large-scale, high-dimensional datasets like Fashion-MNIST. ICP is also model-agnostic and can be applied post hoc to any pre-trained CNN, providing flexibility in practical deployment while offering valid and interpretable uncertainty estimates.

\subsubsection{Bayesian Approximation using MC Dropout}
\label{sec:Bay}

Bayesian inference provides a probabilistic framework for modelling uncertainty by estimating distributions over parameters rather than relying on fixed point estimates \citep{Lindholm2021}. It incorporates prior beliefs about model parameters, which are updated using observed data to compute the posterior distribution. This process is governed by Bayes' theorem:

\begin{equation}
P(\theta \mid y) = \frac{P(\theta) \cdot P(y \mid \theta)}{P(y)},
\label{eq:bayes_theorem}
\end{equation}

where \( P(\theta \mid y) \) denotes the posterior distribution, \( P(\theta) \) the prior, \( P(y \mid \theta) \) the likelihood, and \( P(y) \) the marginal likelihood (evidence). While powerful, the Bayesian approach faces practical limitations in deep learning contexts. The reliance on prior distributions introduces subjectivity, and the inference procedure can become intractable in high-dimensional parameter spaces \citep{Jospin2022, 9745083}. 

Bayesian neural networks (BNNs) generalize conventional deep learning models by placing distributions over the weights and biases, thereby enabling the quantification of epistemic uncertainty—uncertainty arising from limited data or model structure \citep{Chandra2021, Essbai2024}. Unlike traditional deterministic networks that risk overfitting by memorizing training data, BNNs produce probabilistic outputs that more accurately reflect model confidence. However, full Bayesian inference in deep neural networks is computationally expensive, requiring either sampling-based methods or variational approximations, both of which can be prohibitive for large-scale models.

To overcome these challenges, this study adopts Monte Carlo (MC) Dropout, a scalable and efficient Bayesian approximation technique introduced by \citet{Gal2016}. Originally proposed as a regularization method, dropout involves randomly deactivating units during training to prevent overfitting. MC Dropout extends this mechanism to the inference phase by keeping dropout active during testing. Each stochastic forward pass with dropout yields a different output for the same input, approximating samples from the model’s predictive posterior.

By performing \( T \) stochastic forward passes for each input \( \mathbf{x}^* \), MC Dropout estimates the predictive distribution as follows:

\begin{equation}
    p(y^* \mid \mathbf{x}^*, \mathcal{D}) \approx \frac{1}{T} \sum_{t=1}^{T} p\left(y^* \mid \mathbf{x}^*, \widehat{\mathbf{W}}_t\right),
    \label{eq:mc_dropout_predictive}
\end{equation}

where \( \widehat{\mathbf{W}}_t \) represents the randomly sampled network weights at iteration \( t \), and \( p(y^* \mid \mathbf{x}^*, \widehat{\mathbf{W}}_t) \) is the softmax output for that pass. The mean of these outputs provides the final prediction, while their variance serves as an estimate of epistemic uncertainty.

MC Dropout thus enables uncertainty-aware prediction without modifying the training objective or architecture, making it especially useful for convolutional neural networks applied to complex image datasets such as Fashion-MNIST. The technique is lightweight and well-suited for high-dimensional tasks, offering a balance between computational efficiency and the interpretability of Bayesian methods. Prior studies suggest that 50 forward passes yield a reliable trade-off between uncertainty estimation quality and inference cost \citep{Gal2016, Abdar2021}.

Importantly, MC Dropout has also been interpreted as a form of variational inference, approximating the posterior distribution over weights without requiring explicit sampling or expensive reparameterization strategies \citep{shridhar2019bcnn, Gal2017}. This makes it a practical tool for incorporating uncertainty into deep learning pipelines, especially where full Bayesian implementations are infeasible. Overall, MC Dropout provides an accessible and effective mechanism for estimating model uncertainty in deep learning, enhancing both reliability and interpretability of neural predictions. 

\subsection{Evaluation Metrics}
\label{sec:metrics1}

This section outlines the evaluation metrics used to assess the performance of the proposed uncertainty quantification methods. It covers both general model evaluation criteria and metrics specific to Conformal Prediction (CP) and Bayesian approximation using Monte Carlo (MC) Dropout.

\subsubsection{General Evaluation Metrics}
\label{sec:metrics2}

\vspace{1em}
{\normalfont\itshape Sparsity\par}
\vspace{0.3em}
Sparsity is typically analyzed in the context of model compression and pruning techniques aimed at reducing network complexity without compromising predictive performance. One such method is Global Magnitude Pruning (Global MP), a widely used and computationally efficient approach that removes weights with magnitudes below a predefined threshold. This threshold, denoted by \( t \), is computed based on a target sparsity level \( \kappa_\text{target} \) and serves as a universal cut-off across all network layers, in contrast to layer-wise or uniform pruning strategies that require separate thresholds per layer \citep{gupta2024complexityrequiredneuralnetwork}.

Formally, the pruning rule is expressed as follows:

\begin{equation}
f(w) =
\begin{cases}
0 & \text{if } |w| < t \\
w & \text{otherwise}
\end{cases}
\label{eq:pruning_rule}
\end{equation}

In this study, we do not implement pruning explicitly. Rather, we adopt the Global MP thresholding concept as a diagnostic tool to assess the inherent sparsity of the trained network. By quantifying the proportion of weights falling below various threshold levels, we investigate the underlying weight distribution and structural redundancy, without modifying the network's architecture or altering its predictive performance. This approach allows for a nuanced evaluation of sparsity across different configurations.

To facilitate this analysis, sparsity is compared between two configurations: a baseline convolutional neural network (CNN) and its Bayesian counterpart utilizing MC Dropout. This comparison is pertinent because CP operates post hoc and does not influence the training process or the learned weight distribution. In contrast, the Bayesian framework incorporates a prior, specifically, a zero-mean Gaussian prior, over the weights. This prior promotes weight shrinkage by penalizing large values, thereby encouraging weights to cluster around zero. Consequently, this regularizing effect contributes to higher weight sparsity, which is captured by analyzing the number of parameters falling below predefined magnitude thresholds.

\subsubsection{Conformal Prediction Evaluation Metrics}
\label{sec:metrics3}

\vspace{1em}
{\normalfont\itshape Validity\par}
\vspace{0.3em}
The validity of Conformal Prediction (CP) refers to the extent to which the prediction sets include the true class label, as guaranteed by a predefined confidence level \( 1 - \alpha \), where \( \alpha \) is the significance level \citep{Shafer2008}. This property is quantified through empirical coverage, defined as the proportion of test instances for which the true class lies within the corresponding prediction set \citep{zhou2025conformalpredictiondataperspective}. Mathematically, empirical coverage is computed as:

\begin{equation}
\text{Coverage} = \frac{1}{n - v} \sum_{i = v + 1}^{n} \mathbf{1}(y_i \in \Gamma^{\alpha}(x_i))
\label{eq:conformal_coverage}
\end{equation}

Here, \( y_i \) represents the true label of the \( i \)-th test sample, \( \Gamma^{\alpha}(x_i) \) denotes the prediction set for the corresponding input \( x_i \), and \( \mathbf{1}(\cdot) \) is the indicator function that returns 1 if the condition is satisfied and 0 otherwise. The summation is computed over the test set, indexed from \( v+1 \) to \( n \). 

In implementation, coverage is estimated by checking whether the true label is present in each prediction set, followed by averaging over all test instances. If the empirical coverage falls significantly below the nominal level \( 1 - \alpha \), this may indicate a violation of the exchangeability assumption, thereby compromising the theoretical guarantees of CP.

\vspace{1em}
{\normalfont\itshape Efficiency\par}
\vspace{0.3em}
Efficiency measures how informative or precise the prediction sets are, with smaller sets indicating higher efficiency \citep{Fontana2023}. In classification problems, efficiency is commonly assessed by computing the average number of labels included in the prediction sets across the test set. Formally, it is expressed as:

\begin{equation}
\text{Efficiency} = \frac{1}{n - v} \sum_{i = v + 1}^{n} \left| \Gamma^{\alpha}(x_i) \right|
\label{eq:conformal_efficiency}
\end{equation}

In this equation, \( \Gamma^{\alpha}(x_i) \) is the prediction set for input \( x_i \), derived at significance level \( \alpha \), and the summation is performed over all test instances. A prediction set containing only a single label is considered most efficient, as it reflects maximum confidence in the model's prediction. While CP guarantees validity under minimal assumptions, efficiency depends on the model’s capacity to distinguish between classes. Thus, evaluating both validity and efficiency provides a comprehensive understanding of a model’s uncertainty quantification performance.

\subsubsection{Bayesian Inference Evaluation Metrics}
\label{sec:metrics}

\vspace{1em}
{\normalfont\itshape Predictive Entropy (Class-Level and Sample-Level)\par}
\vspace{0.3em}
Predictive entropy quantifies the overall uncertainty in the model’s output distribution for a given input and is particularly relevant in Bayesian inference settings. It is computed as the entropy of the mean predictive distribution obtained from multiple stochastic forward passes using MC Dropout \citep{Gal2016}. Formally, it is defined as:

\begin{samepage}
\begin{equation}
H(p) = - \sum_{c=1}^{C} p_c \log p_c
\label{eq:predictive_entropy}
\end{equation}
\end{samepage}

where \( p_c \) denotes the mean predictive probability for class \( c \), averaged over the Monte Carlo samples. A higher entropy value indicates greater uncertainty in the model's prediction, typically observed when predicted probabilities are evenly distributed across multiple classes. Conversely, a lower entropy value, approaching zero, reflects high model confidence concentrated on a single class prediction.

In this study, predictive entropy is evaluated at both the sample and class levels. At the sample level, it serves to quantify the model’s confidence in individual predictions. At the class level, the entropy is averaged across all test samples belonging to each class, offering insights into which categories the model finds more ambiguous or uncertain. This dual-level analysis supports a more nuanced understanding of model performance, particularly in high-dimensional classification tasks.

\vspace{1em}
{\normalfont\itshape Expected Calibration Error (ECE)\par}
\vspace{0.3em}
The Expected Calibration Error (ECE), introduced by \citet{Naeini2015}, is a widely used metric to assess the calibration of predicted probabilities. It quantifies the average discrepancy between a model’s predicted confidence and the actual accuracy, thereby evaluating how well the confidence scores reflect true correctness \citep{Guo2017}.

To compute ECE, predicted probabilities are first grouped into \( M \) equally spaced bins according to their confidence levels. For each bin, the absolute difference between the average predicted confidence and the empirical accuracy is computed. The final ECE is the weighted average of these differences across all bins, formulated as follows:

\begin{equation}
\text{ECE} = \sum_{m=1}^{M} \frac{|B_m|}{n} \left| \text{acc}(B_m) - \text{conf}(B_m) \right|
\label{eq:ece}
\end{equation}

Here, \( B_m \) denotes the set of predictions falling into the \( m \)-th confidence bin, \( \text{acc}(B_m) \) is the empirical accuracy within the bin, \( \text{conf}(B_m) \) is the average predicted confidence, and \( n \) is the total number of predictions. A lower ECE indicates that the model’s confidence estimates are well-calibrated, i.e., closely aligned with actual prediction accuracy, while a higher ECE suggests over- or under-confidence.

In Bayesian methods, which yield predictive distributions rather than single-point estimates, ECE plays a crucial role in evaluating the reliability of uncertainty quantification. High ECE values imply that confidence scores may be misleading, thereby compromising the interpretability and safety of uncertainty, driven decisions. Notably, ECE is not applicable to Conformal Prediction (CP), since CP produces sets of possible labels rather than scalar confidence values. In such cases, alternative evaluation criteria such as validity and efficiency are used instead.

\vspace{1em}
{\normalfont\itshape Standard Deviation Across Samples\par}
\vspace{0.3em}
While entropy provides a measure of total predictive uncertainty, combining both aleatoric (data-driven) and epistemic (model-driven) sources, it does not distinguish between them. In contrast, variance across predictions offers a direct approximation of epistemic uncertainty by capturing the variability of model outputs across multiple stochastic forward passes \citep{Gal2016}. High variability implies significant epistemic uncertainty, indicating that the model is uncertain about its internal parameters.

In this study, standard deviation is computed per class for each test input. Specifically, each input is passed through the network multiple times using MC Dropout, resulting in a distribution of softmax outputs. The standard deviation of these outputs across all passes reflects the uncertainty associated with each class prediction. These values are then aggregated to estimate the overall epistemic uncertainty at the sample level.

To assess the consistency of uncertainty estimates across methods, we compare the predictive entropy derived from MC Dropout with the prediction set sizes generated by Conformal Prediction (CP). In cases of high certainty, we expect CP to produce smaller prediction sets and MC Dropout to yield lower entropy values. This comparison provides insights into the coherence of uncertainty estimates between the two frameworks.

\vspace{1em}
{\normalfont\itshape Mutual Information (Epistemic Uncertainty)\par}
\vspace{0.3em}
Mutual Information (MI) is a more refined measure for isolating epistemic uncertainty. While predictive entropy accounts for both aleatoric and epistemic components, MI quantifies the reducible portion of uncertainty that arises due to uncertainty in model parameters. It is defined as the difference between the entropy of the mean predictive distribution and the average entropy across individual stochastic forward passes:

\begin{equation}
\text{MI}(y, \theta \mid x) = H\left[\bar{p}(y \mid x)\right] - \frac{1}{T} \sum_{t=1}^{T} H\left[p_t(y \mid x)\right]
\label{eq:mutual_information}
\end{equation}

In this equation, \( H[\bar{p}(y \mid x)] \) denotes the entropy of the average predictive distribution, and \( H[p_t(y \mid x)] \) is the entropy of the prediction in the \( t \)-th stochastic pass. The difference quantifies the extent to which the model's predictions fluctuate under dropout, providing a clear indicator of epistemic uncertainty.

Mutual Information is particularly useful in identifying uncertainty driven by model ambiguity, which often occurs in regions of the input space with limited or conflicting training data. It complements other metrics such as predictive entropy by enabling a more granular understanding of the uncertainty decomposition.

\vspace{1em}
{\normalfont\itshape Average Entropy (Aleatoric Uncertainty)\par}
\vspace{0.3em}
Average entropy across multiple stochastic forward passes serves as an approximation of aleatoric uncertainty, which arises from intrinsic noise or ambiguity in the input data. Unlike epistemic uncertainty, which can be reduced with more data or better modeling, aleatoric uncertainty reflects irreducible randomness and is often due to overlapping class boundaries or low-quality observations.

This metric is derived by averaging the entropy of the softmax outputs obtained from each of the \( T \) MC Dropout passes. Formally, it is computed as:

\begin{equation}
\frac{1}{T} \sum_{t=1}^{T} H\left[p_t(y \mid x)\right]
\label{eq:average_entropy}
\end{equation}

Here, \( H[p_t(y \mid x)] \) denotes the entropy of the predicted class probabilities in the \( t \)-th stochastic pass. The resulting average quantifies the level of uncertainty inherent in the data for a given input.
Together with mutual information, average entropy facilitates the decomposition of total predictive uncertainty into epistemic and aleatoric components, enabling a more nuanced understanding of model confidence and decision reliability.

\vspace{0.3em}
In summary, the evaluation metrics discussed in this section collectively provide a comprehensive framework for assessing both the predictive performance and the quality of uncertainty quantification methods. By combining classical measures such as sparsity and calibration with more advanced probabilistic metrics, this study enables a deeper and more interpretable analysis of model reliability across different uncertainty estimation techniques.

\section{Empirical Analysis}
\label{sec:empirical2}

This section presents the empirical results across several performance metrics, including classification accuracy, uncertainty quantification, prediction set efficiency, and validity. Although the H-CNN VGG16 architecture produces multiple hierarchical outputs, our analysis focuses exclusively on the final predictions for the ten Fashion-MNIST classes.
We begin by evaluating baseline performance in terms of accuracy, overfitting, sparsity, and Expected Calibration Error (ECE) for both H-CNN VGG16 and GoogLeNet, as well as their Bayesian counterparts. We then assess predictive reliability using Conformal Prediction, focusing on prediction set sizes, empirical coverage, and class-wise confidence variation.

Next, we investigate uncertainty through Bayesian approximation using Monte Carlo Dropout. We analyse overall model uncertainty by visualising the distribution and confidence intervals of predictive entropy, and decompose this uncertainty into epistemic and aleatoric components. We compare predictive entropy across correct classifications and misclassifications to explore how uncertainty relates to prediction correctness. To gain deeper insight into class-level ambiguity, we conclude this section with class-wise comparisons of predictive entropy and corresponding confidence intervals for both models.

We then extend the analysis by examining the relationship between empirical and probabilistic reliability measures, specifically comparing Conformal Prediction set sizes with predictive entropy derived from Bayesian inference. Collectively, these analyses shed light on how the two architectures express and manage uncertainty under different estimation frameworks, and how this impacts both predictive reliability and model behaviour.

\subsection{Overall Performance}
\label{sec:overall_performance}
We begin by evaluating the H-CNN VGG16 and GoogLeNet architectures, along with their Bayesian counterparts, across multiple performance dimensions including accuracy, training duration, sparsity, overfitting tendencies, and Expected Calibration Error (ECE). 

For classification accuracy on the Fashion-MNIST dataset, the best-performing H-CNN VGG16 model achieves 92.99\%, with a five-fold cross-validation average of 92.62\%. In comparison, GoogLeNet attains a maximum accuracy of 89.72\%, with an average of 88.24\% across folds. These results are consistent with prior studies where \citet{Seo2019} reported an accuracy of 93.52\% for H-CNN VGG16, while \citet{Vives-Boix2021} documented GoogLeNet achieving 91.89\%.

\subsubsection{Accuracy and Duration}

\begin{table}[htbp]
\centering
\caption{Performance summary for H-CNN VGG16 and GoogLeNet architectures}
\label{tab:performance_summary}
\small
\begin{tabular}{l p{3cm} c c}
\toprule
\textbf{Model} & \textbf{Metric} & \textbf{Baseline} & \textbf{Bayesian} \\
\midrule
\multirow{7}{*}{\shortstack[l]{H-CNN\\VGG16}} 
    & Accuracy (Best)             & 92.99\%       & 92.47\%       \\
    & Accuracy (5-Fold Avg.)      & 92.62\%       & 92.29\%       \\
    & Duration (s)                & 12,342.65     & 13,417.11     \\
    & Trainable Parameters        & \multicolumn{2}{c}{90,312,274} \\
    & Non-Trainable Params    & \multicolumn{2}{c}{2,944}       \\
    & Optimizer Parameters        & \multicolumn{2}{c}{90,312,276} \\
    & \textbf{Total Parameters}   & \multicolumn{2}{c}{\textbf{180,627,494}} \\
\midrule
\multirow{6}{*}{\shortstack[l]{GoogLe\\Net}} 
    & Accuracy (Best)             & 89.72\%       & 88.68\%       \\
    & Accuracy (5-Fold Avg.)      & 88.24\%       & 87.49\%       \\
    & Duration (s)                & 1,428.88      & 1,471.92      \\
    & Trainable Parameters        & \multicolumn{2}{c}{5,977,530}  \\
    & Optimizer Parameters        & \multicolumn{2}{c}{5,977,532}  \\
    & \textbf{Total Parameters}   & \multicolumn{2}{c}{\textbf{11,955,062}} \\
\bottomrule
\end{tabular}
\end{table}

Although H-CNN VGG16 outperforms GoogLeNet in accuracy, this comes at a significant computational cost. With more than 180 million parameters, substantially exceeding the parameter count of GoogLeNet, H-CNN VGG16 requires markedly longer training times. On average, each fold takes nearly ten times longer to complete. This stark difference in model complexity directly underlies the observed gap in computational efficiency between the two architectures.

After implementing the Bayesian approximation using Monte Carlo Dropout, both architectures exhibit a slight decrease in best-case accuracy. This decrease is attributable to the stochastic nature of MC Dropout, which introduces additional variance into the predictions.  Furthermore, inference time increases substantially, as each prediction requires 50 stochastic forward passes. Nevertheless, under the Bayesian setting, H-CNN VGG16 continues to outperform GoogLeNet in terms of classification accuracy.

Overall, both the standard and Bayesian variants of H-CNN VGG16 outperform their GoogLeNet counterparts in terms of classification accuracy. However, this improvement comes at a significantly higher computational cost. While the Bayesian implementation further increases training duration per fold, it yields only marginal gains in accuracy, suggesting limited efficiency benefits relative to its added complexity.

\subsubsection{Class-Wise Accuracy Confusion Matrices}
\begin{figure}[h]
    \centering
    \includegraphics[width=0.45\textwidth]{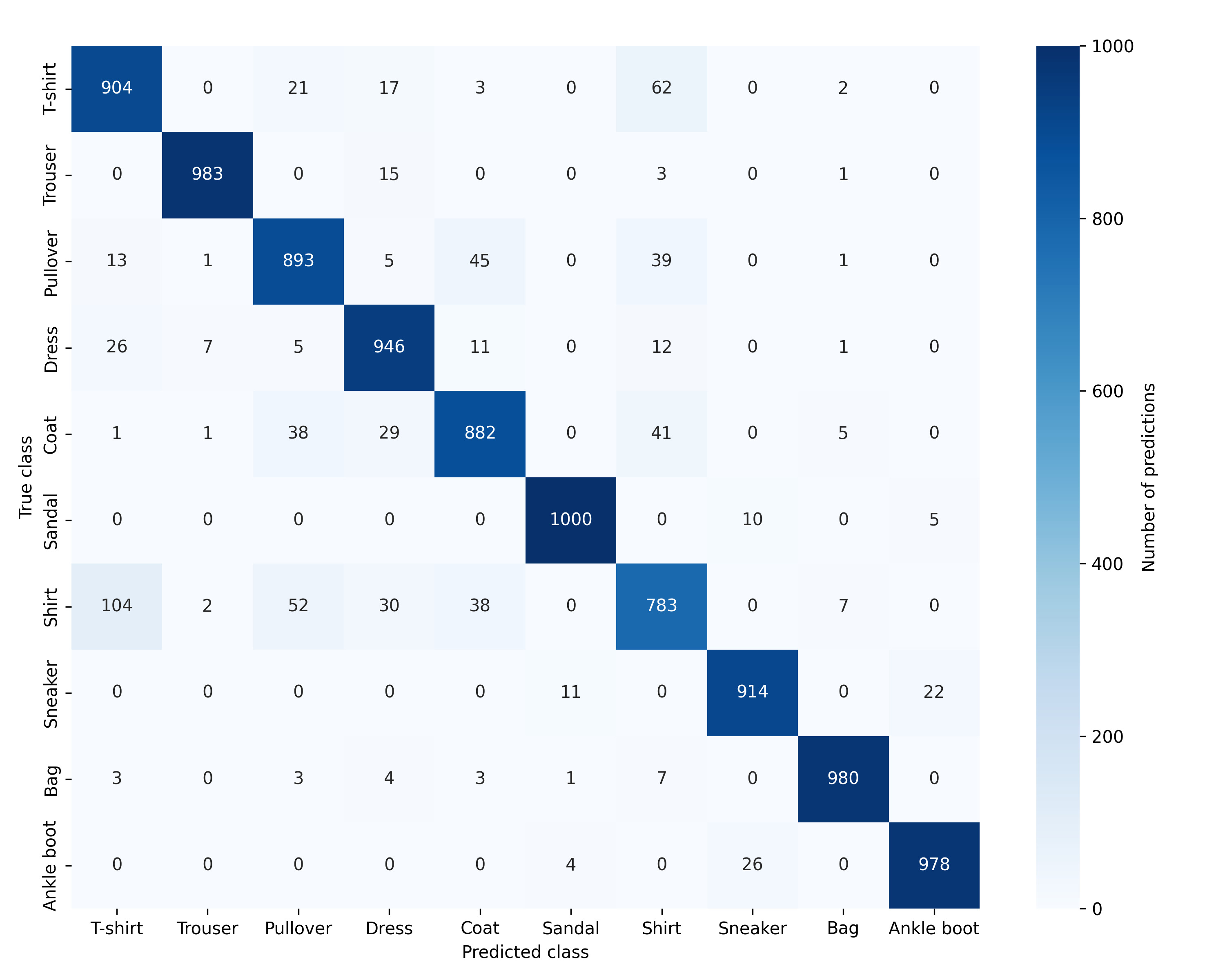}
     \includegraphics[width=0.45\textwidth]{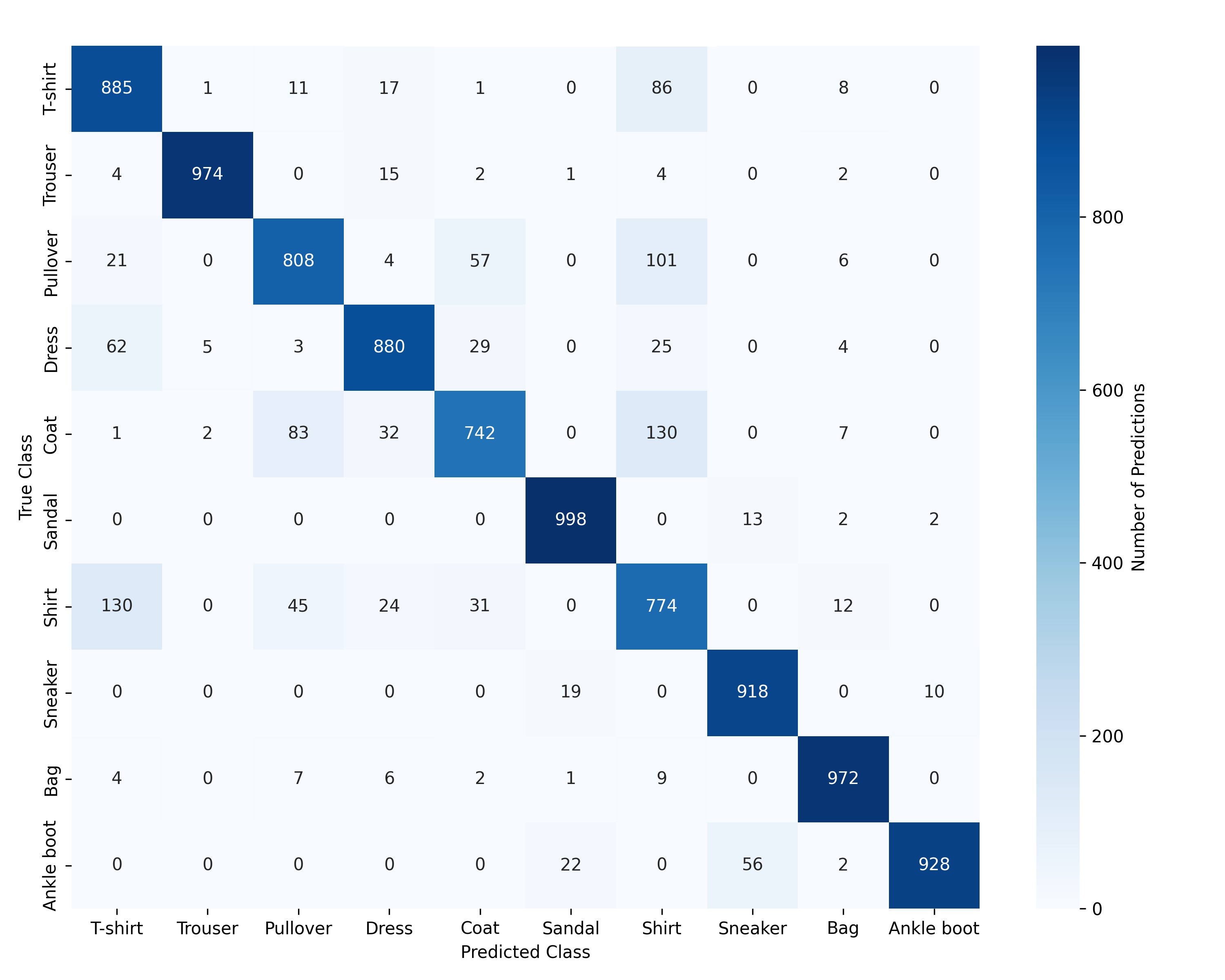}
    \caption{Confusion Matrix for H-CNN VGG16 and GoogLeNet}
    \label{fig:Figure.1}
\end{figure}

To further evaluate model performance beyond overall accuracy, confusion matrices were generated to examine class-wise prediction behaviour (Figure ~\ref{fig:Figure.1}). The general findings of the baseline models are consistent with those of their Bayesian alternatives. To avoid redundancy, the Bayesian confusion matrices are provided in the Appendix, H-CNN VGG16 and GoogLeNet in Figure~\ref{fig:19}. These visualizations highlight systematic misclassification patterns and reveal which classes are most frequently confused with one another.
Importantly, such recurring misclassification trends also provide a foundation for the subsequent analysis of uncertainty measures, where predictive entropy and calibration are used to quantify the reliability of the models’ class-level decisions. 

The H-CNN VGG16 model demonstrates strong classification performance across most categories as can be seen in  (Figure ~\ref{fig:Figure.1} (left)), achieving perfect accuracy for Sandal (1000) and high accuracy for Trouser (983), Bag (980), Ankle boot (974), Sneaker (914). These items are visually distinct, contributing to the model’s high performance. By contrast, the highest concentration of misclassifications occurs in the Shirt class, which is frequently confused with T-shirt (104), Pullover (52), and Coat (38). This reflects substantial visual similarity among these categories. This confusion also occurs in reverse where T-shirt are often predicted as Shirts (62), and occasionally as Pullovers (39) or Coats (41). Similarly, pullovers are sometimes misclassified as Coats (45), underscoring the model’s difficulty distinguishing between classes with overlapping visual features. 

%
%
%

For GoogLeNet (Figure ~\ref{fig:Figure.1} (right)), the overall classification patterns are broadly similar.  While the model demonstrates strong performance across most categories, it struggles notably with Shirts (670 correct out of 1000) and Coats (759). Shirts are frequently misclassified as T-shirts (112) or Pullovers (47), whereas Coats are often predicted as Shirts (132) or Pullovers (68). These patterns are similar to the trends observed in the H-CNN VGG16 model, indicating that these classes are inherently more difficult to separate, most likely due to substantial visual overlap in their features.

Overall, both models demonstrate strong classification performance on visually distinct items, yet face difficulties classifying classes that exhibit substantial visual overlap, particularly Shirt, Pullover, and Coat.

\subsubsection{Overfitting Analysis}
\begin{figure}[H]
    \centering
    \includegraphics[width=0.8\textwidth]{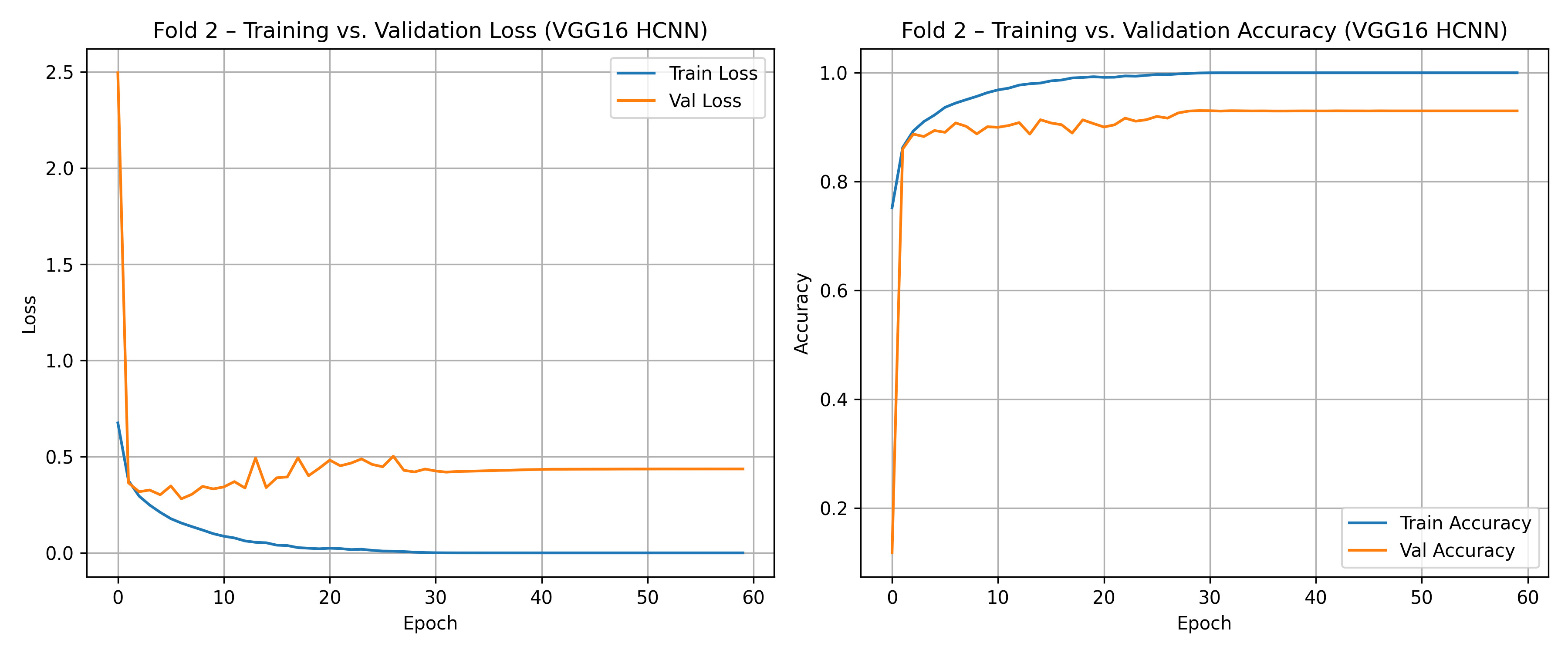}
    \caption{Training vs. Validation Loss and Accuracy for H-CNN VGG16 (Fold 2)}
    \label{fig:Figure.3}
\end{figure}

To evaluate the extent of overfitting in both baseline models and their Bayesian counterparts, training and validation loss and accuracy curves were analyzed across 60 epochs. All models were trained using 5-fold cross-validation, however, for clarity, only the best-performing fold is presented for each baseline model (Figure ~\ref{fig:Figure.3} and Figure ~\ref{fig:Figure.4}). The plots for the remaining folds, along with their Bayesian variants, are provided in Appendix Figures~\ref{fig:21}--\ref{fig:34}, as they do not exhibit substantial differences from the non-Bayesian counterparts.

The training curves for H-CNN VGG16 reveal a pronounced gap between training and validation performance. While training loss steadily decreases and accuracy approaches 100\%, validation loss plateaus early and exhibits minor fluctuations, with validation accuracy stabilising around 93\%. This pattern indicates that the model fits the training data well but shows limited improvement on unseen data, suggesting a degree of overfitting.  The architecture used here follows the design by \citet{Seo2019}, which includes dropout, batch normalization, and loss weighting to support generalisation. These techniques appear effective in stabilising the training process, although the model’s substantial  parameter count likely contributes to its tendency to overfit.

\begin{figure}[H]
    \centering
    \includegraphics[width=0.8\textwidth]{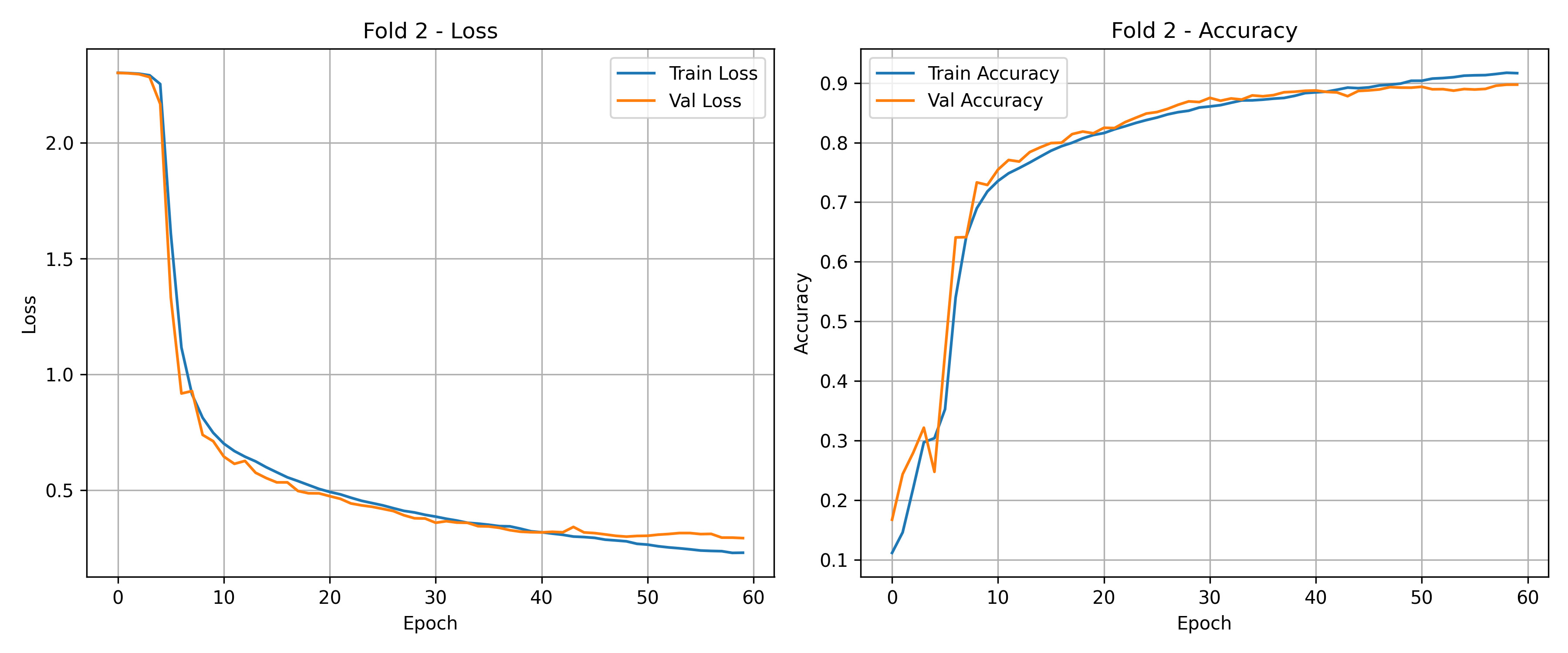}
    \caption{Training vs. Validation Loss and Accuracy for GoogLeNet (Fold 2)}
    \label{fig:Figure.4}
\end{figure}

GoogLeNet, on the other hand, demonstrates more consistent training behaviour. Training and validation losses decrease in parallel, and validation accuracy closely follows training accuracy throughout, indicating good generalisation and limited overfitting. This implementation, adapted from \citet{Vives-Boix2021} for the Fashion-MNIST dataset, benefits from a more compact architecture and a substantially lower parameter count, which likely contribute to its stable performance.

In both architectures, introducing MC Dropout during inference had minimal impact on the training dynamics. The Bayesian models showed nearly identical learning curves to their respective baselines, which supports the decision to move those plots to the Appendix for reference.

In summary, H-CNN VGG16 achieves higher training accuracy but exhibits moderate overfitting, whereas GoogLeNet maintains a more balanced relationship between training and validation performance. Considering its computational efficiency and stronger generalisation, GoogLeNet may represent the more practical choice in scenarios where resource constraints or overfitting risks are critical concerns.

\subsubsection{Sparsity}
\label{sec:sparsity}
\begin{figure}[H]
    \centering
    \includegraphics[width=0.5\textwidth]{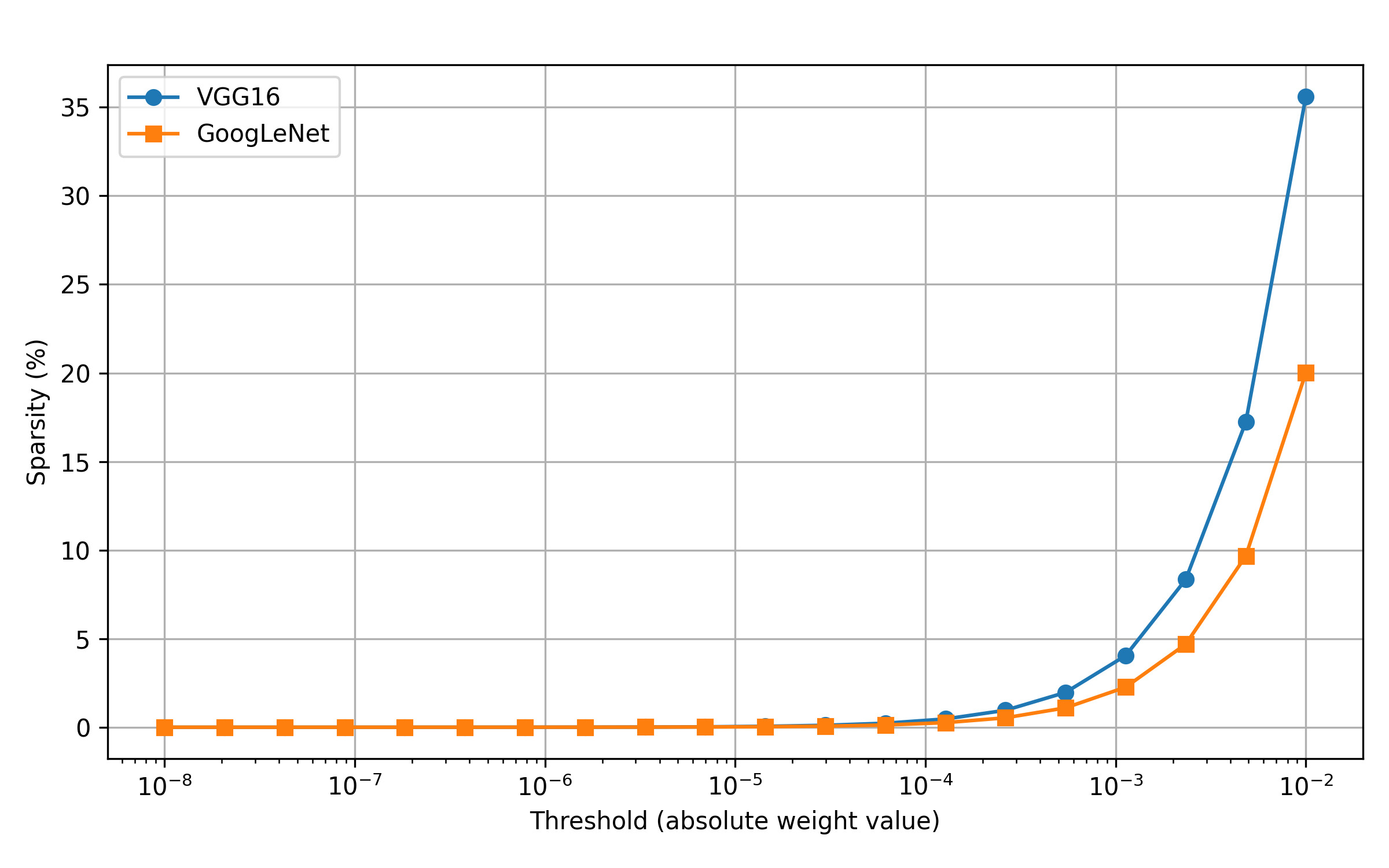}
    \caption{Sparsity vs. Threshold. \textit{Note.} Sparsity as a percentage of total trainable weights}
    \label{fig:Figure.5}
\end{figure}

As introduced in Section 4, sparsity is examined here through both graphical and tabular summaries for the H-CNN VGG16 and GoogLeNet models (Figure~\ref{fig:Figure.5}). For brevity, this section presents only the visualisations and tables corresponding to the MC Dropout implementation, while the baseline visualisations and weight tables are provided in the Appendix Figures~\ref{fig:39} and Table~\ref{tab:sparsity_table}.

The plots provide a visual overview of the cumulative sparsity across a range of thresholds. Both models exhibit similar elbow-shaped curve in their sparsity profiles. For both models, sparsity remains low at small thresholds but begins to increase sharply around 0.001. Closer inspection of this region, however, reveals important differences in how the two architectures distribute their weights, highlighting distinct sparsity patterns.

The H-CNN VGG16 model contains significantly more weights, which is expected given its substantially larger number of parameters. It also exhibits a higher proportion of weights close to zero. This is a noteworthy observation considering the model's  earlier difficulties with generalisation, as discussed above. According to the magnitude pruning framework proposed by \citet{gupta2024complexityrequiredneuralnetwork}, such near-zero weights contribute  little to model performance and therefore constitute strong candidates for pruning.

\begin{table}[htbp]
\centering
\caption{Sparsity Comparison Across Thresholds for   (1) Bayesian H-CNN VGG16 and (2) GoogLeNet}
\renewcommand{\arraystretch}{1.15}
\setlength{\tabcolsep}{2pt}
\scriptsize

\begin{tabular}{@{}l
                >{\raggedleft\arraybackslash}p{1.2cm}
                >{\raggedleft\arraybackslash}p{1cm}
                >{\raggedleft\arraybackslash}p{0.7cm}
                >{\raggedleft\arraybackslash}p{0.5cm}
                >{\raggedleft\arraybackslash}p{0.7cm}
                >{\raggedleft\arraybackslash}p{0.5cm}@{}}
\toprule
\textbf{Weight Range} &
\multicolumn{2}{c}{\textbf{\# Weights}} &
\multicolumn{2}{c}{\textbf{\% Total}} &
\multicolumn{2}{c}{\textbf{Cum. \%}} \\
\cmidrule(lr){2-3}\cmidrule(lr){4-5}\cmidrule(lr){6-7}
 & (1) & (2) & (1) & (2) & (1) & (2) \\
\midrule
\textless{}0.00001     & 34,427     & 1,548     & 0.04  & 0.03  & 0.04   & 0.03   \\
0.00001–0.00005        & 134,757    & 4,812     & 0.15  & 0.08  & 0.19   & 0.11   \\
0.00005–0.0001         & 165,450    & 6,033     & 0.18  & 0.10  & 0.37   & 0.21   \\
0.0001–0.0005          & 1,297,613	  & 48,000    & 1.44  & 0.80  & 1.81   & 1.01   \\
0.0005–0.001           & 1,609,454	  & 59,512    & 1.78  & 1.00  & 3.59   & 2.01   \\
0.001–0.005            & 12,857,487 & 478,938    & 14.24 & 8.01  & 17.83  & 10.04  \\
$\ge$ 0.005            & 74,216,030 & 5,377,523 & 82.17 & 89.96 & 100 & 100 \\
\bottomrule
\end{tabular}
\end{table}

In contrast, GoogLeNet demonstrates a more efficient utilisation of its smaller parameter set. The increase in sparsity across thresholds is more gradual, with fewer weights falling below the different threshold values. This supports previous observations that GoogLeNet’s more compact architecture is structurally more constrained, thereby promoting more effective generalisation.

In particular, the implementation of MC Dropout does not significantly affect the weight sparsity of the H-CNN VGG16 or GoogLeNet models. This is expected, as MC Dropout is applied during inference to estimate predictive uncertainty and does not influence the underlying weight magnitudes during training. As a result, the sparsity pattern remains unchanged and the models are not further compressed. However, this stands in contrast to a full Bayesian approach, where the choice of prior distributions can induce regularisation during training \citep{Abdar2021,9745083}.

\subsubsection{Expected Calibration Error (ECE)}
\label{sec:ece}

Expected Calibration Error (ECE) quantifies the discrepancy between predicted confidence and actual accuracy. For instance, a perfectly calibrated model would assign 80\% confidence to predictions that are correct precisely 80\% of the time. Lower ECE values therefore indicate better calibration, and a stronger alignment between confidence and correctness. This study uses ECE to evaluate how effectively the Bayesian versions of H-CNN VGG16 and GoogLeNet capture and express predictive uncertainty.

\begin{table}[ht]
\centering
\caption{Comparison of ECE before and after Bayesian modeling}
\label{tab:ece_comparison}
\begin{tabular}{lcc}
\toprule
\textbf{Architecture} & \textbf{ECE (Baseline)} & \textbf{ECE (Bayesian)} \\
\midrule
H-CNN VGG16      & 5.66\% & 5.61\% \\
GoogLeNet  & 2.82\% & 1.37\% \\
\bottomrule
\end{tabular}
\end{table}

The results presented above show a clear contrast between the two models. GoogLeNet demonstrates a significant improvement in calibration, with  ECE dropping from 2.82\% in the baseline model to 1.37\% under  Bayesian inference. This indicates that Monte Carlo Dropout effectively improves the model’s ability to reflect uncertainty in its confidence scores. In contrast, H-CNN VGG16 shows only a marginal improvement, with ECE decreasing slightly from 5.66\% to 5.61\%. Even with Bayesian inference, the model remains comparatively poorly calibrated relative to  GoogLeNet. A more detailed interpretation of these findings is provided in Section 5.2.2.

\subsection{Uncertainty Estimation}
\label{sec:uq total}

This section provides a rigorous empirical analysis of Conformal Predictions and Bayesian approximation with MC Dropout. Both methods are evaluated using multiple metrics to assess the predictive reliability across two neural network architectures. Moreover, both approaches are compared to understand the relationship between Conformal Prediction set sizes and predictive entropy derived from Bayesian approximation.

\subsubsection{Conformal Prediction}
\label{sec:conformal}

Conformal Predictions requires an additional calibration split. Therefore, the data set is partitioned into  60,000 observations for training, 2,000 for calibration and 8,000 for testing. This design follows the standard Inductive Conformal Prediction (ICP) framework, in which the calibration set is used to compute conformity scores and derive a quantile threshold that controls  prediction set sizes while ensuring the desired coverage level (validity) on unseen test data \citep{vovk2005algorithmic}. This section also reports the empirical coverage achieved by each model. Because the confidence band is adaptively adjusted, the empirical coverage will not match the nominal 95\% level exactly, as is typical in ICP.

\begin{figure}[H]
    \centering
    \includegraphics[width=0.45\textwidth]{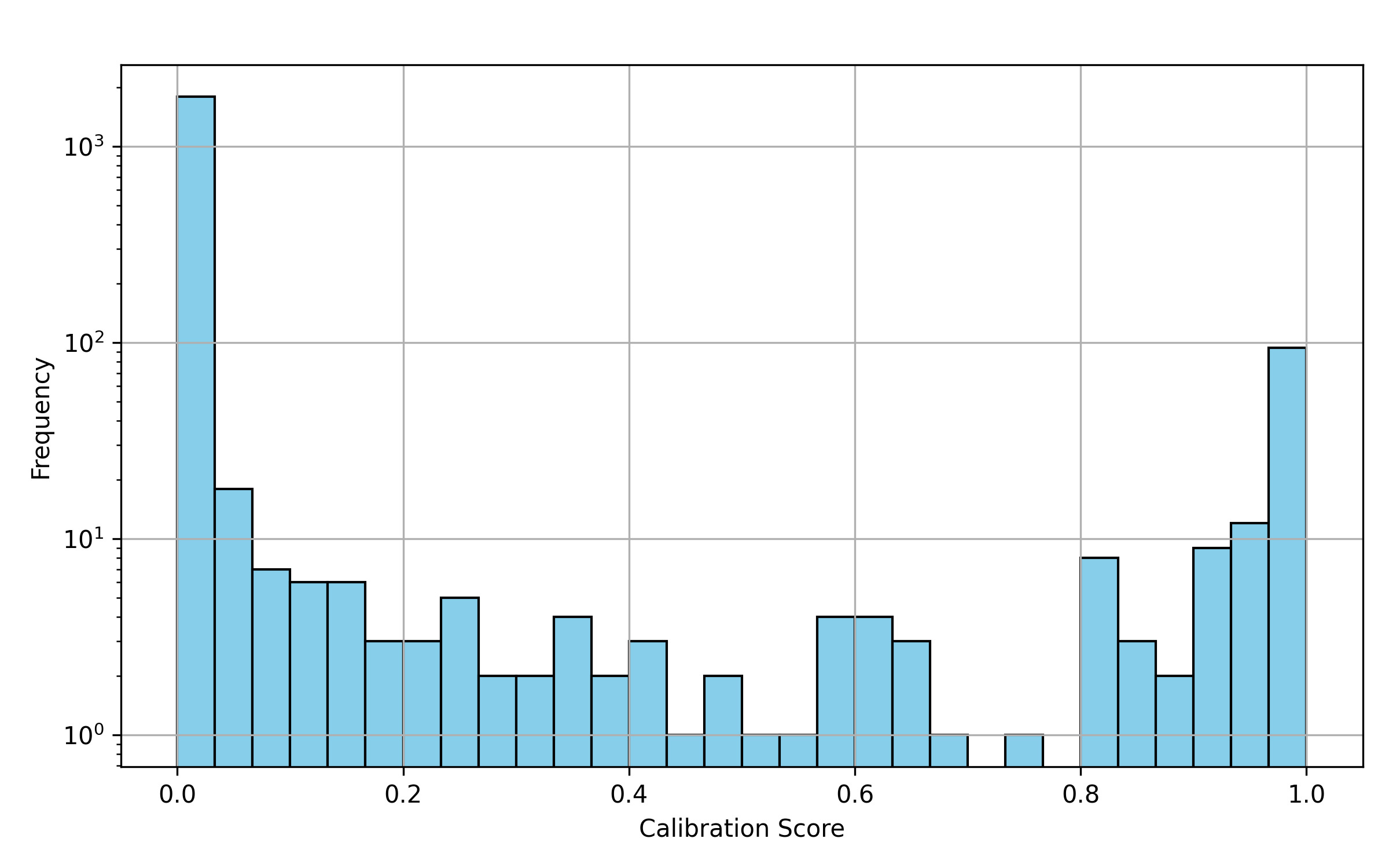}
     \includegraphics[width=0.45\textwidth]{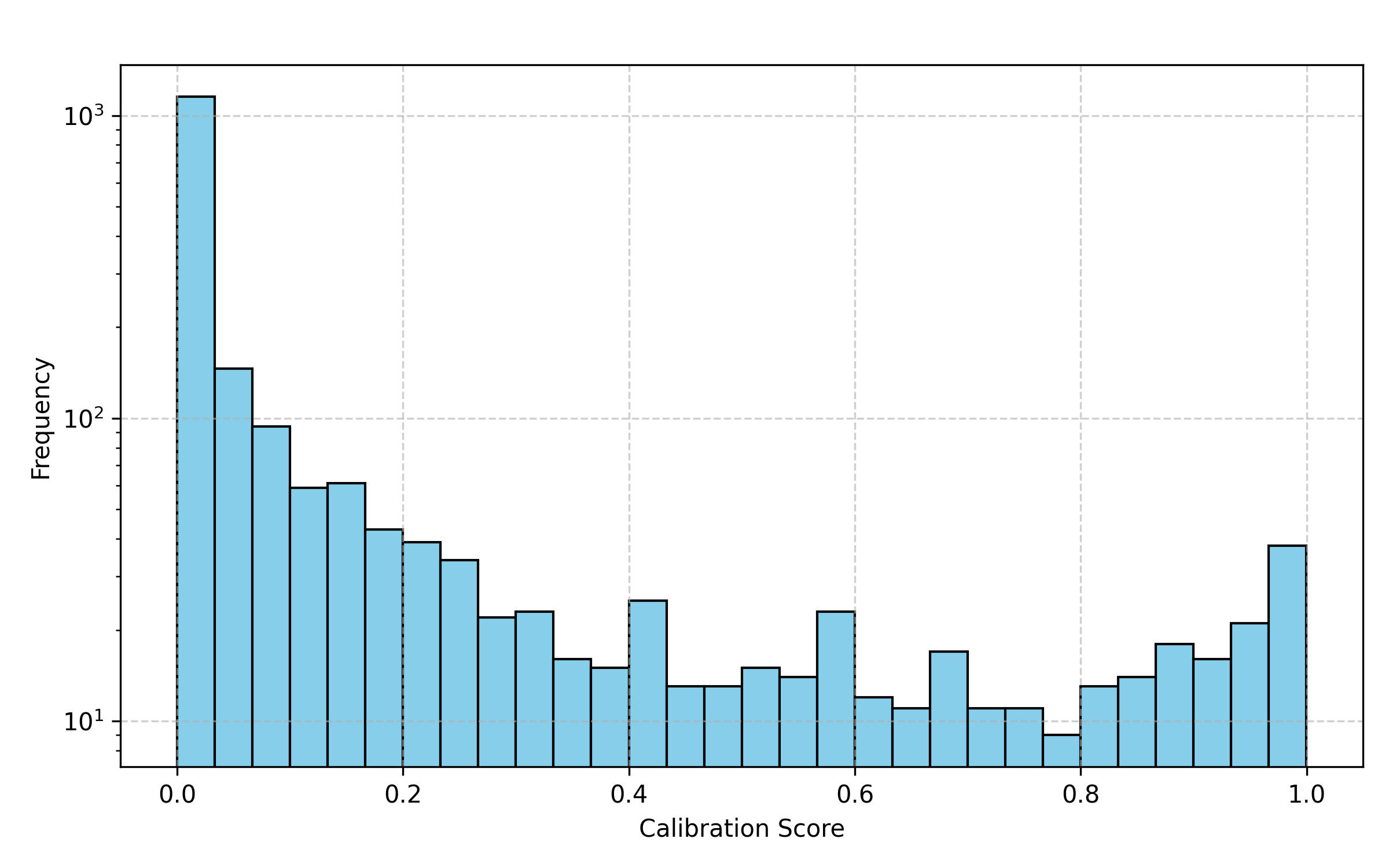}
    \caption{H-CNN VGG16 \& GoogLeNet Calibration Scores}
    \label{fig:Figure.6}
\end{figure}

The histograms below illustrate the distribution of calibration scores for both baseline models (Figure ~\ref{fig:Figure.6}). The scores represent how confident the model was in the prediction of the true label where values closer to zero indicate higher confidence assigned to the true class, whereas larger values suggest greater uncertainty. Both models achieve an empirical coverage of 95\%, demonstrating that the ICP method is well calibrated overall and successfully achieves validity.

For the H-CNN VGG16 model (Figure ~\ref{fig:Figure.6} (right)), the calibration score distribution shows a pronounced peak near zero, indicating that the model assigns high confidence to the majority of its predictions.  This pattern is consistent with the model’s overall high accuracy. However, another peak appears on the right-hand side, suggesting reduced confidence for certain predictions. As shown later in the efficiency plots, this lower-confidence region is primarily associated with the Shirt class.

%
%
%
%

GoogleNet's distribution of calibration scores is broader and less structured, indicating different confidence behaviour (see (Figure ~\ref{fig:Figure.6} (left)). Like H-CNN VGG16, it displays a peak at zero, indicating that both models assign high confidence to many correct predictions. However, GoogLeNet exhibits greater variability overall. In the case of H-CNN VGG16, there is a high peak of calibration scores near 1, meaning that in some cases the model assigns low probability to the true class, leading to lower confidence. This suggests that GoogLeNet is generally less overconfident and adopts a more cautious stance when making predictions.  At the same time, this broader distribution makes it more difficult to judge the reliability of individual predictions based solely on calibration scores.


The information below depicts efficiency evaluation, which is measured via the prediction set size. A smaller set size indicates higher efficiency, meaning the model includes fewer labels in the prediction, making it more certain. If the set size is larger, the model is less confident and tries to include more labels to reach the coverage guarantee.

\vspace{-0.5em}  
\begin{table}[!htbp]
\centering
\caption{Prediction set sizes for GoogLeNet and H-CNN VGG16 models}
\renewcommand{\arraystretch}{1.2}
\begin{tabular}{lcccc}
\toprule
\textbf{Set Size} & 1 & 2 & 3 & 4 \\
\midrule
\textbf{GoogLeNet Count} & 6,431 & 1,377 & 185 & 7 \\
\textbf{H-CNN VGG16 Count} & 7,551 & 398 & 44 & 7 \\
\bottomrule
\end{tabular}
\label{tab:efficiency_combined}
\end{table}

\begin{figure}[H]
    \centering
    \includegraphics[width=0.45\textwidth]{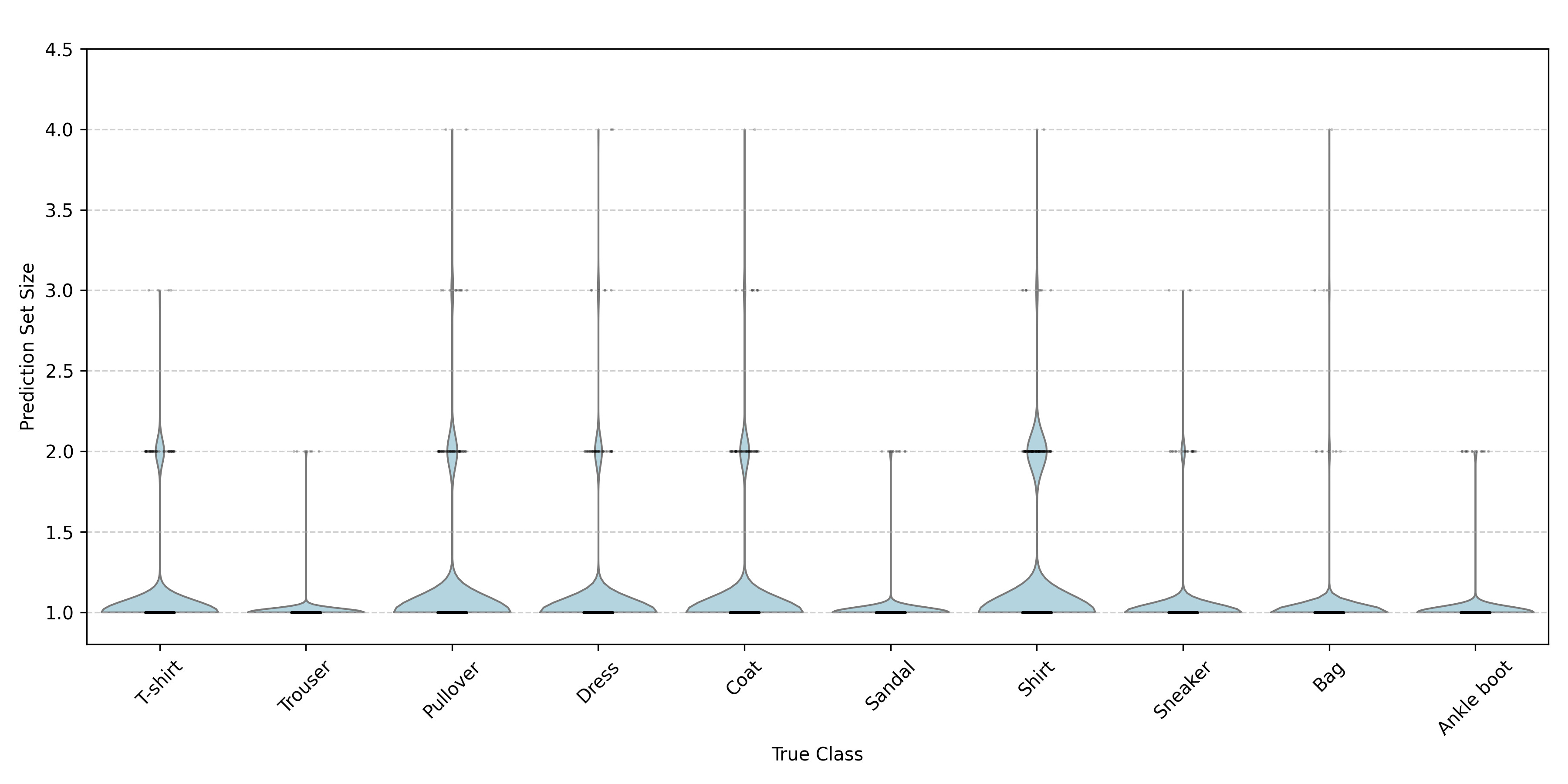}
        \includegraphics[width=0.45\textwidth]{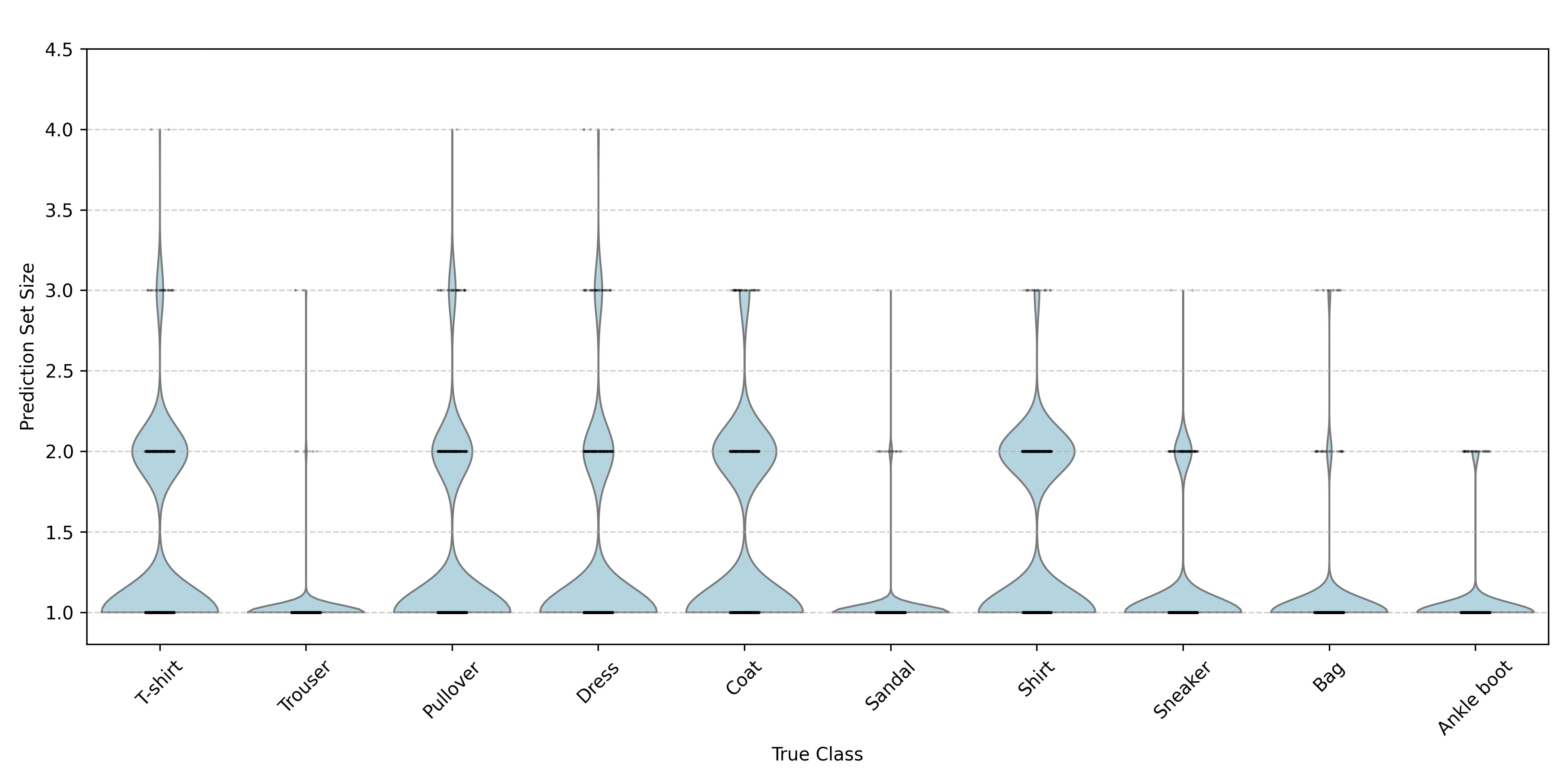}
    \caption{Prediction Set Size Per Class H-CNN VGG16}
    \label{fig:Figure.8}
\end{figure}

In the case of H-CNN VGG16 (Figure ~\ref{fig:Figure.8} (left)), the majority of predictions consist of a single label, indicating high efficiency. This is evident in the plot below, where the distribution of the prediction set size is centred around 1 for most classes. The main exception is the Shirt category, which has a broader distribution at a prediction set size of two, indicating lower model confidence. This observation aligns with the right tail of calibration score distribution discussed above, suggesting that Shirt instances frequently generate high nonconformity scores. Moreover, Pullover appears as the only category with a prediction set size of five, representing a rare case of particularly low confidence. Overall, maintains both high coverage (validity) and compact prediction sets.

GoogLeNet (Figure ~\ref{fig:Figure.8} (right)) produces fewer prediction sets of size one and significantly more sets of size two and three. This pattern corresponds directly to the earlier calibration score histograms, where GoogLeNet exhibited fewer instances of extreme confidence. The distribution of prediction set sizes is also more dispersed compared to H-CNN VGG16. As with the previous model, the Shirt category shows the widest spread, underscoring its difficulty to classify. This finding is consistent with the broader calibration score distribution observed for GoogLeNet, which was less sharply peaked than that of H-CNN VGG16.

%
%
%

Overall, H-CNN VGG16 achieves high empirical coverage with prediction sets that are typically small, but it also exhibits overconfidence, particularly for visually ambiguous classes. At the same time, GoogLeNet produces a broader distribution of confidence scores and generates broader prediction sets, reflecting a greater ability to signal uncertainty when the model is unsure.

\subsubsection{Uncertainty in Bayesian Inference}
\label{sec:bayesian_uncertainty}

Bayesian uncertainty estimation is implemented with MC Dropout, following the approach of \citet{Gal2016}. This method relies on dropout being applied during testing, resulting in the model performing multiple stochastic forward passes to approximate the posterior distribution. In the article, all results are based on fifty Monte Carlo passes per observation, with the mean predictive entropy and its standard deviation computed  from these.

This section provides a general overview of the model's uncertainty and the predictive entropy, which reflects the combined contribution of  aleatoric and epistemic uncertainties, as outlined in Section 4. In essence, predictive entropy quantifies how uncertain the model is on average across its predictions.


Figure ~\ref{fig:Figure.10} below displays predictive confidence values,  sorted from lowest to highest. The dark blue line represents the mean predicted confidence, while the shaded area illustrates the variation (standard deviation) across the fifty dropout passes.

\begin{figure}[H]
    \centering
    \includegraphics[width=0.6\textwidth]{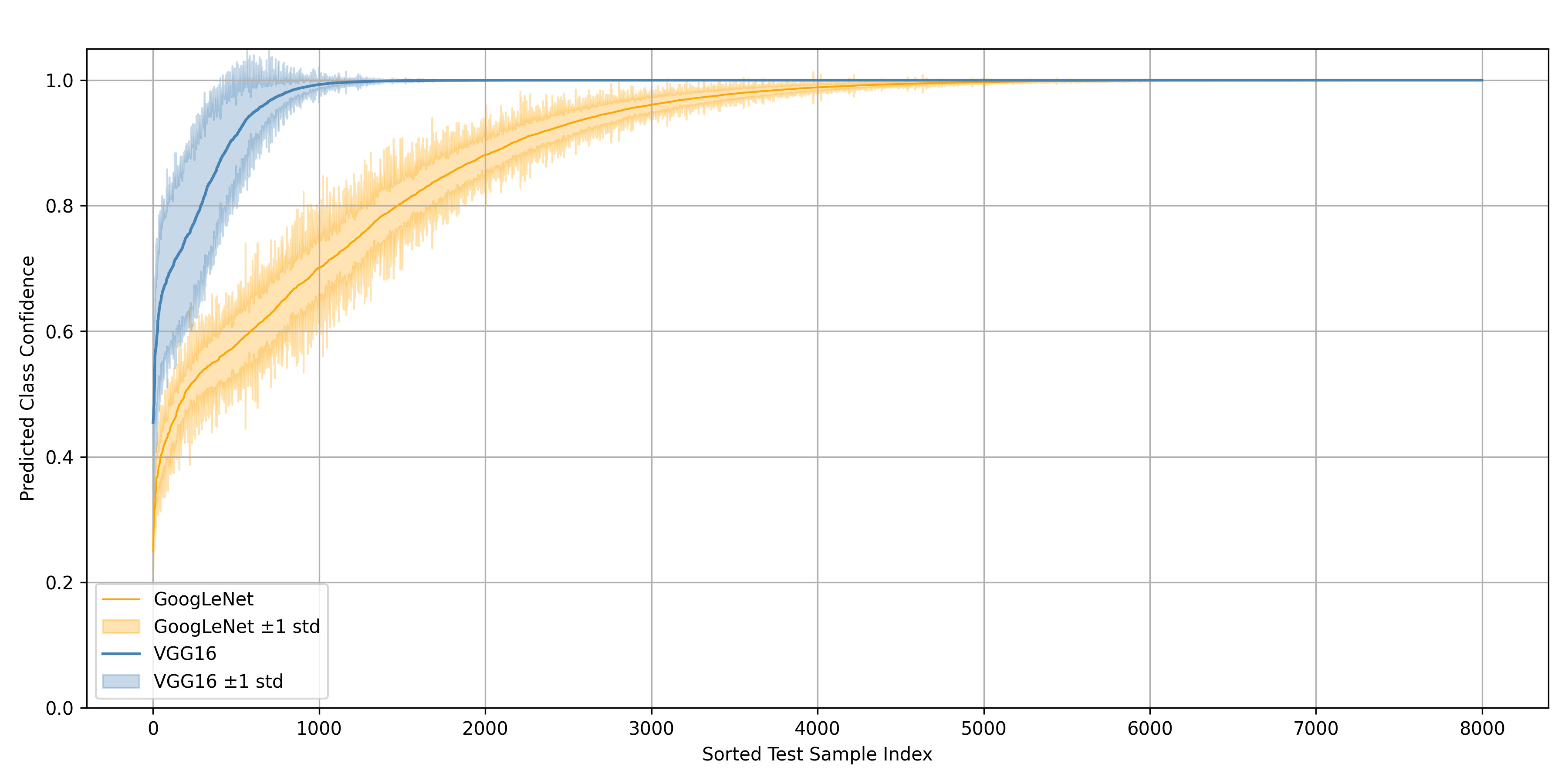}
    \caption{Overall Confidence Intervals Bayesian}
    \label{fig:Figure.10}
\end{figure}


The confidence curve for H-CNN VGG16 rises sharply, reaching near-perfect confidence within the first 1,000 samples. Although the shaded region begins relatively wide, it narrows quickly, indicating that the model becomes highly consistent across dropout passes.  However, this consistency appears excessive; the model shows little hesitation even on inputs that are likely to be ambiguous. This behaviour echoes earlier findings of overfitting, with overconfidence persisting even under the Bayesian setting. The model attains certainty too rapidly and exhibits minimal variation between stochastic passes, suggesting that it underestimates epistemic uncertainty and fails to adequately signal when it is unsure.

For GoogLeNet, the increase in confidence is smooth and more gradual. The model begins with relatively low certainty for the first few hundred samples and then steadily gains confidence, eventually flattening near 1.0. Notably, the variation across dropout passes is most pronounced in the lower-confidence region and diminishes as confidence increases. This behaviour is expected; different dropout passes yield different predictions when the model is uncertain. However, once the model is sure, the variability in predictions diminishes. These dynamics indicate that GoogLeNet not only becomes confident but also appropriately reflects its uncertainty, behaving as expected for a well-calibrated Bayesian model.

To further support the findings on model confidence and uncertainty, predictive entropy distributions were examined for both H-CNN VGG16 and GoogLeNet. As shown in Figure~\ref{fig:41} in the Appendix. H-CNN VGG16's entropy values are predominantly low and right-skewed, reflecting consistently high confidence across most predictions. In contrast, GoogLeNet exhibits a wider range of entropy values with a longer tail, indicating more frequent high-uncertainty predictions. These patterns are consistent with the confidence interval plots discussed earlier; H-CNN VGG16 remains confident and consistent, even on more ambiguous inputs, while GoogLeNet demonstrates greater variability and appears more responsive to uncertainty.


As described in Section 4, total predictive uncertainty can be decomposed  into two main components. Predictive entropy captures the overall uncertainty in the model’s output, combining contributions from both epistemic and aleatoric uncertainties. Epistemic uncertainty, often quantified using mutual information, reflects uncertainty about the model parameters. For example,when the model has limited exposure to similar data and predictions vary across dropout passes. Aleatoric uncertainty, measured through 
the expected entropy, arises from noise or intrinsic ambiguity in the data itself, such as visually similar categories (e.g. Shirt and T-shirt)  that are inherently difficult to distinguish. The following analysis disentangles these two components to provide a more detailed characterisation of each model’s uncertainty profile.

\begin{figure}[H]
    \centering
    \includegraphics[width=0.45\textwidth]{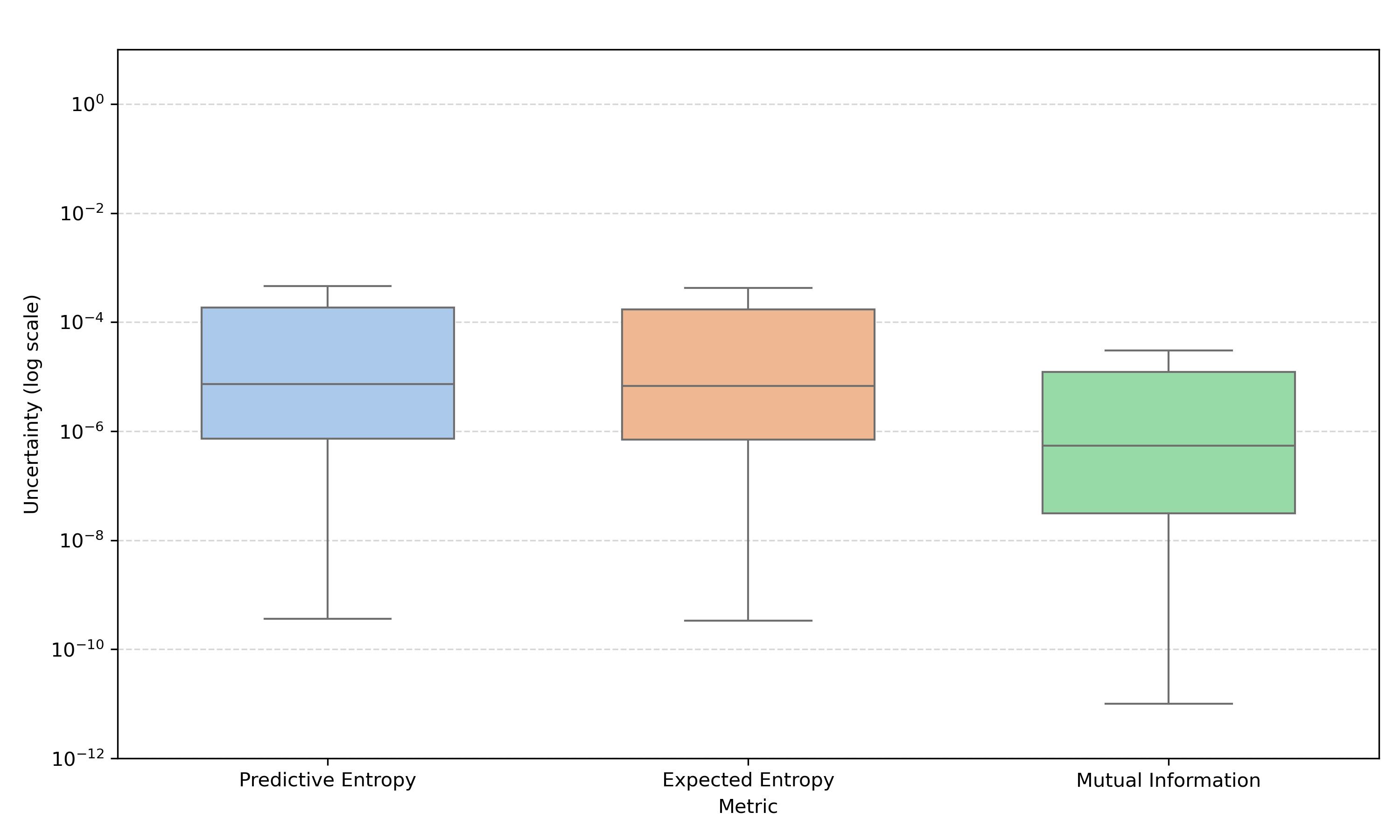}
        \includegraphics[width=0.45\linewidth]{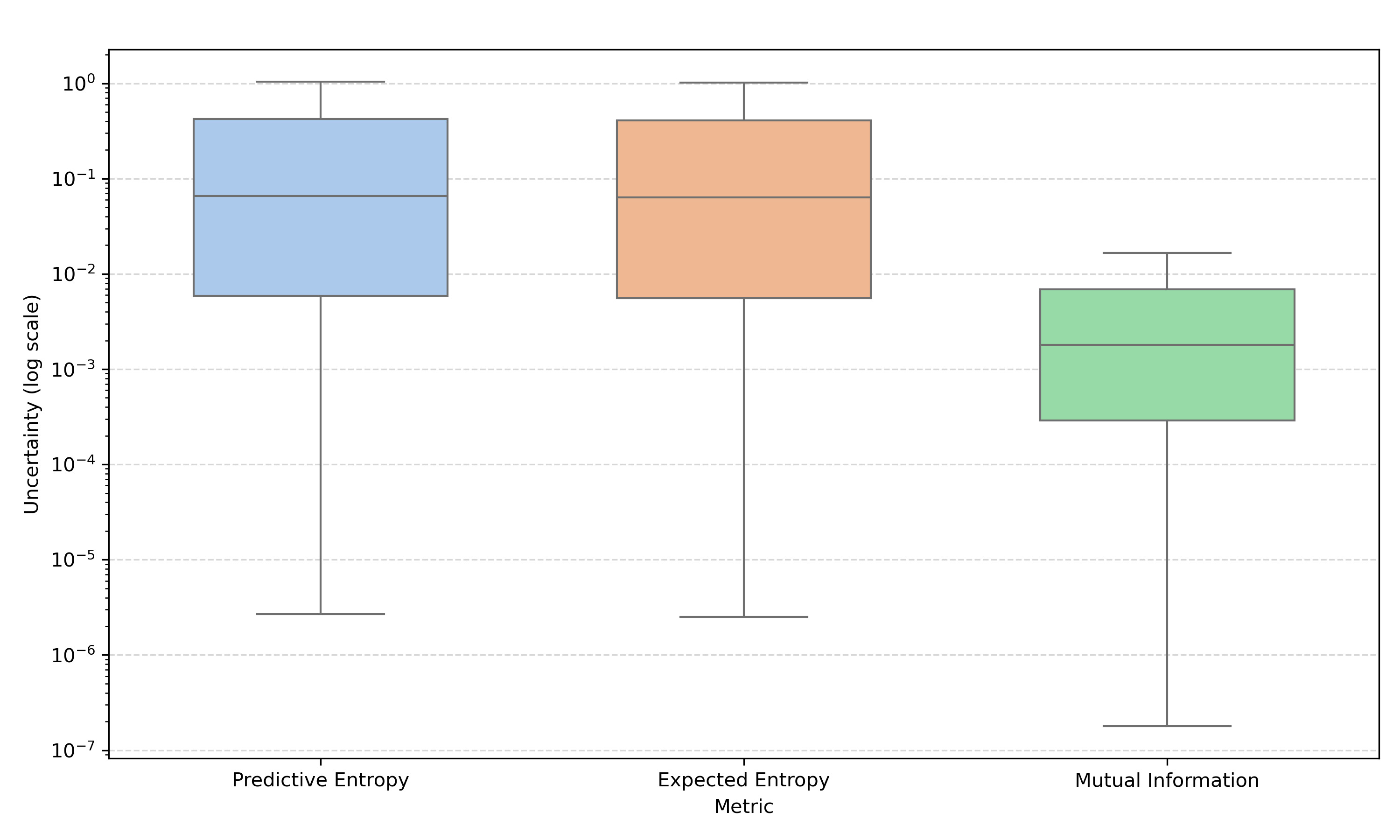}
    \caption{Uncertainty Decomposition for H-CNN VGG16 \& GoogLe Net} 
    \label{fig:Figure.11}
  \end{figure}

In the H-CNN VGG16 plot (Figure ~\ref{fig:Figure.11} (left)), we see that both predictive entropy and expected entropy exhibit similar distributions, while the mutual information remains noticeably lower. This suggests that most of the model’s uncertainty is aleatoric, the predictions are relatively stable across dropout passes, but the model still expresses uncertainty when the input is ambiguous. The narrow distribution of mutual information indicates limited epistemic uncertainty and is overall confident in its predictions, reinforcing earlier observations that H-CNN VGG16 becomes confident rapidly and shows minimal variation across passes.

GoogLeNet (Figure ~\ref{fig:Figure.11} (right)) displays a distinct pattern compared to H-CNN VGG16. Both predictive and expected entropy values are higher, and mutual information is substantially larger. This suggests that GoogLeNet expresses greater epistemic uncertainty; the model’s predictions vary more across dropout passes, indicating increased uncertainty in the model parameters. At the same time, the elevated expected entropy shows that the model also captures data-related ambiguity.  The wider spread of all three metrics, especially mutual information, indicates that GoogLeNet is more expressive in signalling when unsure, which aligns with its more gradual confidence rise and wider uncertainty bands in earlier plots.


As a result, H-CNN VGG16 tends to rely heavily on its learned decision boundaries and rarely adjusts its predictions, even in cases where uncertainty would be warranted—an indication of overconfidence. GoogLeNet, on the other hand, exhibits more flexible behaviour, capturing both model and data uncertainty more clearly. This  supports the view that GoogLeNet is better calibrated and more reliable in representing meaningful uncertainty.

%

\begin{figure}[H]
    \centering
    \includegraphics[width=0.45\textwidth]{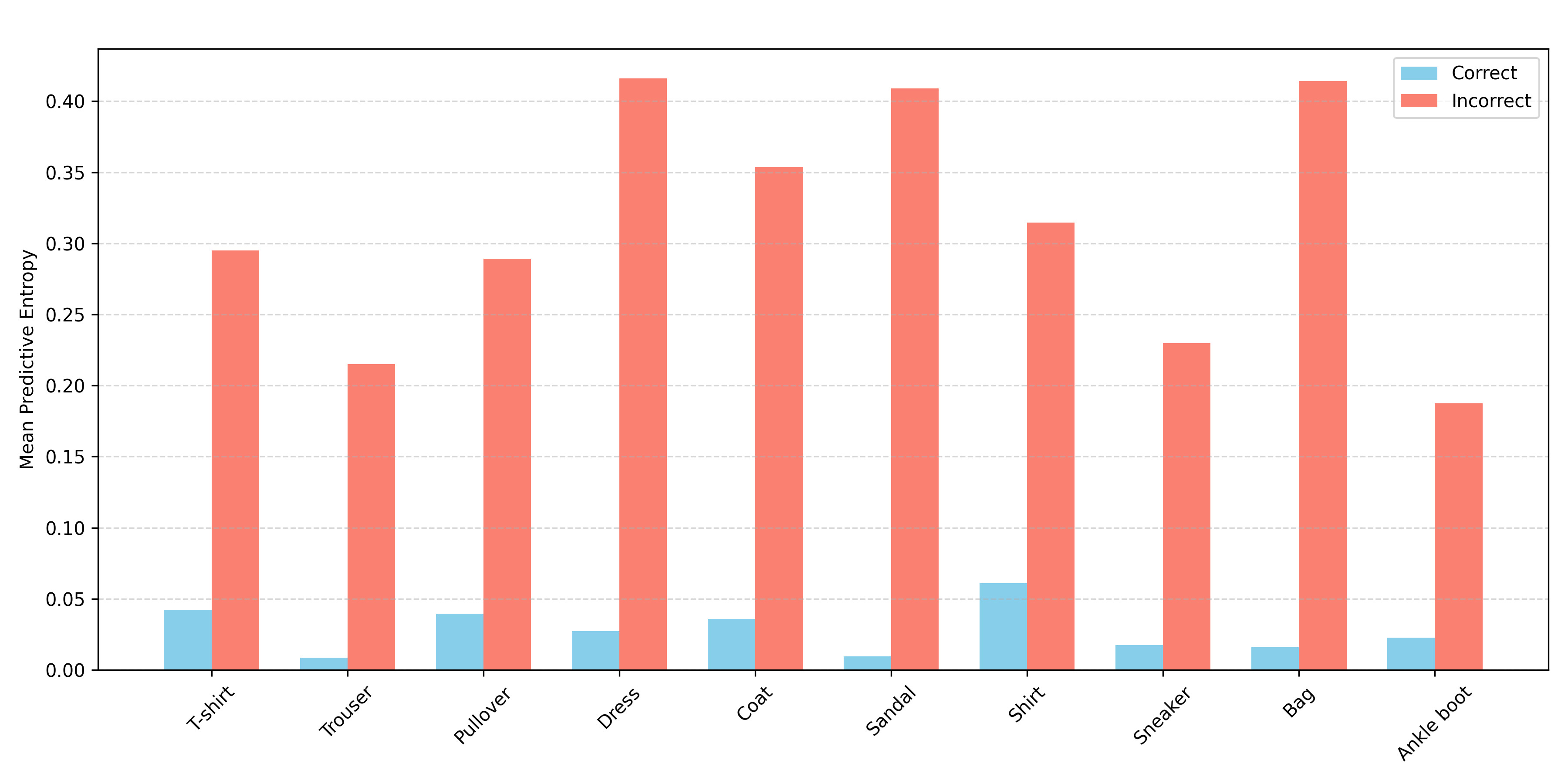}
        \includegraphics[width=0.45\textwidth]{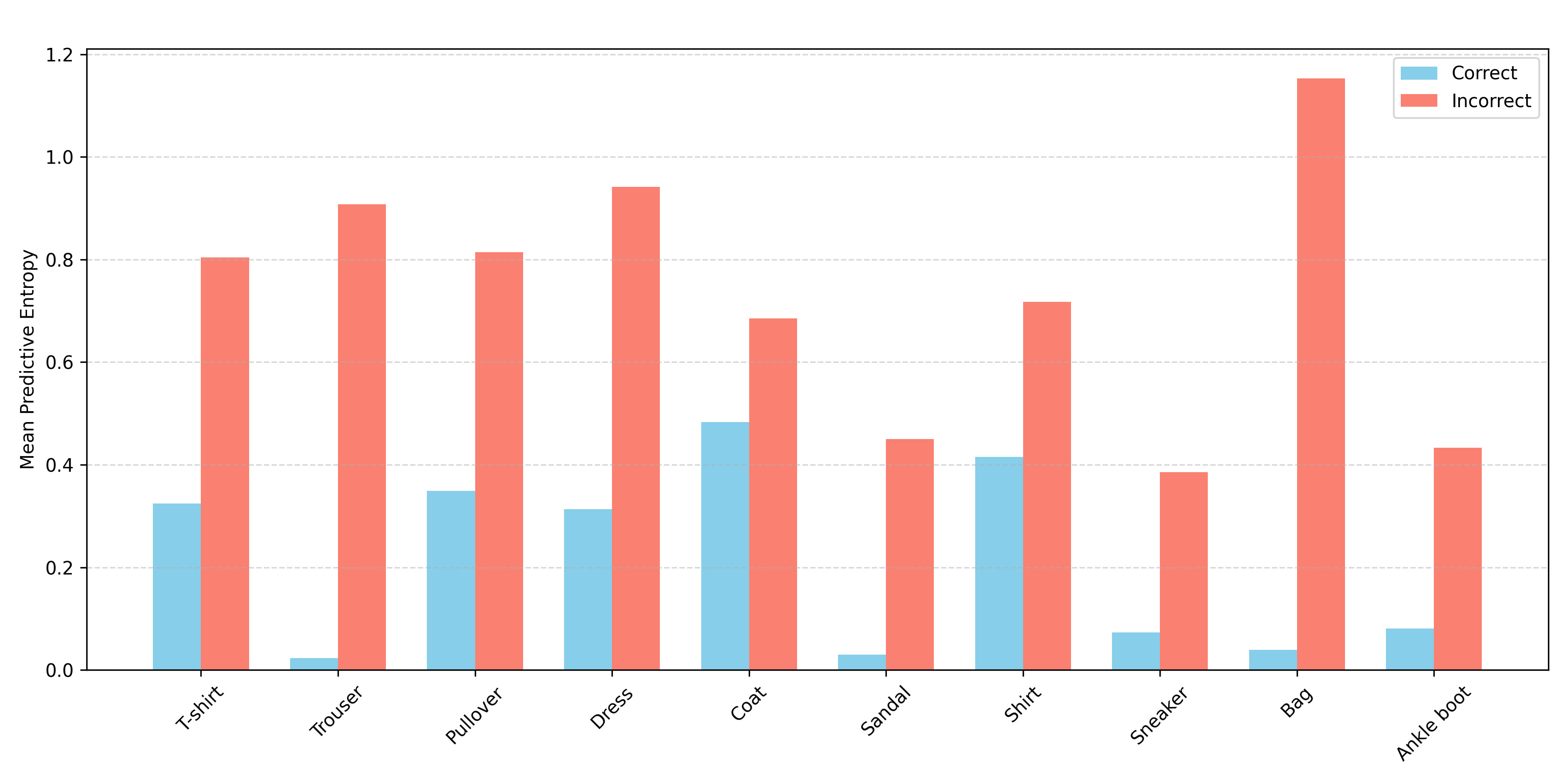}
    \caption{Entropy per Class for Correct vs. Incorrect Predictions for H-CNN VGG16 \& GoogLeNet}
    \label{fig:Figure.13}
\end{figure}

Previously, uncertainty was aggregated across multiple forward passes for each individual observation.  In this plot (Figure ~\ref{fig:Figure.13}), we group predictions by class and separate them into correct and incorrect cases. This approach allows us to see how confident the model is on average when it classifies correctly versus when it misclassifies. Ideally, a well-calibrated model should exhibit low entropy for correct classifications and higher entropy for errors, thereby signalling uncertainty appropriately.  Consequently, we aim to maximise the difference to reflect the model’s ability to distinguish between classes in a meaningful and reliable manner.

We see a familiar pattern in H-CNN VGG16 (Figure~\ref{fig:Figure.13} (left)): the model shows low entropy for correct predictions and high entropy for misclassified ones, with a clear separation between the two. At first glance, this is  desirable, as it suggests the model can express uncertainty when it errs. However, considering our earlier findings, this sharp separation may also be a sign of overfitting. The model  becomes highly confident very quickly, which may not always reflect genuine uncertainty, particularly for ambiguous inputs. This pattern is evident in classes like Pullover and Coat, which were previously identified as frequently misclassified. For Shirt, the worst-performing class according to the confusion matrix, the gap between correct and incorrect predictions is noticeably smaller, suggesting that the model does express higher uncertainty when less certain, consistent with behaviour expected from a well-calibrated model.

GoogLeNet shows a different behaviour though (see (Figure~\ref{fig:Figure.13} (right)). While the model generally assigns higher entropy to incorrect predictions than to correct ones, which is desirable, the separation between the two is less pronounced than in H-CNN VGG16. For certain classes, such as Shirt, Coat, T-shirt, and Dress, the entropy for correct predictions remains relatively high. This aligns with previous findings from the confusion matrix, which highlighted GoogLeNet’s difficulty in distinguishing among these visually similar categories. The model often expresses uncertainty even when it predicts correctly, indicating that its correct classifications may be closer to the decision boundary and at times verge on misclassification.

%
%

GoogLeNet shows smaller gaps between correct and incorrect entropy, and higher uncertainty even for correct predictions. Although H-CNN VGG16 is more confident and sharp in its separation, GoogLeNet appears to be more cautious and uncertain overall. This suggests that H-CNN VGG16 may still be overconfident, while GoogLeNet is more hesitant but potentially better calibrated.

\begin{figure}[H]
  \centering
    \includegraphics[width=0.45\textwidth]{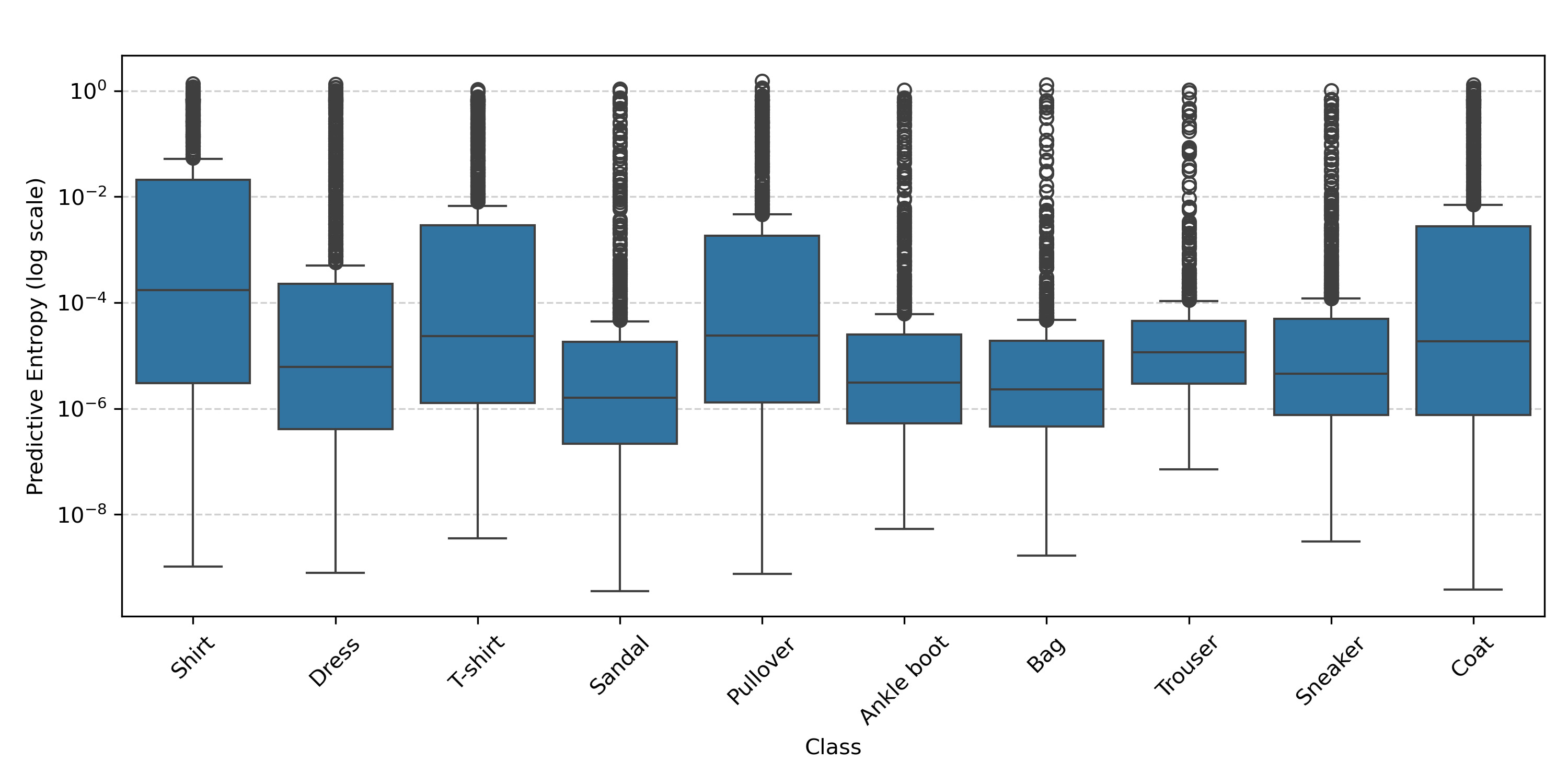}
        \includegraphics[width=0.45\textwidth]{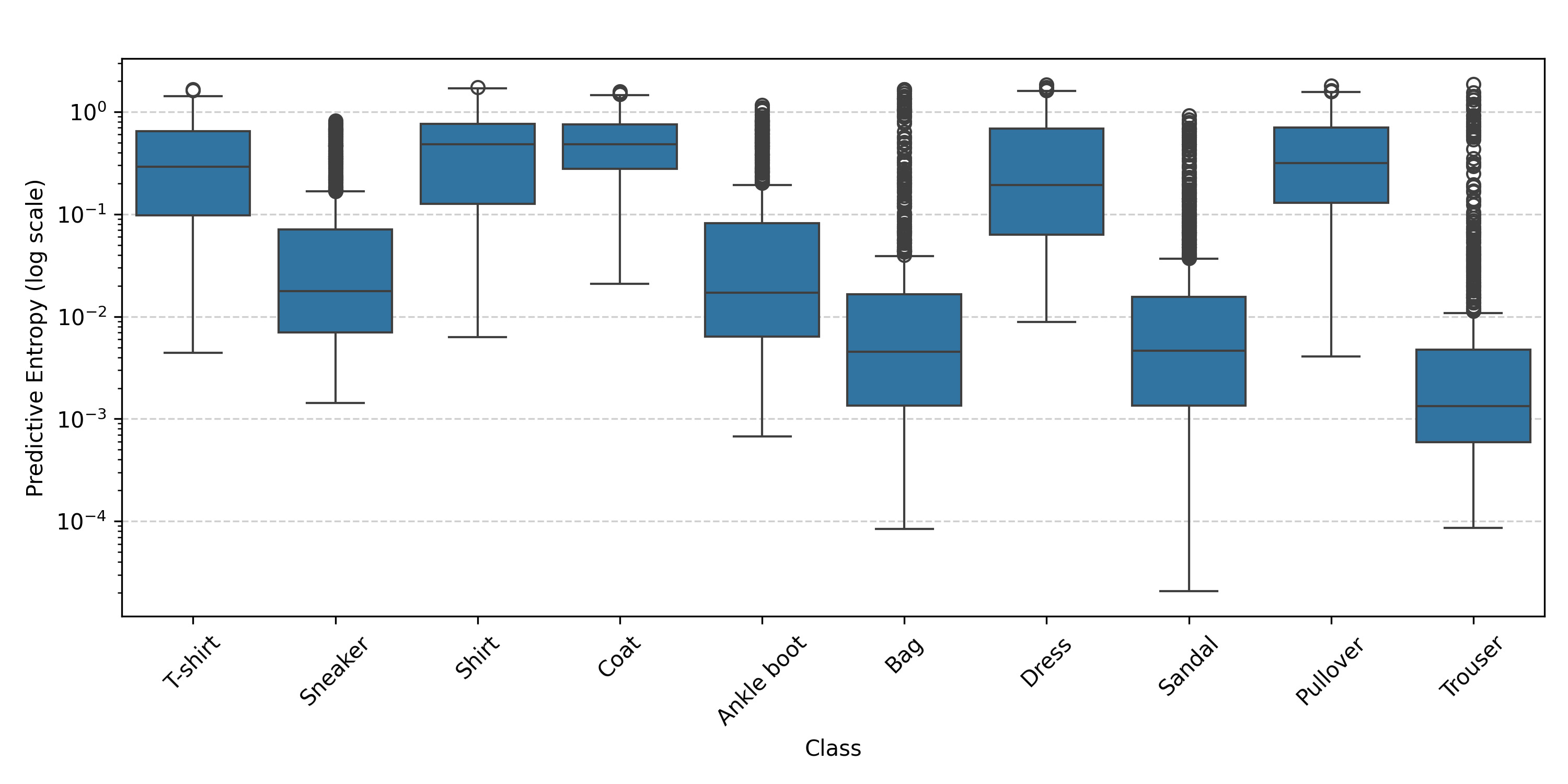}
    \caption{Predictive Entropy by Class for H-CNN VGG16 \& GoogLeNet}
    \label{fig:Figure.15}
\end{figure}

To better understand how uncertainty is expressed across the model, we examine predictive entropy and confidence at the class level rather than relying on overall averages. This approach allows us to identify whether certain classes are more challenging to predict and to evaluate whether the model expresses uncertainty in a manner consistent with their performance. The following analysis presents the distribution of predictive entropy values per class for both neural networks (Figure~\ref{fig:Figure.15}), using a logarithmic scale to emphasise variation across both straightforward and ambiguous classes. This approach also enables a direct link between uncertainty and the misclassification trends observed earlier, providing a more complete picture of model behaviour.


For H-CNN VGG16 (Figure~\ref{fig:Figure.15} (left)), predictive entropy remains low across most classes, with relatively tight distributions and few high-entropy outliers. This indicates that the model is generally confident in its predictions, regardless of class. However, slightly elevated entropy values appear for classes such as T-shirt, Coat, Pullover, and Shirt, among the most frequently misclassified classes in the confusion matrix. Nevertheless, the overall variation remains limited, indicating that the model’s confidence does not adjust substantially between easier and more difficult classes. This behaviour reinforces earlier indications of overfitting, as H-CNN VGG16 tends to remain confident even on ambiguous or borderline inputs.

For GoogLeNet (Figure~\ref{fig:Figure.15} (rigth)), predictive entropy remains consistently low across most classes, with relatively narrow distributions and only a few high-entropy outliers. This suggests that the model is generally confident in its predictions, regardless of class. Slightly elevated entropy values appear for categories such as T-shirt, Coat, Pullover, and Shirt, also among the most frequently misclassified in the confusion matrix. Nevertheless, the overall variation remains limited, indicating that the model’s confidence does not adjust substantially between easier and more difficult classes. This behaviour reinforces earlier indications of overfitting, as H-CNN VGG16 tends to remain confident even on ambiguous or borderline inputs.

In contrast, GoogLeNet shows a broader spread of predictive entropy across classes. Classes like Shirt, T-shirt, Coat, Pullover, and Dress exhibit notably higher median and upper-quartile entropy values, corresponding to the most frequently misclassified classes in the confusion matrix. This indicates that GoogLeNet is more sensitive to uncertainty in challenging cases and reflects that uncertainty more clearly in its predictions. Compared to H-CNN VGG16, GoogLeNet appears better calibrated and more capable of expressing doubt where appropriate, particularly for visually similar or ambiguous classes. Taken together, these findings highlight a trade-off; H-CNN VGG16 maintains efficiency through strong confidence, while GoogLeNet prioritises reliability by more explicitly expressing uncertainty.


To assess the robustness of our results, we examined the mean predicted class confidence together with its standard deviation, computed from fifty Monte Carlo Dropout forward passes across different classes. As these findings are consistent with the patterns discussed above, the corresponding plots are provided in the Appendix Figures~\ref{fig:43} and ~\ref{fig:44}. This placement avoids redundancy in the main text while still providing full detail for reference.

\subsubsection{Comparative Analysis: Conformal vs. Bayesian}
\label{sec:comparison}
This section examines how Conformal Prediction (CP) set sizes relate to predictive uncertainty  estimated via Monte Carlo Dropout. While both methods quantify uncertainty, they do so in fundamentally different ways. MC Dropout approximates a Bayesian posterior by performing multiple stochastic forward passes during inference, with predictive entropy capturing the dispersion of predicted probabilities across classes as a measure of uncertainty. In contrast, Conformal Prediction does not rely on entropy; instead, it ensures statistically valid coverage by calibrating prediction set sizes according to the softmax probability assigned to the true class, using a held-out calibration set.

This distinction has practical consequences. A model may exhibit high uncertainty (high entropy) while still assigning high confidence to the true label, resulting in a small CP set. Conversely, it may have low entropy but low true-class confidence, triggering CP to expand the prediction set. The correlation between entropy and set size therefore depends on how closely these two forms of uncertainty align in practice.

\begin{figure}[H]
    \centering
    \includegraphics[width=0.45\textwidth]{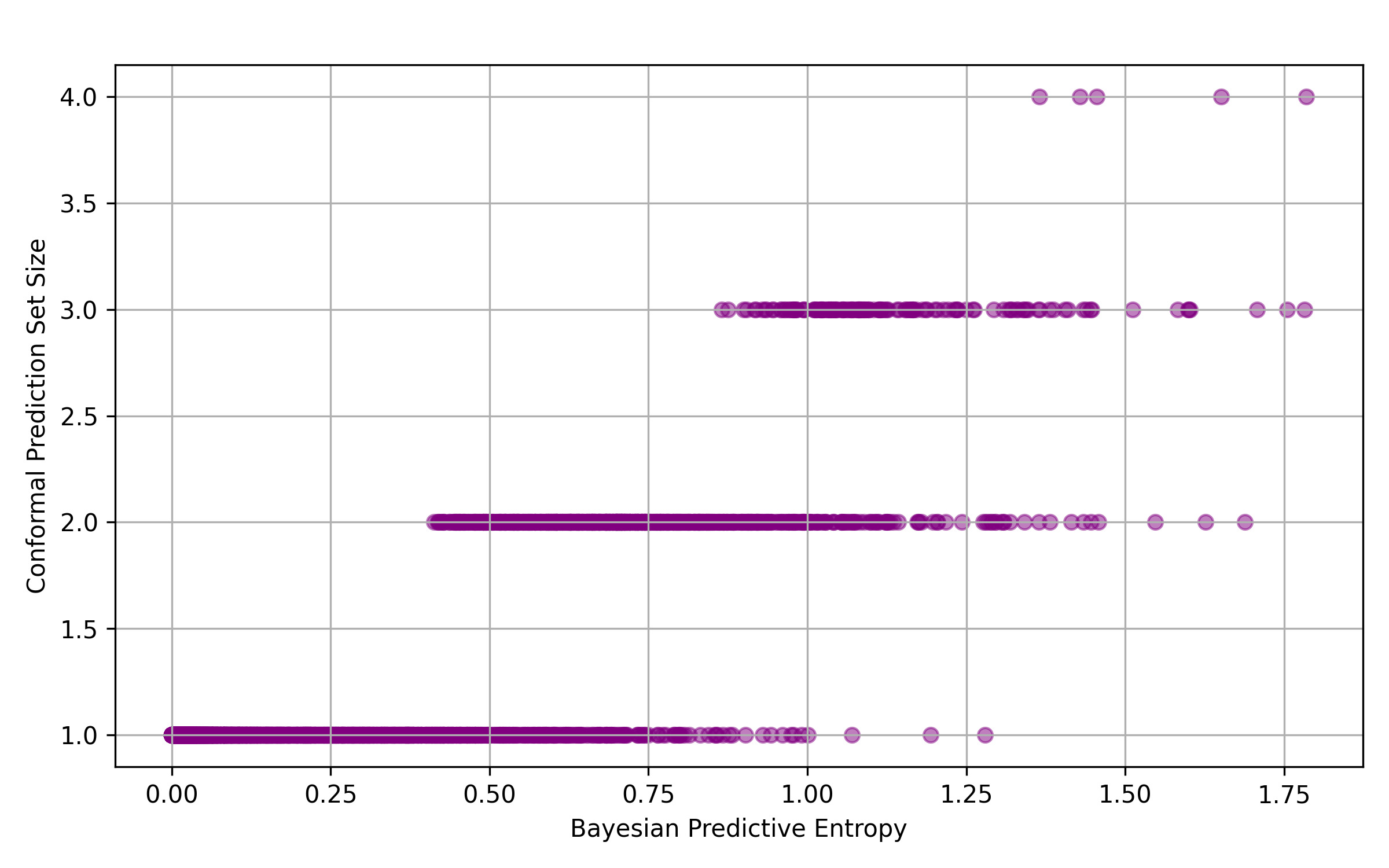}
        \includegraphics[width=0.45\textwidth]{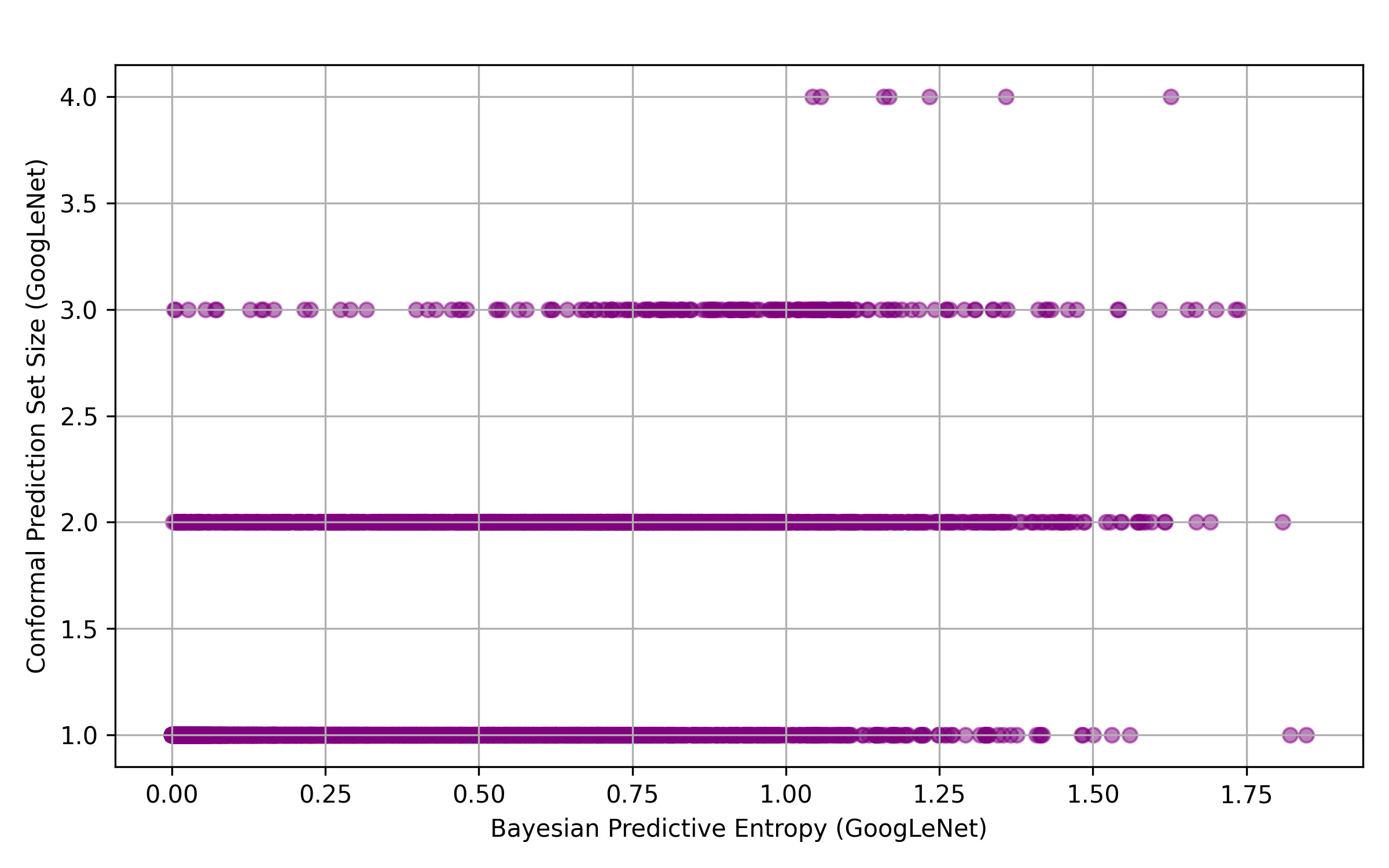}
    \caption{Prediction Set Vs. Predictive Entropy for H-CNN VGG16 \& GoogLeNet}
    \label{fig:Figure.17}
\end{figure}

In the case of H-CNN VGG16 (Figure~\ref{fig:Figure.17} (left)), the relationship between entropy and set size is strong and consistent. As predictive entropy increases, CP prediction sets also expand. The scatter plot shows a clear, monotonic pattern: predictions with low entropy typically correspond to set sizes of 1, whereas higher-entropy predictions more often require sets of size 2, 3, or even 4. This indicates that although H-CNN VGG16 tends to be overconfident overall (as shown in earlier plots), its entropy rankings still provide a reliable signal of prediction difficulty, allowing CP to adapt its set sizes effectively. In short, H-CNN VGG16's entropy may underestimate total uncertainty, but it is internally coherent and aligns well with CP calibration.

For GoogLeNet (Figure~\ref{fig:Figure.17} (right)), the relationship between entropy and set size is less clear. Although entropy generally increases for more difficult predictions, many low-entropy samples still result in large prediction sets. This behavior stems from the model’s cautious probability assignments; GoogLeNet tends to distribute its confidence more evenly across multiple plausible classes, even when its prediction is correct. As a result, the probability assigned to the true class can be relatively modest, not because the model is incorrect, but because it is better calibrated and avoids overconfidence. While this conservatism is advantageous from a reliability perspective, it often causes CP to enlarge prediction sets, thereby weakening the correlation between entropy and set size.


In conclusion, these findings highlight the complementary roles of MC Dropout and Conformal Prediction. MC Dropout captures the model’s internal uncertainty, while CP guarantees empirical coverage regardless of calibration quality. When predictive entropy aligns well with true-label confidence, as in H-CNN VGG16, the two methods work in sync. When this alignment is weaker, as in GoogLeNet, CP acts as a corrective mechanism to maintain reliability, even when entropy alone does not fully explain variation in set sizes. Ultimately, this comparison underscores a trade-off: H-CNN VGG16 favours efficiency through smaller prediction sets, whereas GoogLeNet prioritises reliability by more consistently signalling uncertainty.



\section{Conclusion}
\label{sec:conclusion}

This paper addressed the gap in the existing literature by comparing two fundamentally different approaches to uncertainty estimation in deep convolutional neural networks: Bayesian approximation via Monte Carlo Dropout and the nonparametric method of Conformal Prediction. The analysis was conducted on two distinct architectures, H-CNN VGG16 and GoogLeNet, applied consistently to the Fashion-MNIST dataset to ensure methodological comparability. Our empirical results provide several key insights. First, they clarify how uncertainty is expressed across models, distinguishing between epistemic and aleatoric components. Second, they reveal systematic patterns in class-level ambiguity, showing how models respond to visually similar categories. Finally, they demonstrate the complementary strengths of Bayesian and Conformal approaches: while Bayesian methods capture internal model uncertainty, Conformal Prediction guarantees empirical coverage and corrects for calibration weaknesses. Taken together, these findings advance our understanding of predictive reliability in deep learning and underscore the importance of designing models that not only achieve high accuracy but also convey trustworthy measures of uncertainty.


\bibliographystyle{elsarticle-harv}
\bibliography{references}

@article{barber2023conformal,
  title={Conformal prediction beyond exchangeability},
  author={Barber, Rina Foygel and Cand{\`e}s, Emmanuel J. and Ramdas, Aaditya and Tibshirani, Ryan J.},
  journal={The Annals of Statistics},
  volume={51},
  number={2},
  pages={816--845},
  year={2023},
doi={https://doi.org/10.1214/23-AOS2276}
}

@mastersthesis{fontes2023bayesian,
  author      = {Fontes, Wagner da Silva},
  year        = {2023},
  title       = {Bayesian inference for uncertainty quantification in binary classification tasks with limited data},
  type        = {Undergraduate thesis},
  school = {Universidade Federal de Alagoas},
  address     = {Macei{\'o}, Brazil}
}

@article{SON2026131927,
title = {Improving monte carlo dropout uncertainty estimation with stable output layers},
journal = {Neurocomputing},
volume = {661},
pages = {131927},
year = {2026},
issn = {0925-2312},
author = {Suhan Son and Junhee Seok},
}

@article{FERCHICHI2025130242,
title = {Deep Learning-based Uncertainty Quantification for spatio-temporal environmental Remote Sensing: A systematic literature review},
journal = {Neurocomputing},
volume = {639},
pages = {130242},
year = {2025},
issn = {0925-2312},
author = {Aya Ferchichi and Ahlem Ferchichi and Fatma Hendaoui and Mejda Chihaoui and Radhia Toujani},
}

@misc{shridhar2019bcnn,
      title={A comprehensive guide to {Bayesian} convolutional neural networks with variational inference}, 
      author={Kumar Shridhar and Felix Laumann and Marcus Liwicki},
      year={2019},
howpublished = {arXiv preprint arXiv:1901.02731},
  url          = {https://arxiv.org/abs/1901.02731}
}

@inproceedings{Essbai2024,
title={A framework for including uncertainty in robustness evaluation of {Bayesian} neural network nlassifiers},
  author={Essbai, Wasim and Bombarda, Andrea and Bonfanti, Silvia and Gargantini, Angelo},
  booktitle={Proceedings of the 5th IEEE/ACM International Workshop on Deep Learning for Testing and Testing for Deep Learning},
  pages={25--32},
  year={2024},
doi       = {https://doi.org/10.1145/3643786.3648026}

}

@article{Abdar2021,
author = {Moloud Abdar and Farhad Pourpanah and Sadiq Hussain and Dana Rezazadegan and Li Liu and Mohammad Ghavamzadeh and Paul Fieguth and Xiaochun Cao and Abbas Khosravi and U. Rajendra Acharya and Vladimir Makarenkov and Saeid Nahavandi},
title = {A review of uncertainty quantification in deep learning: Techniques, applications and challenges},
journal = {Information Fusion},
volume = {76},
pages = {243-297},
year = {2021},
doi = {https://doi.org/10.1016/j.inffus.2021.05.008}
}

@article{Gawlikowski2023,
   author = {Jakob Gawlikowski and Cedrique Rovile Njieutcheu Tassi and Mohsin Ali and Jongseok Lee and Matthias Humt and Jianxiang Feng and Anna Kruspe and Rudolph Triebel and Peter Jung and Ribana Roscher and Muhammad Shahzad and Wen Yang and Richard Bamler and Xiao Xiang Zhu},
   doi = {https://doi.org/10.1007/s10462-023-10562-9},
   journal = {Artificial Intelligence Review},
   title = {A survey of uncertainty in deep neural networks},
   volume = {56},
   year = {2023},
pages = {1513--1589}
}

@article{Shafer2008,
   author = {Glenn Shafer and Vladimir Vovk},
   journal = {Journal of Machine Learning Research},
   title = {A tutorial on conformal prediction},
   volume = {9},
   year = {2008},
pages = {371--421}
}

@book{vovk2005algorithmic,
  author       = {Vladimir Vovk and Alexander Gammerman and Glenn Shafer},
  title        = {Algorithmic learning in a random world},
  publisher    = {Springer},
  year         = {2005},
  address      = {New York, NY},
  doi          = {https://doi.org/10.1007/b106715},
  pages        = {324},
}

@article{Padarian2022,
  author  = {J. Padarian and B. Minasny and A. B. McBratney},
  title   = {Assessing the uncertainty of deep learning soil spectral models using Monte Carlo dropout},
  journal = {Geoderma},
  volume  = {425},
  pages   = {116063},
  year    = {2022},
  doi     = {https://doi.org/10.1016/j.geoderma.2022.116063}
}

@article{Chandra2021,
  author  = {Rohitash Chandra and Yixuan He},
  title   = {Bayesian neural networks for stock price forecasting before and during {COVID-19} pandemic},
  journal = {PLOS ONE},
  volume  = {16},
  number  = {7},
  pages   = {e0253217},
  year    = {2021},
  doi     = {https://doi.org/10.1371/journal.pone.0253217}
}

@misc{sun2022conformalmethodsquantifyinguncertainty,
  author       = {Sophia Sun},
  title        = {Conformal methods for quantifying uncertainty in spatiotemporal data: A survey},
  year         = {2022},
  howpublished = {arXiv preprint arXiv:2209.03580},
  url          = {https://arxiv.org/abs/2209.03580}
}

@misc{liang2024conformalpredictionquantifyinguncertainty,
  author       = {Aoming Liang and Qi Liu and Lei Xu and Fahad Sohrab and Weicheng Cui and Changhui Song and Moncef Gabbouj},
  title        = {Conformal prediction on quantifying uncertainty of dynamic systems},
  year         = {2024},
  howpublished = {arXiv preprint arXiv:2412.10459},
  url          = {https://arxiv.org/abs/2412.10459}
}

@article{zhou2025conformalpredictiondataperspective,
  author    = {Xiaofan Zhou and Baiting Chen and Yu Gui and Lu Cheng},
  title     = {Conformal prediction: A data perspective},
  journal   = {ACM Computing Surveys},
  year      = {2025},
  doi       = {https://doi.org/10.1145/3736575},
  note      = {In press}
}

@article{Fontana2023,
title     = {Conformal prediction: A unified review of theory and new challenges},
  author    = {Fontana, Matteo and Zeni, Gianluca and Vantini, Simone},
  journal   = {Bernoulli},
  volume    = {29},
  number    = {1},
  pages     = {1--23},
  year      = {2023},
  doi       = {https://doi.org/10.3150/21-BEJ1447}
}

@misc{bbouzidi2024convolutionalneuralnetworksvision,
  author       = {Sonia Bbouzidi and Ghazala Hcini and Imen Jdey and Fadoua Drira},
  title        = {Convolutional neural networks and vision transformers for {Fashion-MNIST} classification: A literature review},
  year         = {2024},
  howpublished = {arXiv preprint arXiv:2406.03478},
  url          = {https://arxiv.org/abs/2406.03478}
}

@article{Krichen2023,
  author  = {Moez Krichen},
  title   = {Convolutional neural networks: A survey},
  journal = {Computers},
  volume  = {12},
  number  = {8},
  pages   = {151},
  year    = {2023},
  doi     = {10.3390/computers12080151}
}

@inproceedings{Gal2017,
  author    = {Yarin Gal and Riashat Islam and Zoubin Ghahramani},
  title     = {Deep {Bayesian} active learning with image data},
  booktitle = {Proceedings of the 34th International Conference on Machine Learning (ICML 2017)},
  pages     = {1183--1192},
  year      = {2017},
  publisher = {JMLR.org},
  address   = {Sydney, NSW, Australia}
}

@inproceedings{Nguyen2015,
  author    = {Anh Nguyen and Jason Yosinski and Jeff Clune},
  title     = {Deep neural networks are easily fooled: High confidence predictions for unrecognizable images},
  booktitle = {2015 IEEE Conference on Computer Vision and Pattern Recognition (CVPR)},
  year      = {2015},
  pages     = {427--436},
  doi       = {10.1109/CVPR.2015.7298640}
}

@inproceedings{Khurjekar2023,
  author    = {Ishan Khurjekar and Peter Gerstoft},
  title     = {DoA uncertainty quantification with conformal prediction},
  booktitle = {2023 IEEE 33rd International Workshop on Machine Learning for Signal Processing (MLSP)},
  year      = {2023},
  pages     = {1--6},
  publisher = {IEEE},
  doi       = {10.1109/MLSP55844.2023.10285924}
}

@inproceedings{Gal2016,
  author    = {Yarin Gal and Zoubin Ghahramani},
  title     = {Dropout as a Bayesian approximation: Representing model uncertainty in deep learning},
  booktitle = {Proceedings of the 33rd International Conference on Machine Learning},
  series    = {Proceedings of Machine Learning Research},
  volume    = {48},
  pages     = {1050--1059},
  year      = {2016},
  address   = {New York, New York, USA},
  publisher = {PMLR}
}

@inproceedings{Janjua2023,
  author    = {Juhi Janjua and Archana Patankar and Aradhya Talan},
  title     = {Exploring pretrained models and transfer learning techniques for image retrieval},
  booktitle = {Proceedings of the 2023 14th International Conference on Computing Communication and Networking Technologies (ICCCNT)},
  year      = {2023},
  pages     = {1--7},
  publisher = {IEEE},
  doi       = {10.1109/ICCCNT56998.2023.10307272}
}

@misc{lu2022fairconformalpredictorsapplications,
  author       = {Charles Lu and Andreanne Lemay and Ken Chang and Katharina Hoebel and Jayashree Kalpathy-Cramer},
  title        = {Fair conformal predictors for applications in medical imaging},
  year         = {2022},
  howpublished = {arXiv preprint arXiv:2109.04392},
  url          = {https://arxiv.org/abs/2109.04392}
}

@inproceedings{Samia2022,
  author    = {Samia Bougareche and Soraya Zehani and Malika Mimi},
  title     = {Fashion images classification using machine learning, deep learning and transfer learning models},
  booktitle = {Proceedings of the 7th International Conference on Image and Signal Processing and their Applications (ISPA 2022)},
  year      = {2022},
  pages     = {1--5},
  doi       = {10.1109/ISPA54004.2022.9786364}
}

@misc{xiao2017fashionmnistnovelimagedataset,
  author       = {Han Xiao and Kashif Rasul and Roland Vollgraf},
  title        = {Fashion-MNIST: A novel image dataset for benchmarking machine learning algorithms},
  year         = {2017},
  howpublished = {arXiv preprint arXiv:1708.07747},
  url          = {https://arxiv.org/abs/1708.07747}
}

@phdthesis{Poceviit2023,
  author    = {Milda Pocevičiūtė},
  title     = {Generalisation and reliability of deep learning for digital pathology in a clinical setting},
  school    = {Linköping University},
  address   = {Linköping},
  year      = {2023},
  type      = {PhD thesis},
  publisher = {Linköping University Electronic Press},
  doi       = {10.3384/9789180753005}
}

@phdthesis{kendall_2019,
  author    = {Alex Guy Kendall},
  title     = {Geometry and uncertainty in deep learning for computer vision},
  school    = {University of Cambridge},
  year      = {2019},
  type      = {PhD thesis},
  doi       = {10.17863/CAM.35260},
  url       = {https://www.repository.cam.ac.uk/handle/1810/287944}
}

@inproceedings{Szegedy2015,
  author = {Christian Szegedy and Wei Liu and Yangqing Jia and Pierre Sermanet and Scott Reed and Dragomir Anguelov and Dumitru Erhan and Vincent Vanhoucke and Andrew Rabinovich},
   doi = {10.1109/CVPR.2015.7298594},
   booktitle = {Proceedings of the IEEE Computer Society Conference on Computer Vision and Pattern Recognition},
   title = {Going deeper with convolutions},
pages={1--9},
   year = {2015}
}

@article{Jospin2022,
   author = {Laurent Valentin Jospin and Hamid Laga and Farid Boussaid and Wray Buntine and Mohammed Bennamoun},
   doi = {10.1109/MCI.2022.3155327},
   number = {2},
   journal = {IEEE Computational Intelligence Magazine},
   title = {Hands-on {Bayesian} neural networks: A tutorial for deep learning users},
   volume = {17},
pages={29--48},
   year = {2022}
}

@article{Seo2019,
   author = {Yian Seo and Kyung Shik Shin},
   doi = {10.1016/j.eswa.2018.09.022},
   journal = {Expert Systems with Applications},
   title = {Hierarchical convolutional neural networks for fashion image classification},
   volume = {116},
pages={328--339},
   year = {2019}
}

@inproceedings{Krizhevsky2012,
   author = {Alex Krizhevsky and Ilya Sutskever and Geoffrey E. Hinton},
   booktitle = {Advances in Neural Information Processing Systems},
   title = {ImageNet classification with deep convolutional neural networks},
   volume = {25},
pages     = {1097--1105},
   year = {2012}
}

@incollection{Papadopoulos08,
author = {Harris Papadopoulos},
title = {Inductive conformal prediction: Theory and application to neural networks},
booktitle = {Tools in Artificial Intelligence},
publisher = {IntechOpen},
address = {Rijeka},
year = {2008},
chapter = {18},
doi = {10.5772/6078},
}

@article{Matiz2019,
   author = {Sergio Matiz and Kenneth E. Barner},
   doi = {10.1016/j.patcog.2019.01.035},
   journal = {Pattern Recognition},
   title = {Inductive conformal predictor for convolutional neural networks: Applications to active learning for image classification},
   volume = {90},
pages={172--182},
   year = {2019}
}

@inproceedings{gupta2024complexityrequiredneuralnetwork,
  author    = {Manas Gupta and Efe Camci and Vishandi Rudy Keneta and Abhishek Vaidyanathan and Ritwik Kanodia and Ashish James and Chuan-Sheng Foo and Min Wu and Jie Lin},
  title     = {Is complexity required for neural network pruning? A case study on global magnitude pruning},
  booktitle = {Proceedings of the 2024 IEEE Conference on Artificial Intelligence (CAI)},
  year      = {2024},
  pages     = {747--754},
  publisher = {IEEE Computer Society},
  address   = {Los Alamitos, CA, USA},
  doi       = {10.1109/CAI59869.2024.00144}
}

@book{Lindholm2021,
  author    = {Andreas Lindholm and Niklas Wahlström and Fredrik Lindsten and Thomas B. Schön},
  title     = {Machine learning: A first course for engineers and scientists},
  publisher = {Cambridge University Press},
  address   = {Cambridge},
  year      = {2021}
}

@inproceedings{Naeini2015,
  author    = {Mahdi Pakdaman Naeini and Gregory F. Cooper and Milos Hauskrecht},
  title     = {Obtaining well calibrated probabilities using Bayesian binning},
  booktitle = {Proceedings of the AAAI Conference on Artificial Intelligence},
  year      = {2015},
  volume    = {29},
  number    = {1},
  pages     = {2901--2907},
  doi       = {10.1609/aaai.v29i1.9602}
}

@inproceedings{Guo2017,
  author    = {Chuan Guo and Geoff Pleiss and Yu Sun and Kilian Q. Weinberger},
  title     = {On calibration of modern neural networks},
  booktitle = {Proceedings of the 34th International Conference on Machine Learning (ICML 2017)},
  pages     = {1321--1330},
  year      = {2017},
  publisher = {JMLR.org},
  address   = {Sydney, Australia}
}

@article{Roth2024,
   author = {Jannik P. Roth and Jürgen Bajorath},
   doi = {10.1038/s41598-024-57135-6},
   number = {1},
   journal = {Scientific Reports},
   pages = {6536},
   title = {Relationship between prediction accuracy and uncertainty in compound potency prediction using deep neural networks and control models},
   volume = {14},
   year = {2024}
}

@article{Alzubaidi2021,
   author = {Laith Alzubaidi and Jinglan Zhang and Amjad J. Humaidi and Ayad Al-Dujaili and Ye Duan and Omran Al-Shamma and J. Santamaría and Mohammed A. Fadhel and Muthana Al-Amidie and Laith Farhan},
   doi = {https://doi.org/10.1186/s40537-021-00444-8},
   number = {1},
   journal = {Journal of Big Data},
   pages = {53},
   title = {Review of deep learning: Concepts, {CNN} architectures, challenges, applications, future directions},
   volume = {8},
   year = {2021}
}

@article{Vives-Boix2021,
   author = {Víctor Vives-Boix and Daniel Ruiz-Fernández},
   doi = {10.1016/j.neucom.2021.08.021},
   journal = {Neurocomputing},
   title = {Synaptic metaplasticity for image processing enhancement in convolutional neural networks},
pages = {534--543},
   volume = {462},
   year = {2021}
}

@inproceedings{Eaton-Rosen2018,
  author    = {Zach Eaton-Rosen and Felix Bragman and Sotirios Bisdas and Sébastien Ourselin and M. Jorge Cardoso},
  title     = {Towards safe deep learning: Accurately quantifying biomarker uncertainty in neural network predictions},
  booktitle = {Medical Image Computing and Computer Assisted Intervention -- MICCAI 2018},
  editor    = {Alejandro F. Frangi and Julia A. Schnabel and Christos Davatzikos and Carlos Alberola-López and Gabor Fichtinger},
  publisher = {Springer International Publishing},
  address   = {Cham},
  year      = {2018},
  pages     = {691--699},
  doi       = {10.1007/978-3-030-00928-1-78}
}

@misc{angelopoulos2022gentleintroductionconformalprediction,
  title        = {A gentle introduction to conformal prediction and distribution-free uncertainty quantification},
  author       = {Anastasios N. Angelopoulos and Stephen Bates},
  year         = {2022},
  howpublished = {arXiv preprint arXiv:2107.07511},
  url          = {https://arxiv.org/abs/2107.07511}
}

@misc{fan2024utopiauniversallytrainableoptimal,
  title        = {UTOPIA: Universally trainable optimal prediction intervals aggregation},
  author       = {Jianqing Fan and Jiawei Ge and Debarghya Mukherjee},
  year         = {2024},
  howpublished = {arXiv preprint arXiv:2306.16549},
  url          = {https://arxiv.org/abs/2306.16549}
}

@inproceedings{Simonyan2015,
   author = {Karen Simonyan and Andrew Zisserman},
   booktitle = {3rd International Conference on Learning Representations (ICLR 2015)},
pages = {1--14},
   title = {Very deep convolutional networks for large-scale image recognition},
   year = {2015}
}

@article{KendallGal2017,
   author = {Alex Kendall and Yarin Gal},
   journal = {Advances in Neural Information Processing Systems},
   title = {What uncertainties do we need in {Bayesian} deep learning for computer vision?},
   volume = {30},
   year = {2017},
pages = {5580--5590}
}

@article{9745083,
  author={Abdullah, Abdullah A. and Hassan, Masoud M. and Mustafa, Yaseen T.},
  journal={IEEE Access}, 
  title={A review on {Bayesian} deep learning in healthcare: Applications and challenges}, 
  year={2022},
  volume={10},
  pages={36538-36562},
  doi     = {https://doi.org/10.1109/ACCESS.2022.3163384}
}

@article{shiman_li,
title = {Evaluation of uncertainty estimation methods in medical image segmentation: Exploring the usage of uncertainty in clinical deployment},
journal = {Computerized Medical Imaging and Graphics},
volume = {124},
pages = {102574},
year = {2025},
doi = {https://doi.org/10.1016/j.compmedimag.2025.102574},
author = {Shiman Li and Mingzhi Yuan and Xiaokun Dai and Chenxi Zhang}}


\newpage
\section{Appendix}

\begin{figure}[tbh!]
    \centering
    \includegraphics[width=0.45\textwidth]{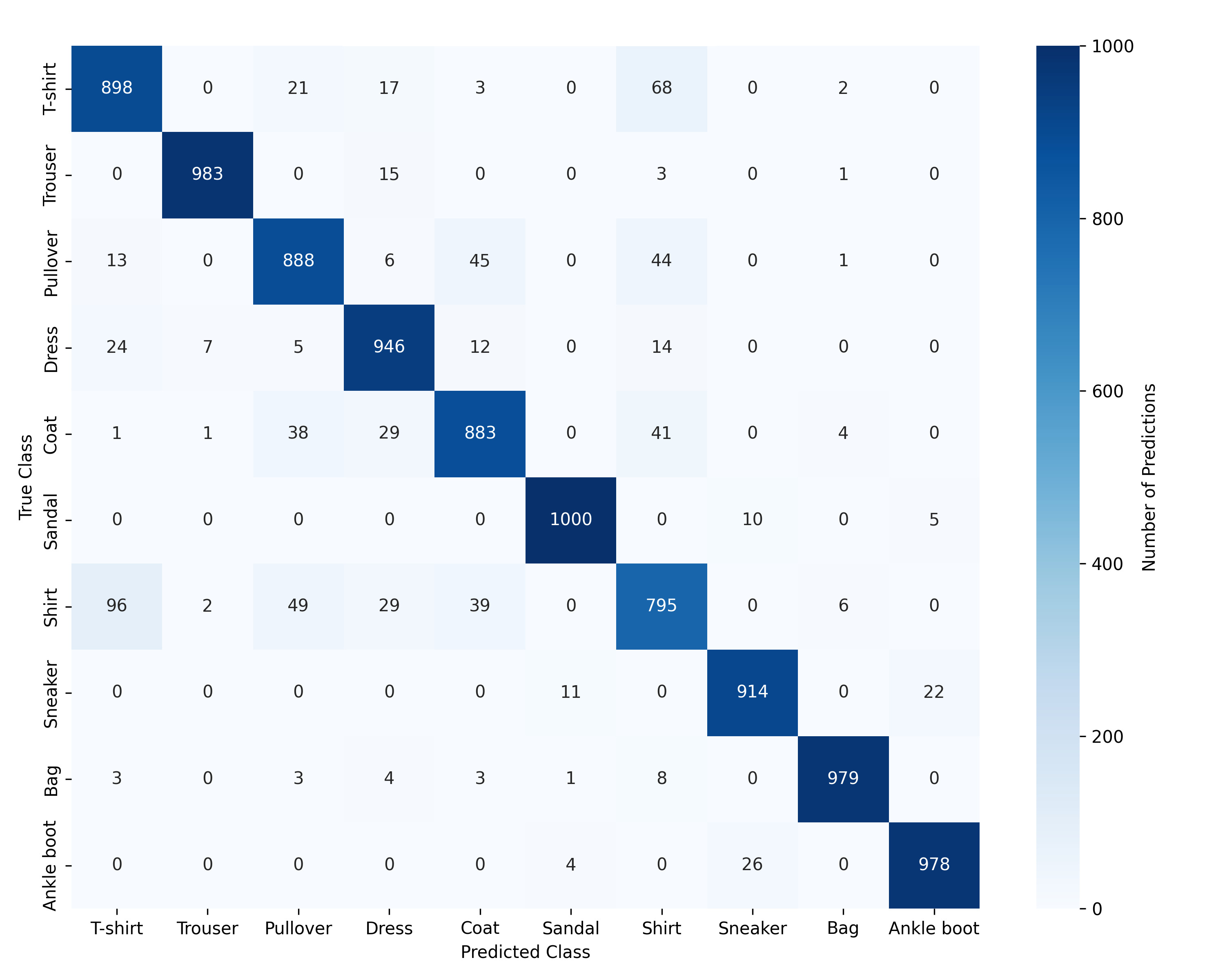}
        \includegraphics[width=0.45\textwidth]{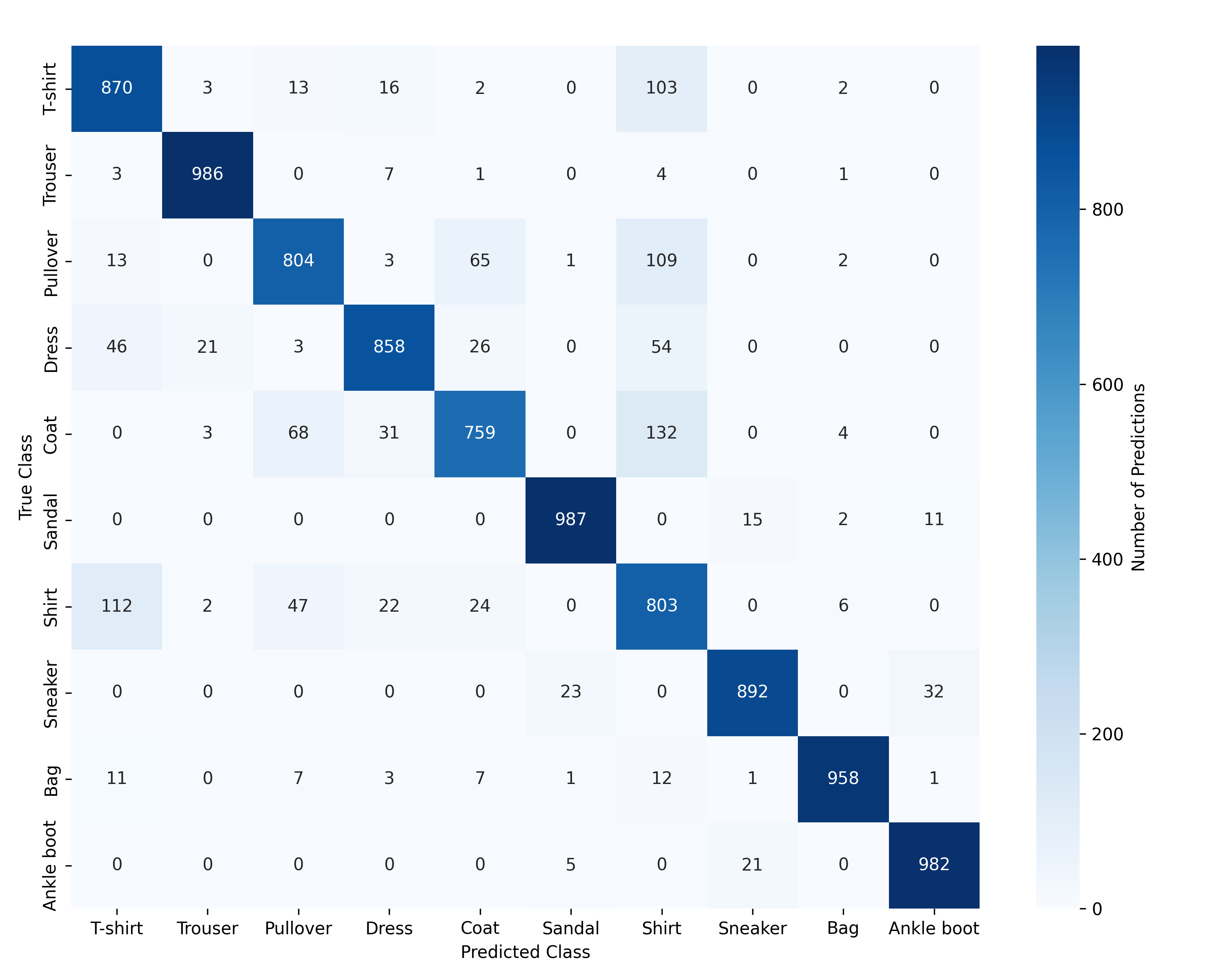}
    \caption{Heatmap for H-CNN VGG16 \& GoogLeNet Bayesian}
    \label{fig:19}
\end{figure}


\begin{figure}
    \centering
    \includegraphics[width=0.6\textwidth]{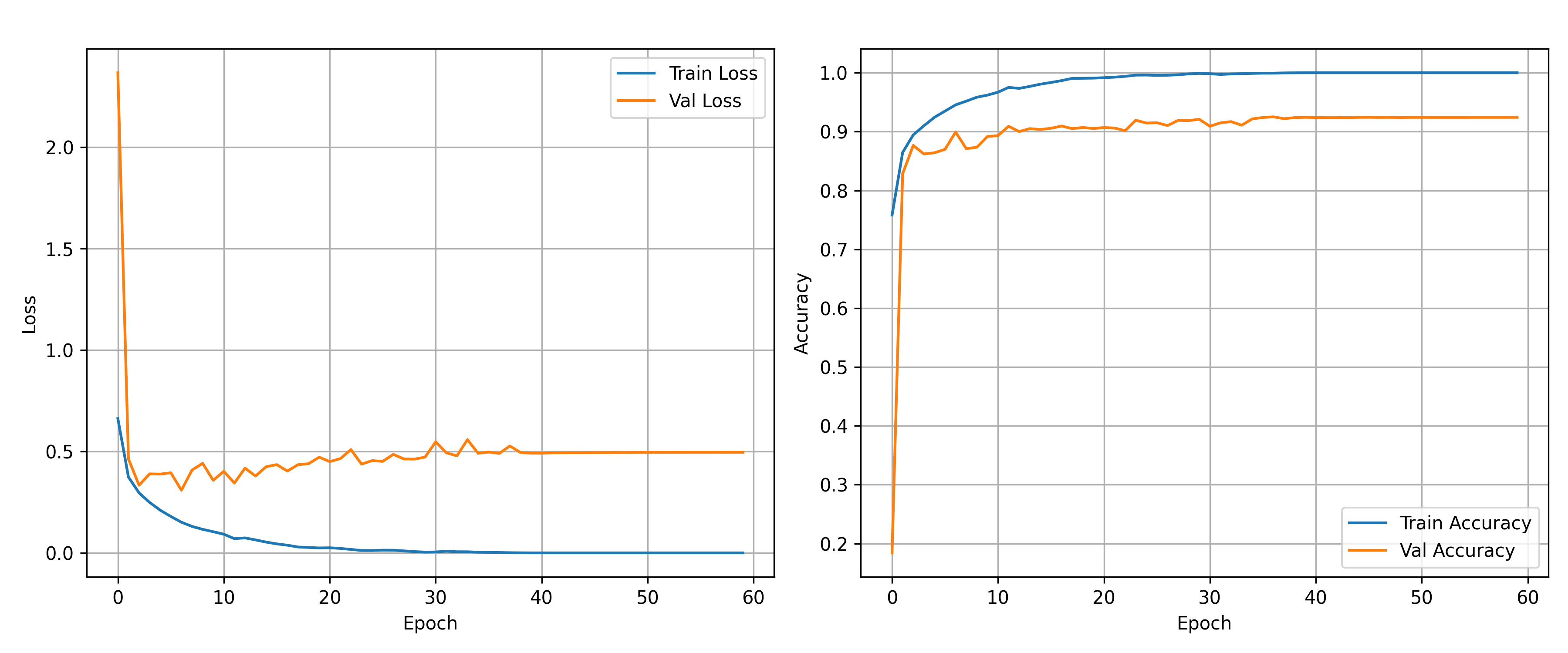}
        \includegraphics[width=0.6\textwidth]{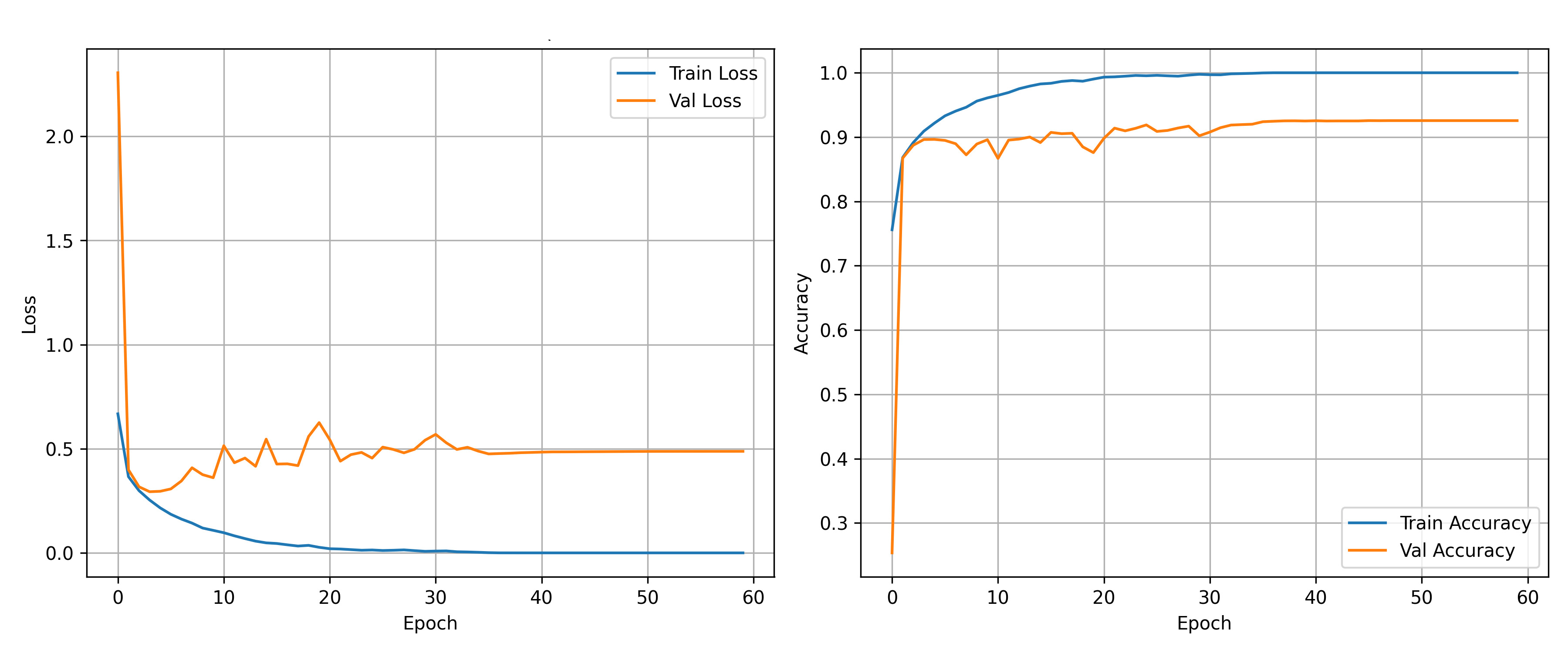}
            \includegraphics[width=0.6\textwidth]{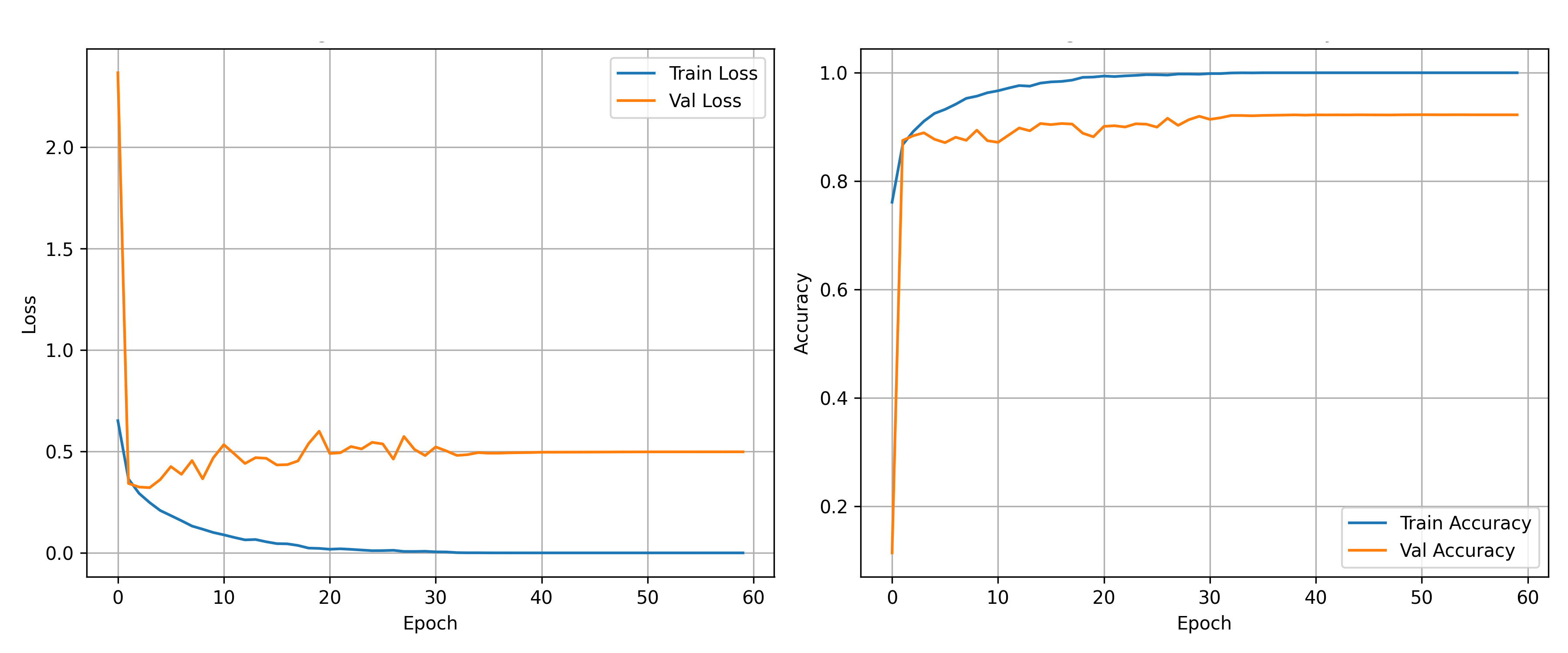}
                \includegraphics[width=0.6\textwidth]{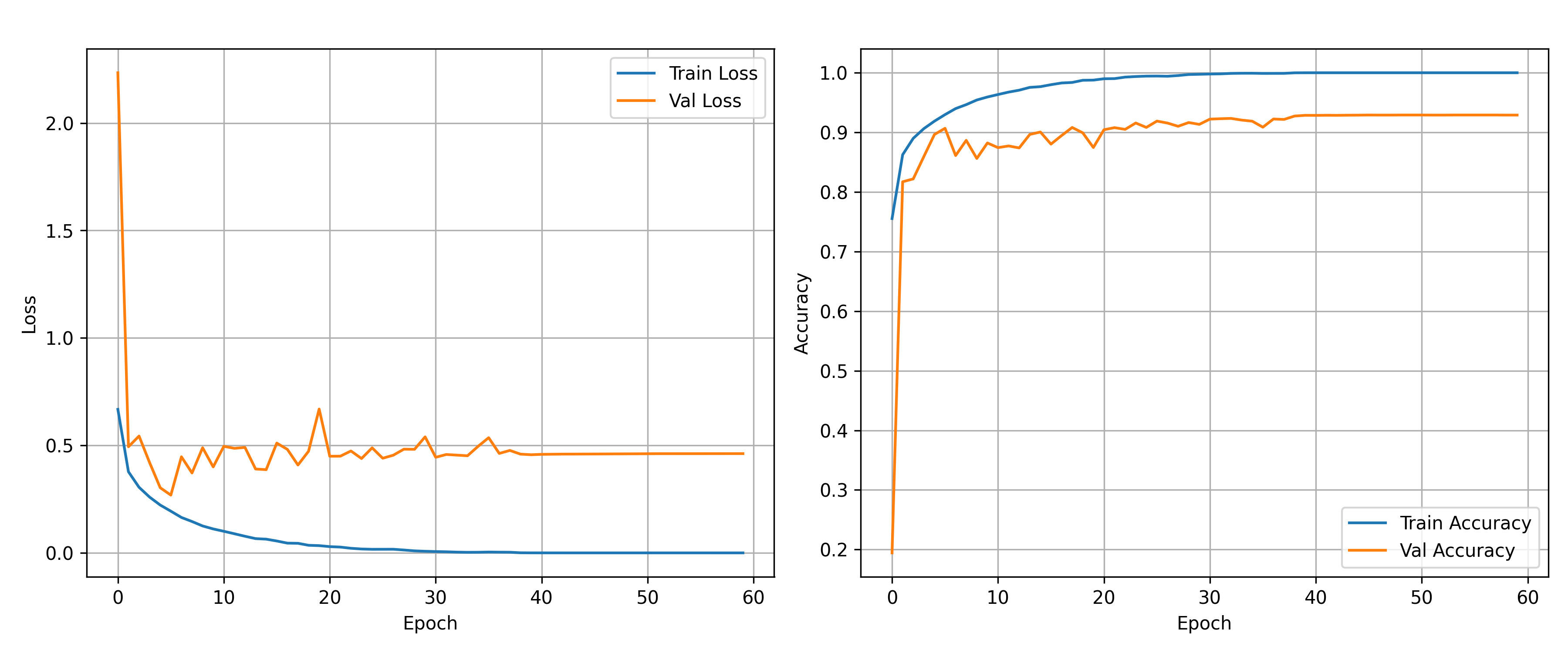}
    \caption{Accuracy Vs. Loss  H-CNN VGG16 (Fold 1, 3-5)}
    \label{fig:21}
\end{figure}

%
%

\begin{figure}
    \centering
    \includegraphics[width=0.6\textwidth]{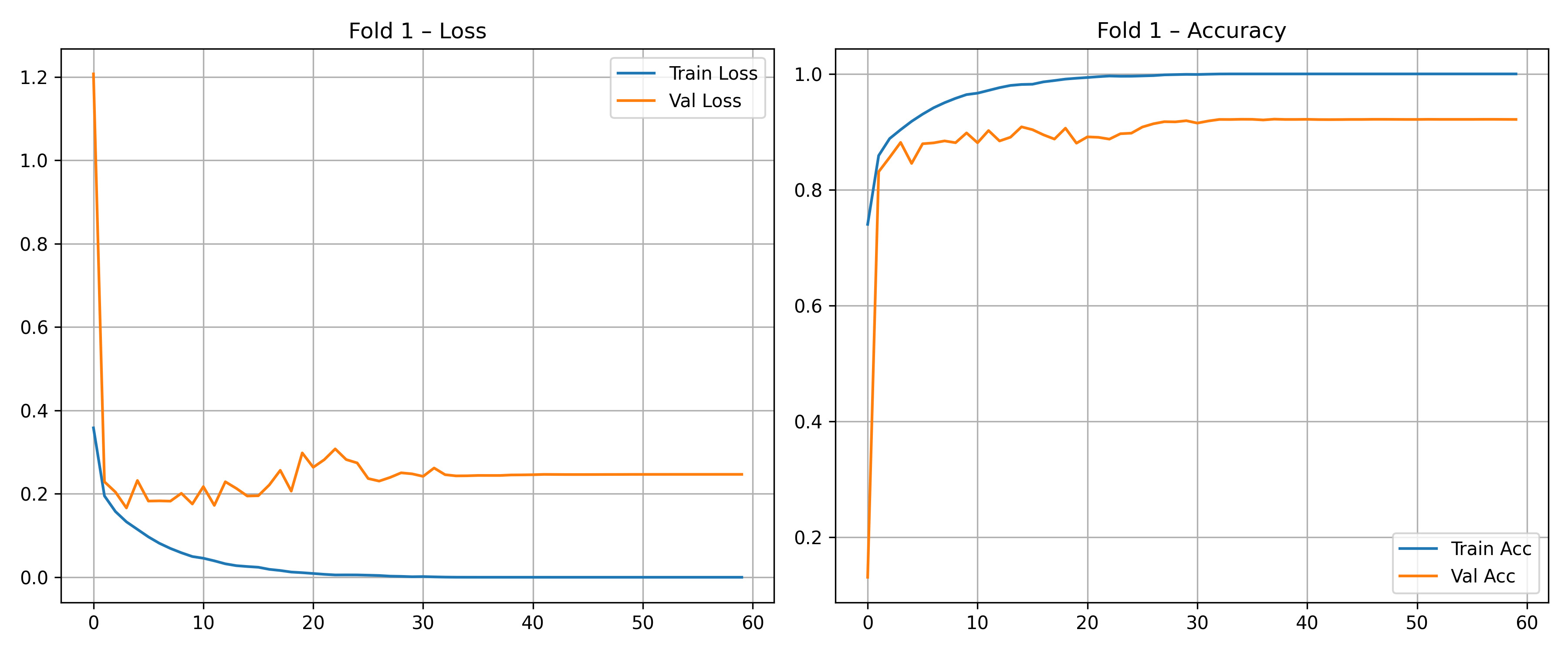}
        \includegraphics[width=0.6\textwidth]{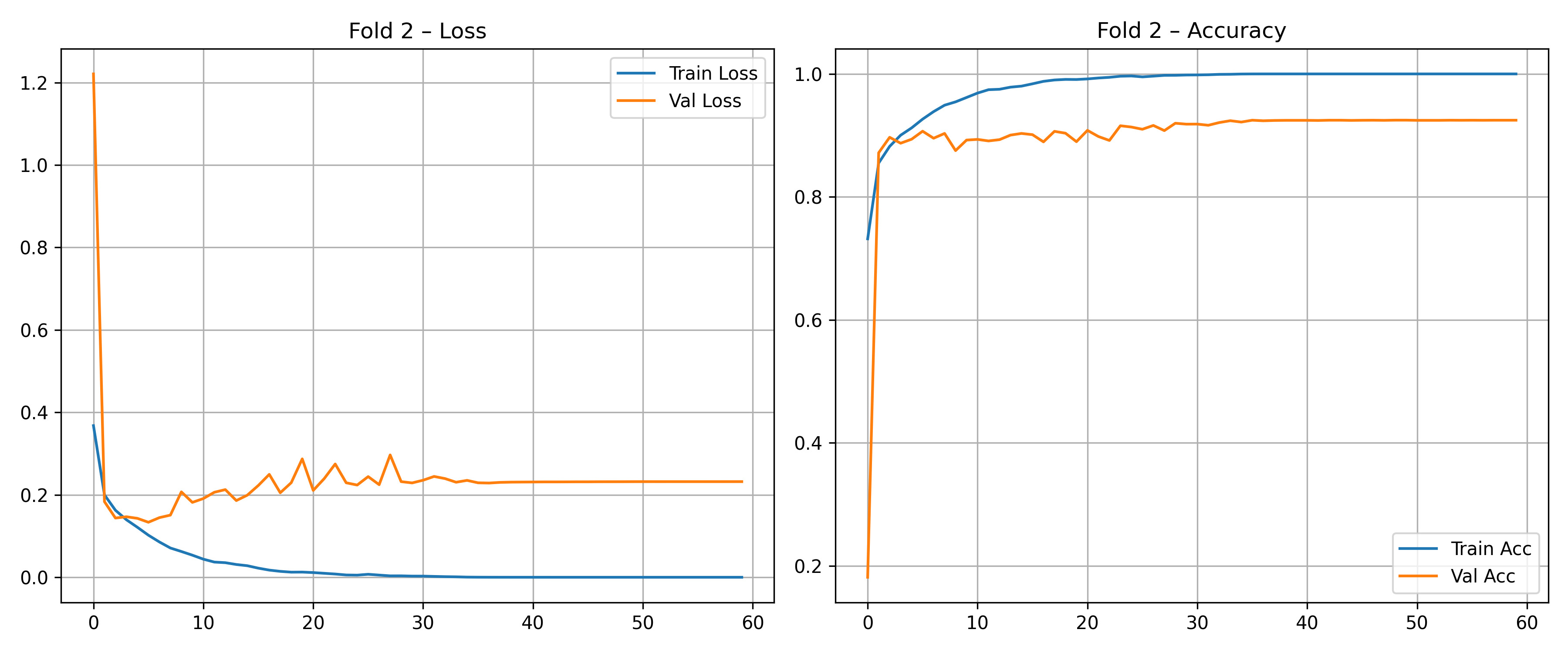}
            \includegraphics[width=0.6\textwidth]{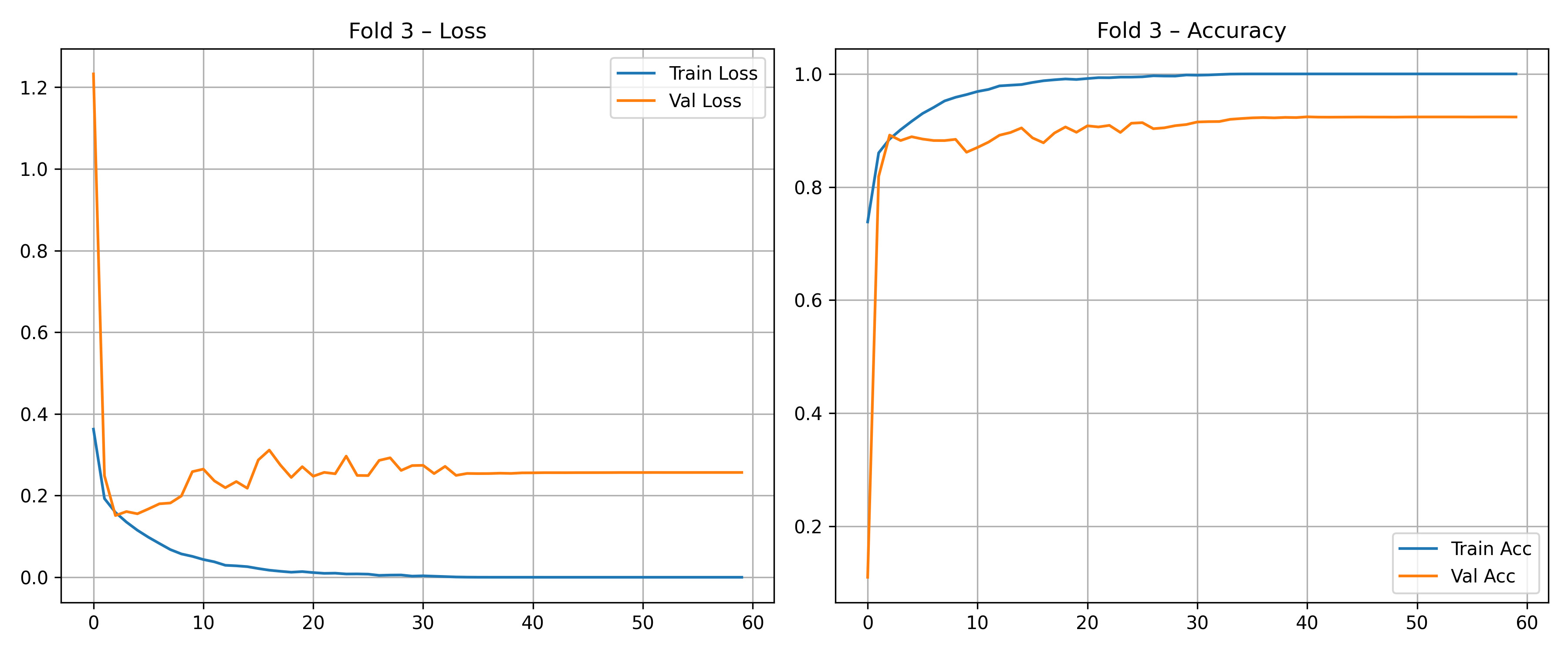}
    \includegraphics[width=0.6\textwidth]{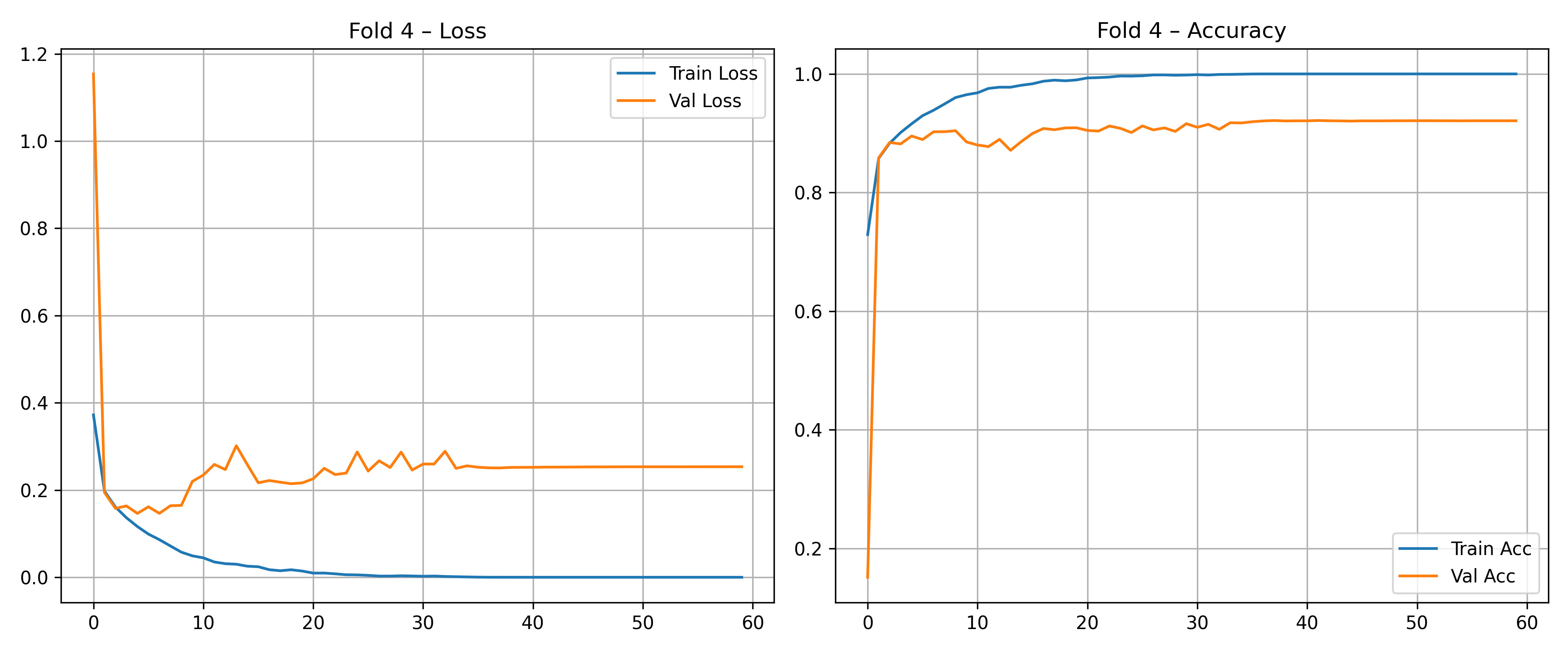}
    \includegraphics[width=0.6\textwidth]{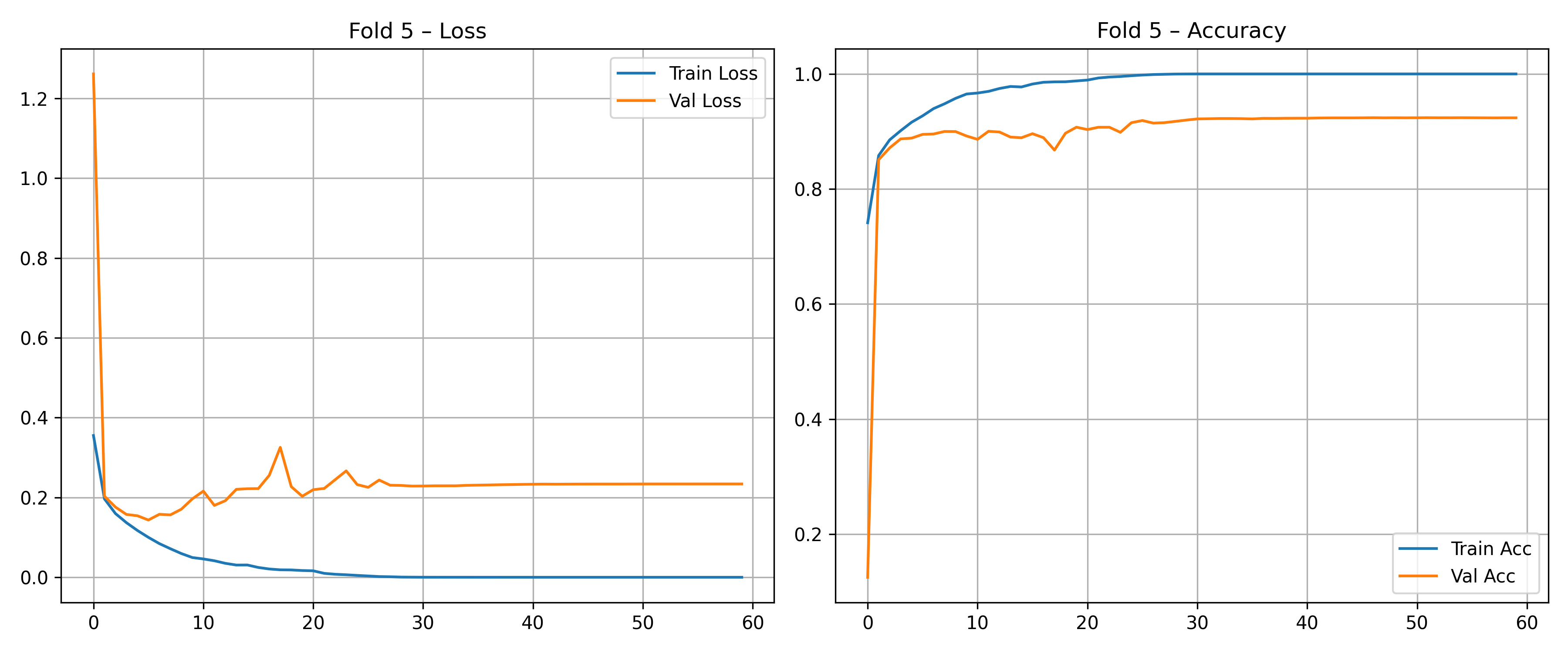}
    \caption{Accuracy Vs. Loss  H-CNN VGG16 Bayesian (Fold 1-5)}
    \label{fig:25}
\end{figure}

%
%

\begin{figure}
    \centering
    \includegraphics[width=0.6\textwidth]{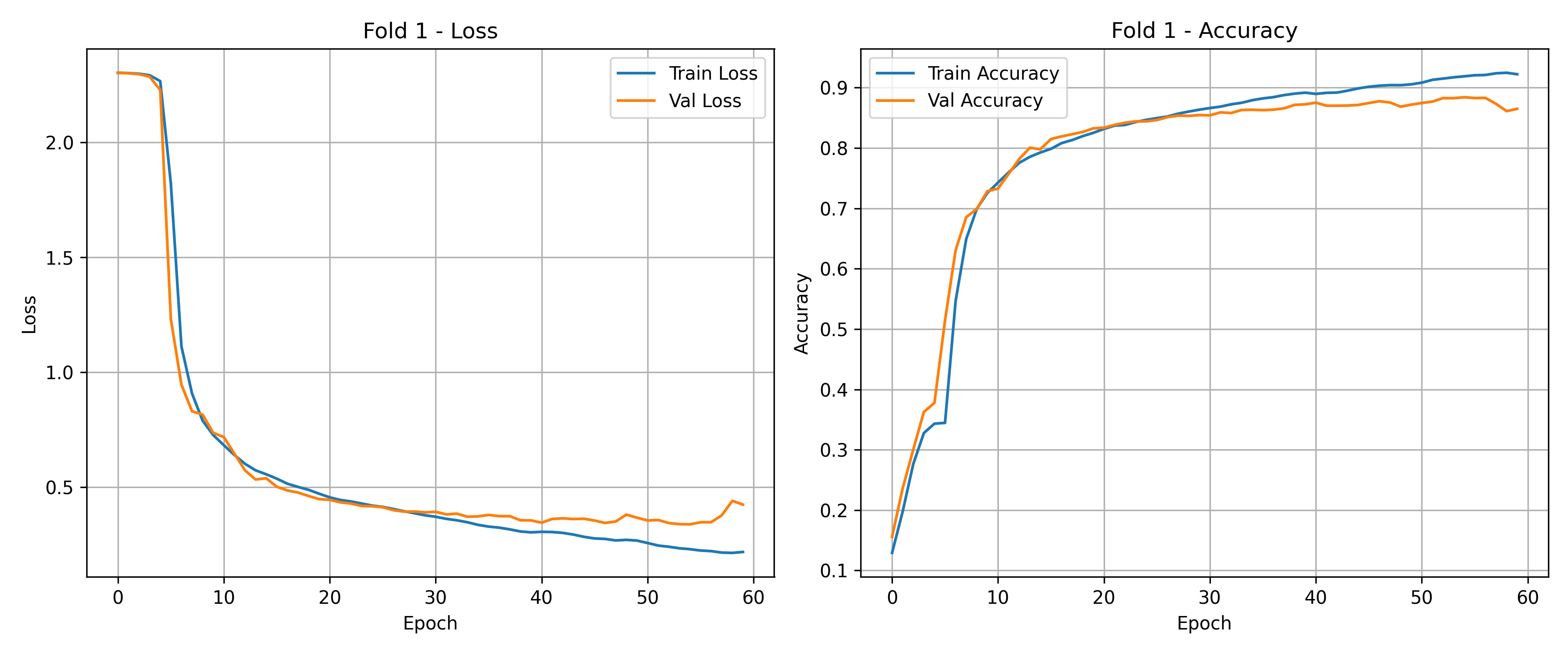}
        \includegraphics[width=0.6\textwidth]{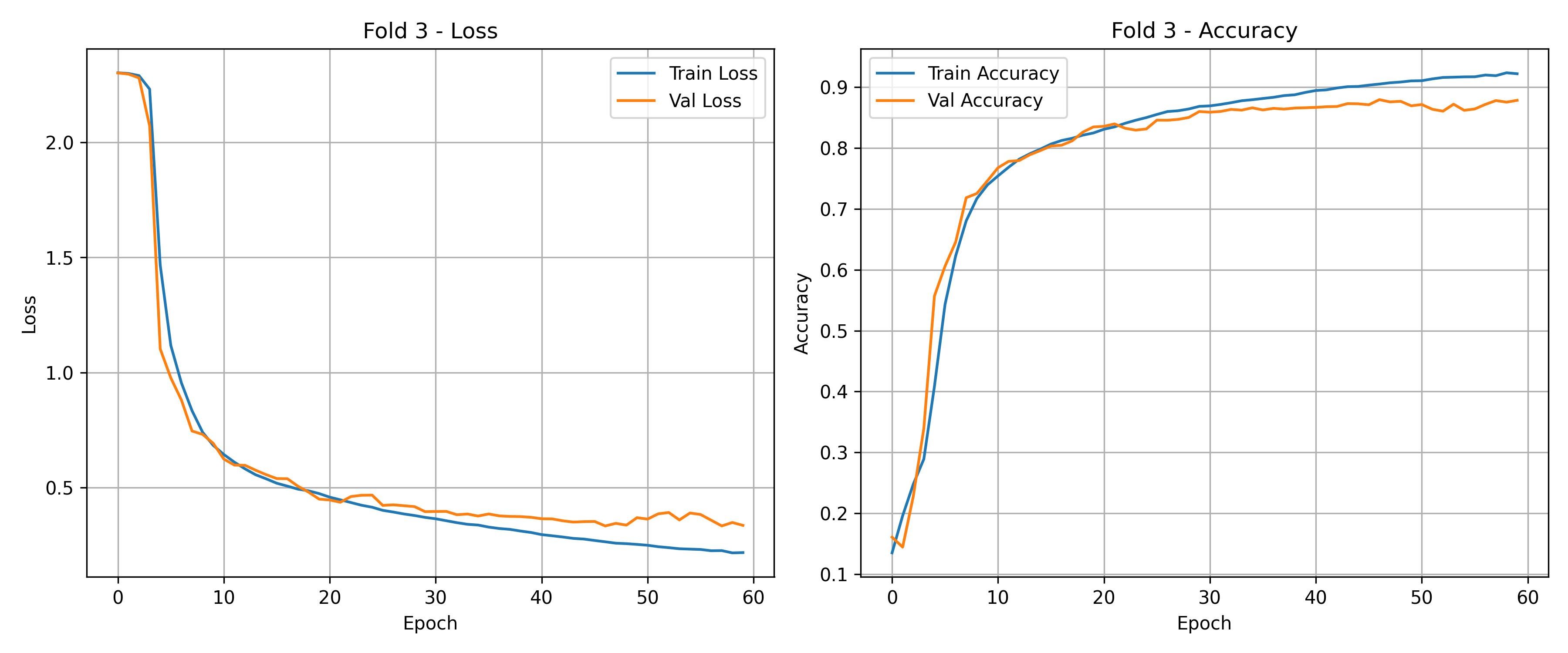}
    \includegraphics[width=0.6\textwidth]{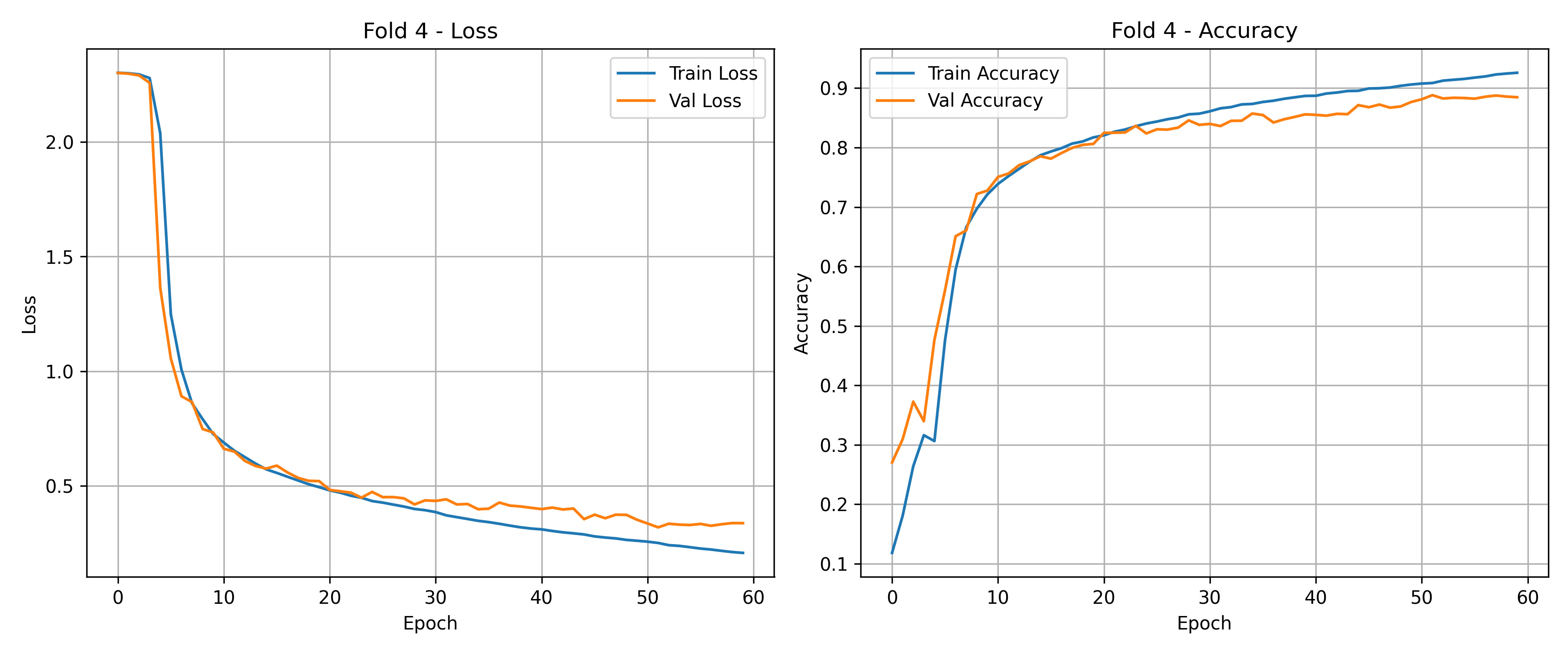}
    \includegraphics[width=0.6\textwidth]{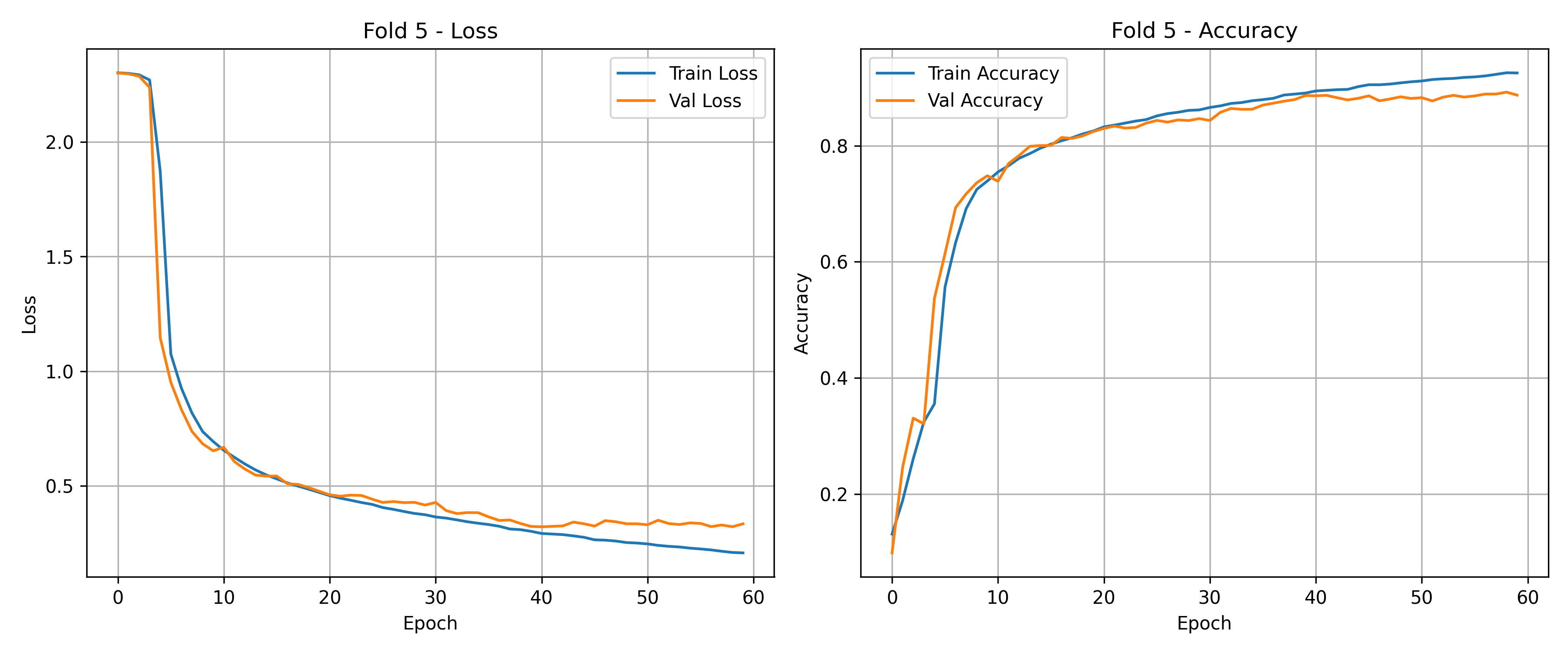}
    \caption{Accuracy Vs. Loss  GoogLeNet (Fold 1, 3-5)}
    \label{fig:30}
\end{figure}

%
%

\begin{figure}
    \centering
    \includegraphics[width=0.6\textwidth]{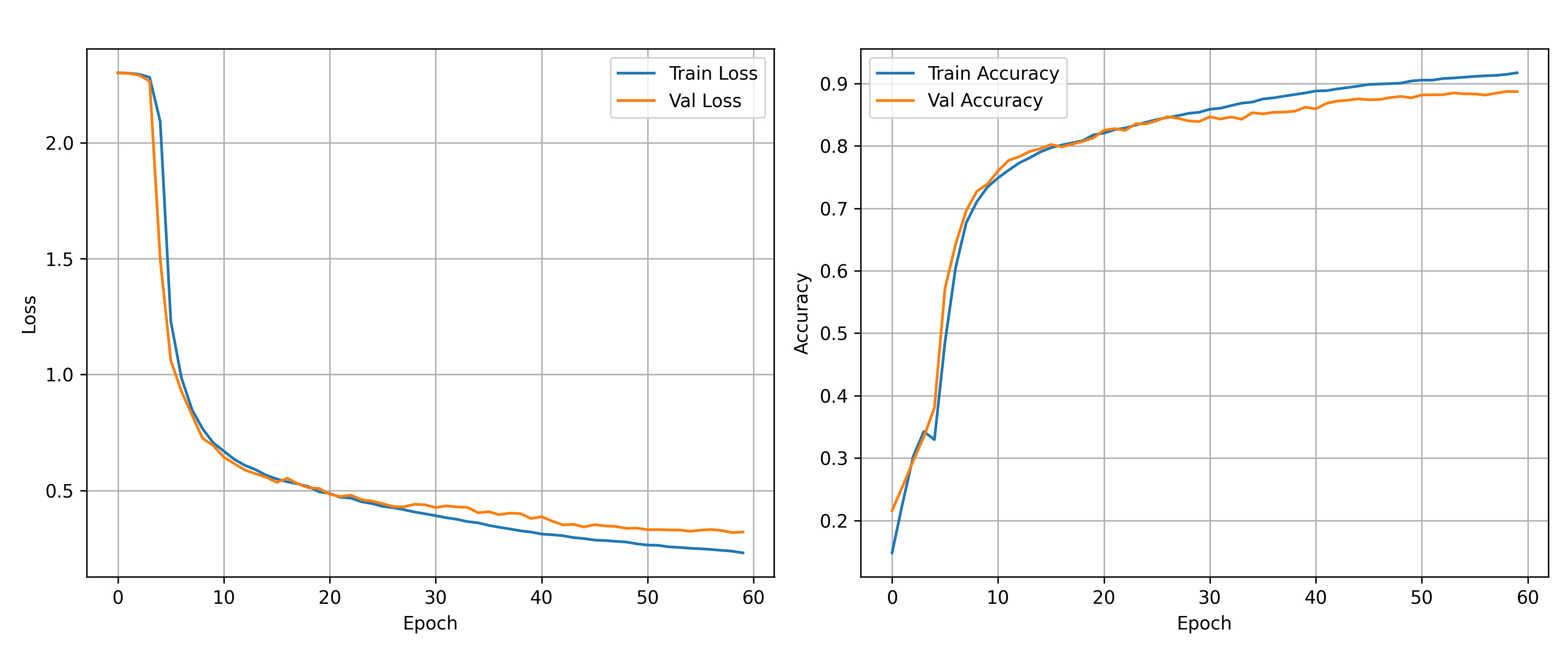}
        \includegraphics[width=0.6\textwidth]{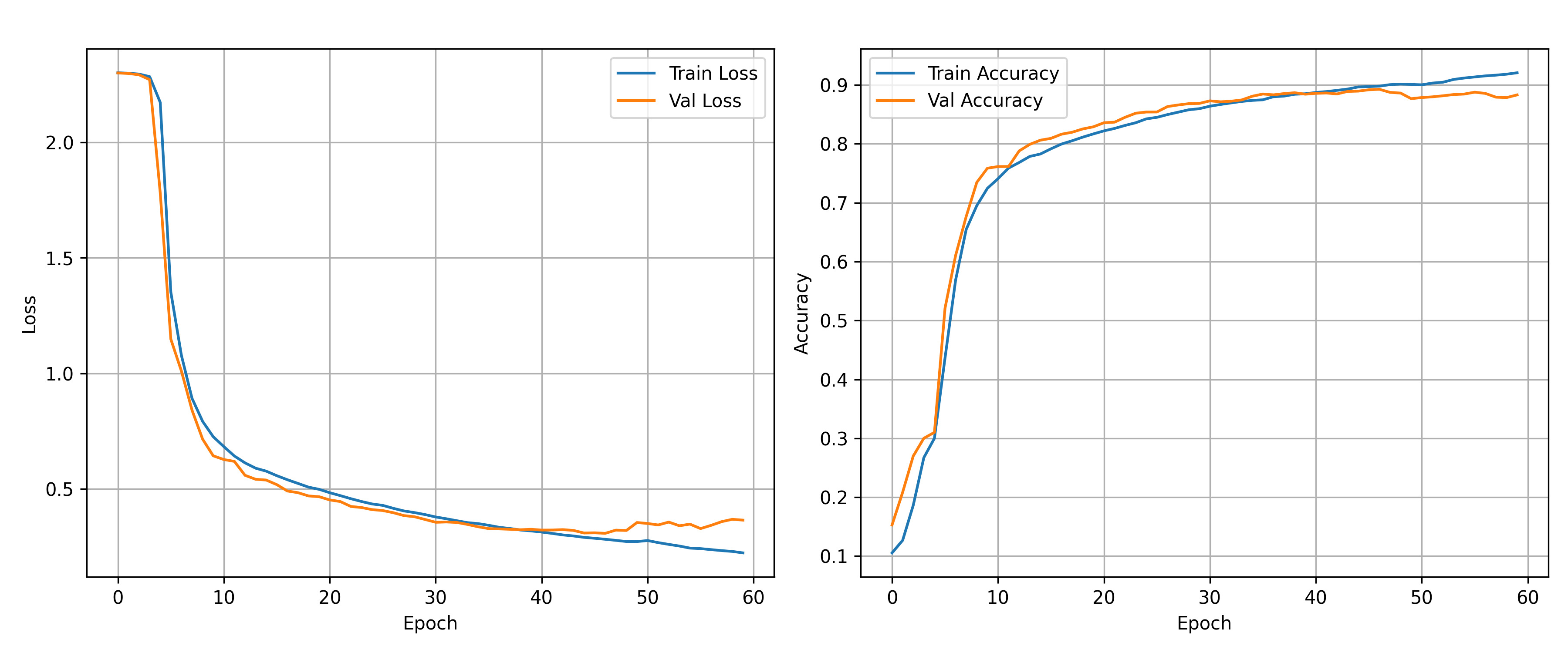}
    \includegraphics[width=0.6\textwidth]{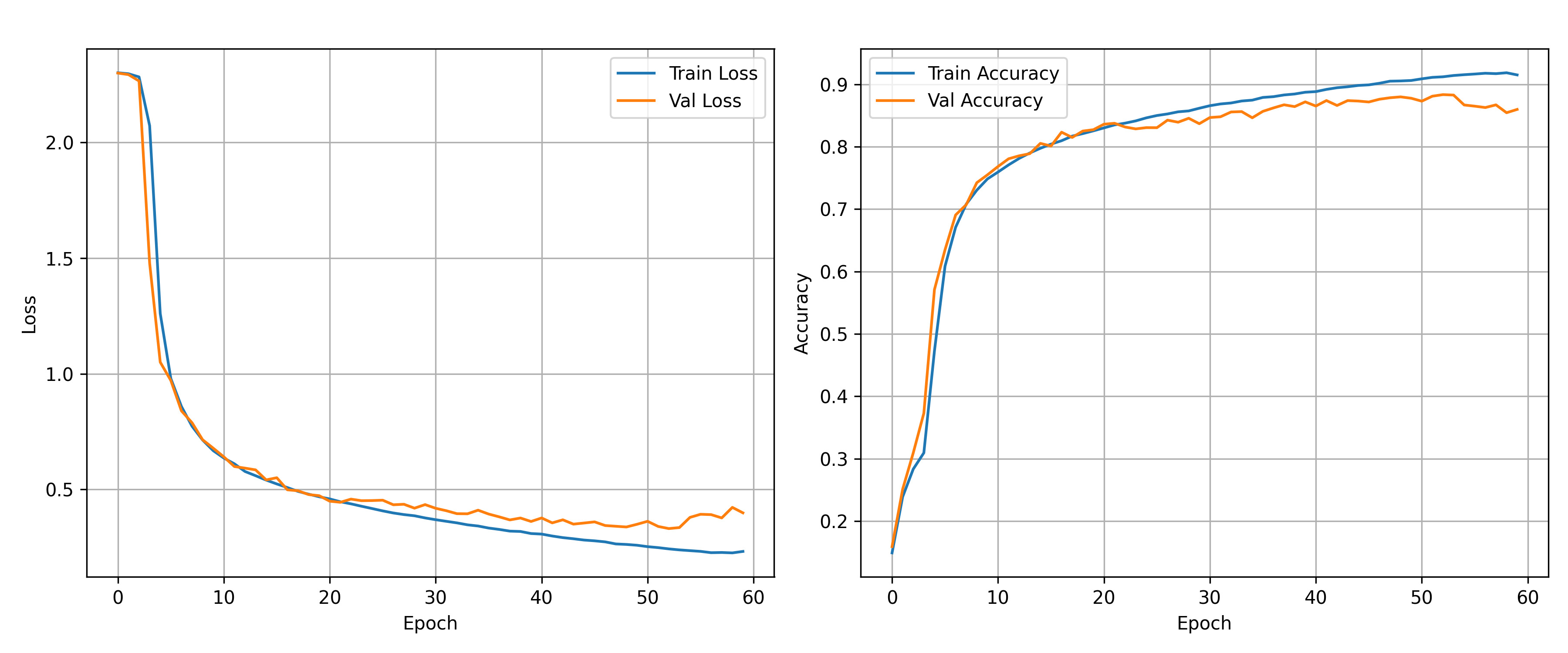}
    \includegraphics[width=0.6\textwidth]{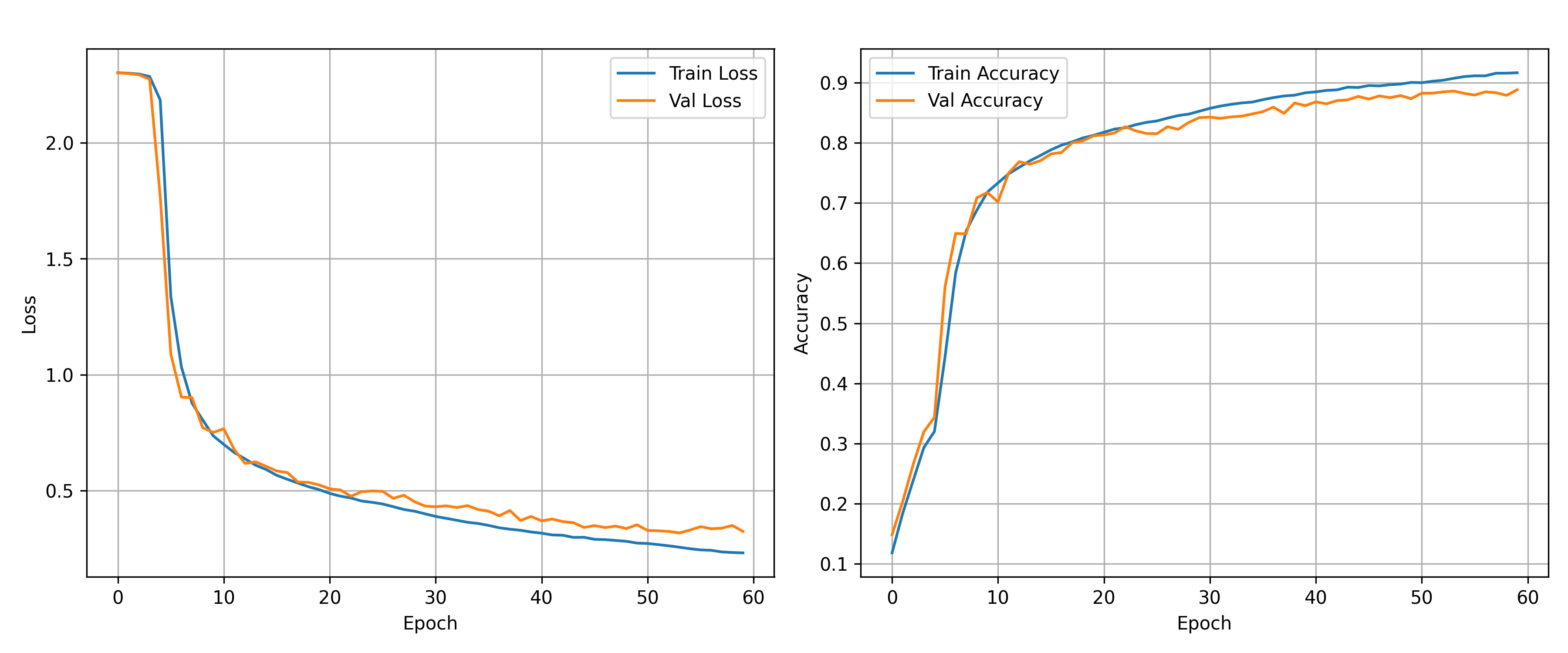}
    \includegraphics[width=0.6\textwidth]{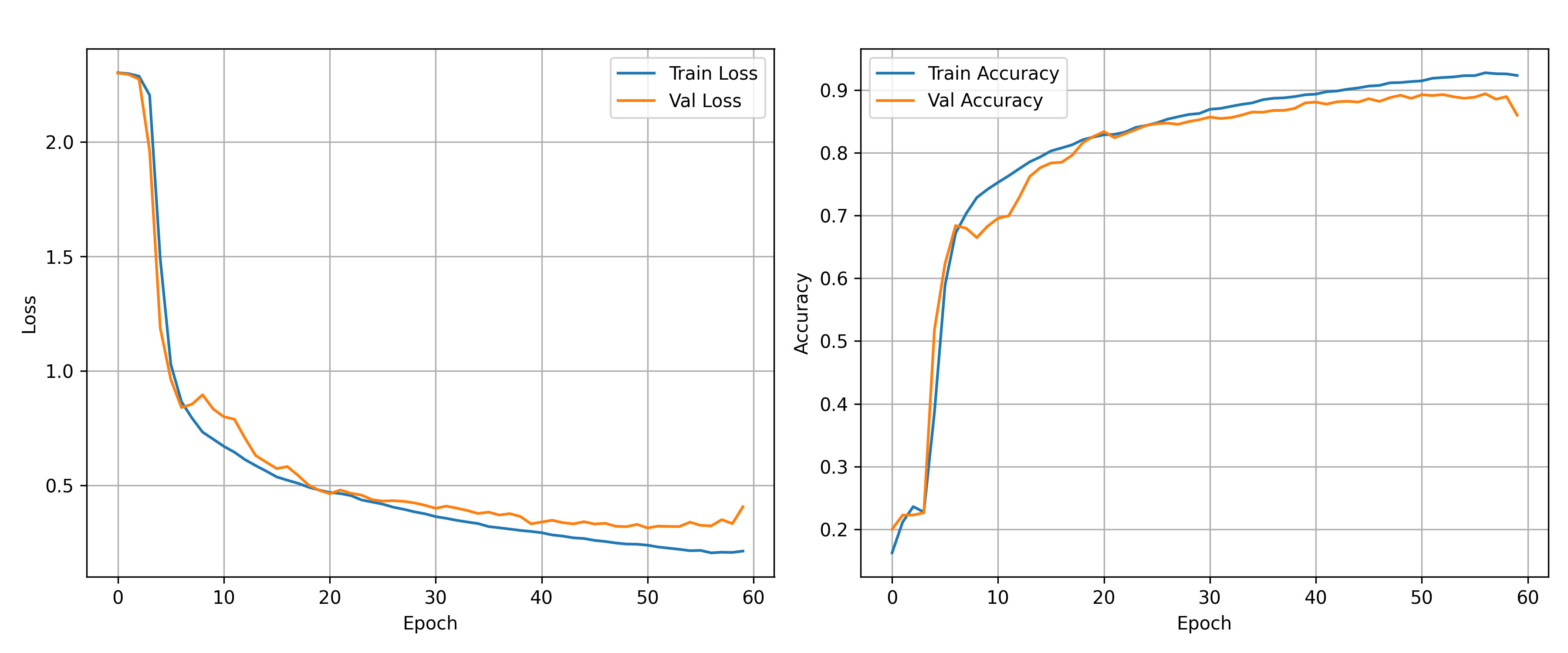}
    \caption{Accuracy Vs. Loss  GoogLeNet Bayesian(Fold 1-5)}
    \label{fig:34}
\end{figure}

%
%
%
%
\begin{figure}[tbh!]
    \centering
    \includegraphics[width=0.45\textwidth]{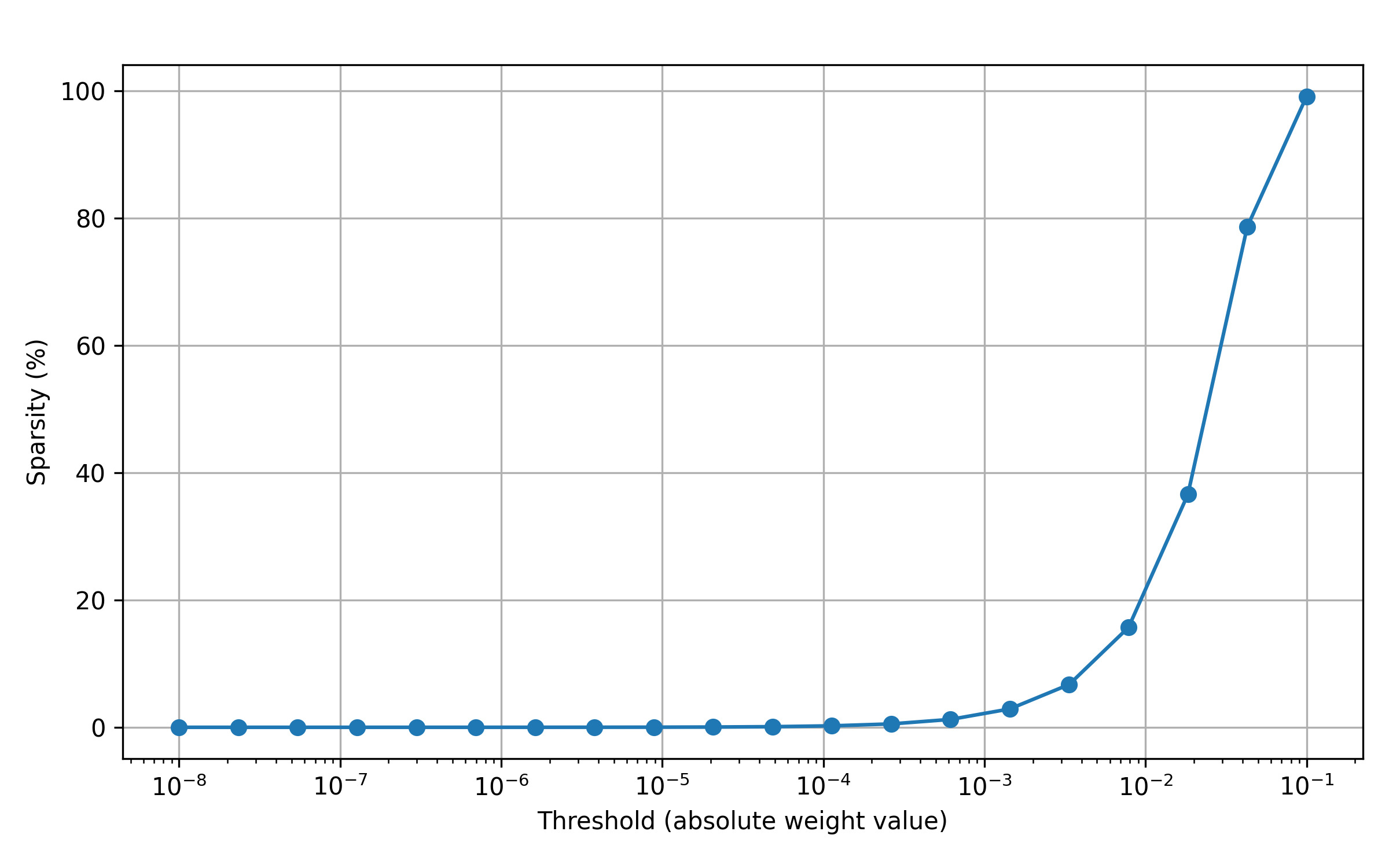}
        \includegraphics[width=0.45\textwidth]{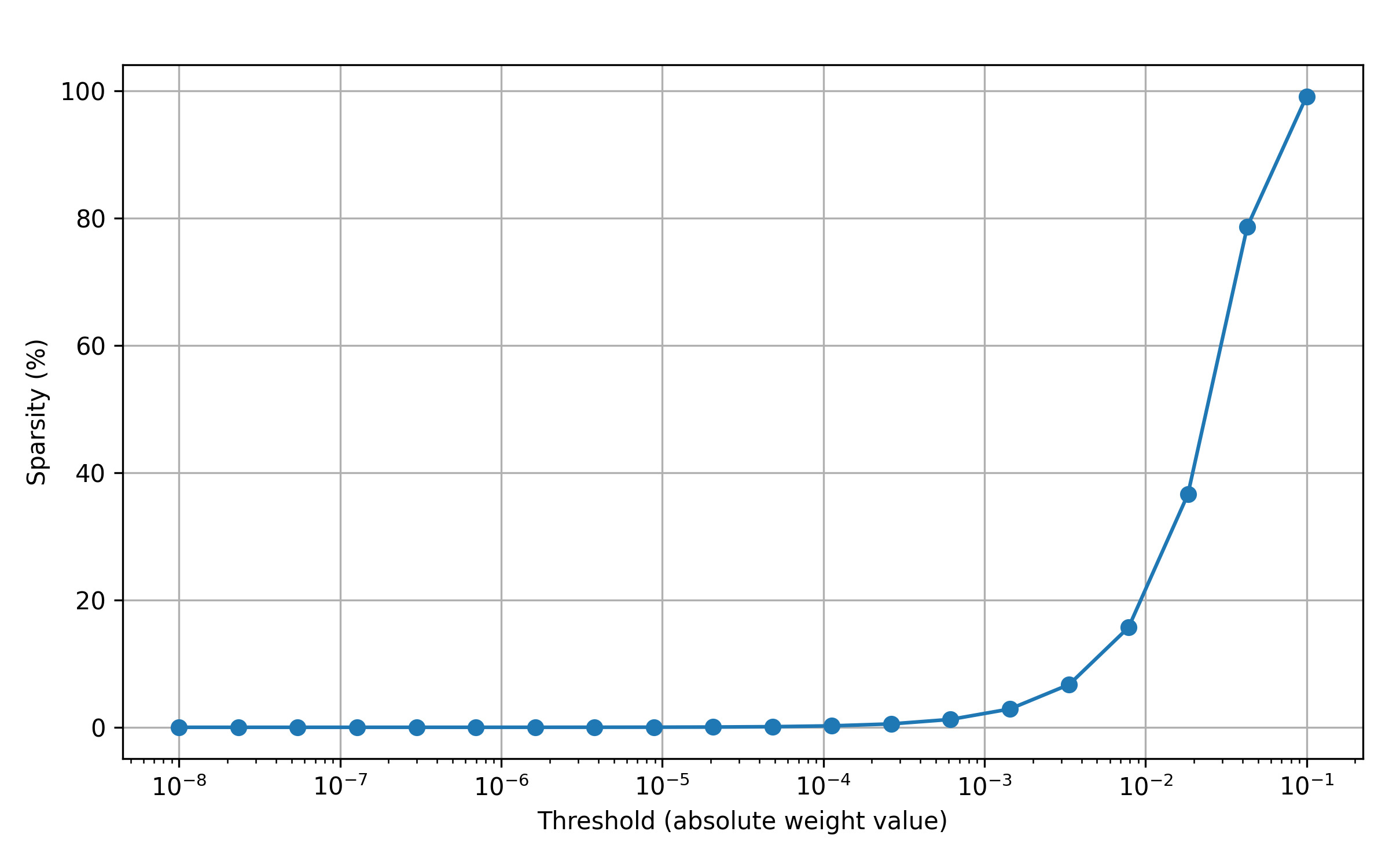}
    \caption{Sparsity vs Threshold VGG16 \& GoogLeNetBayesian \textit{Note.} Sparsity as a percentage of total weights}
    \label{fig:39}
\end{figure}
\begin{figure}[tbh!]
    \centering
    \includegraphics[width=0.45\textwidth]{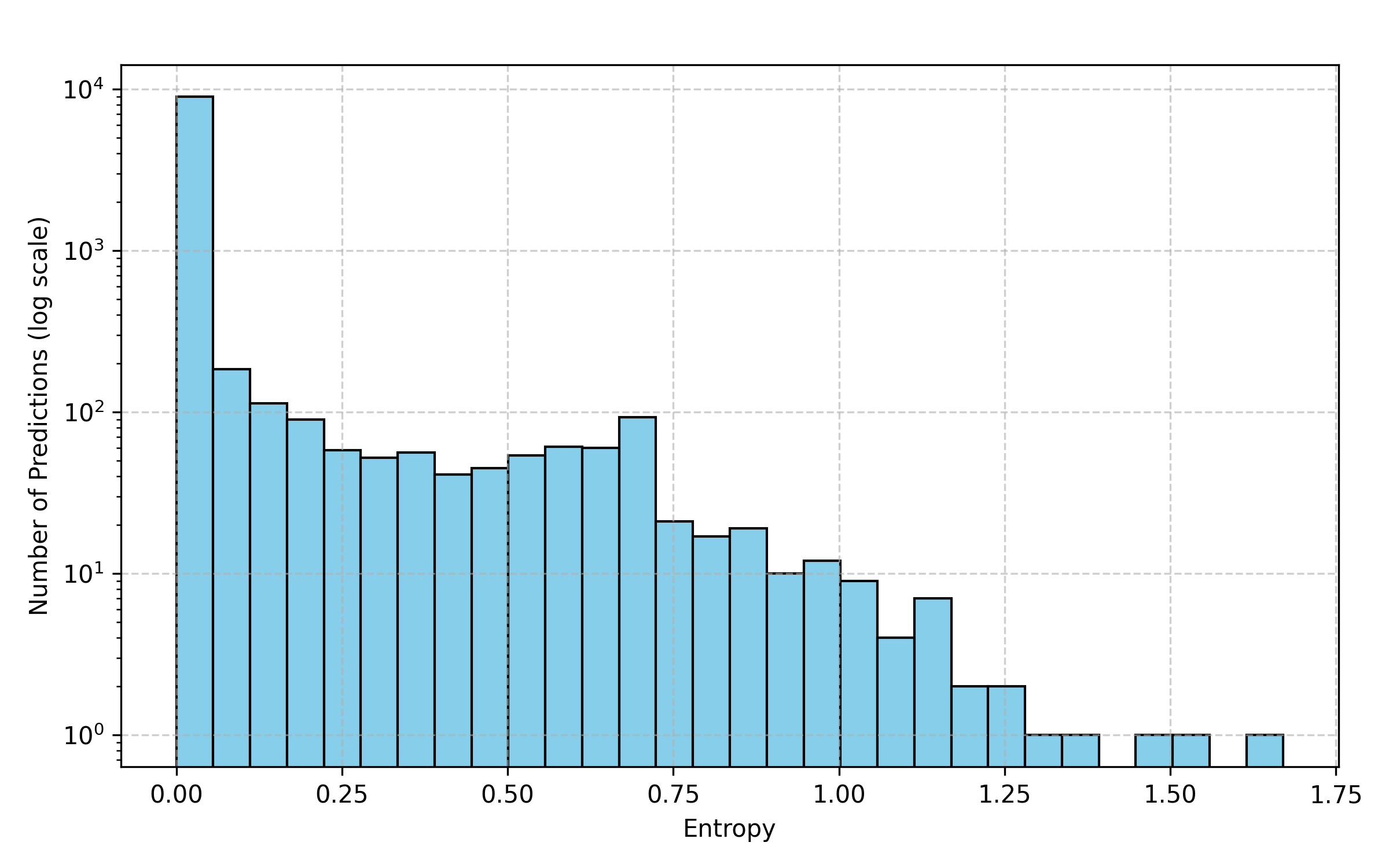}
        \includegraphics[width=0.45\textwidth]{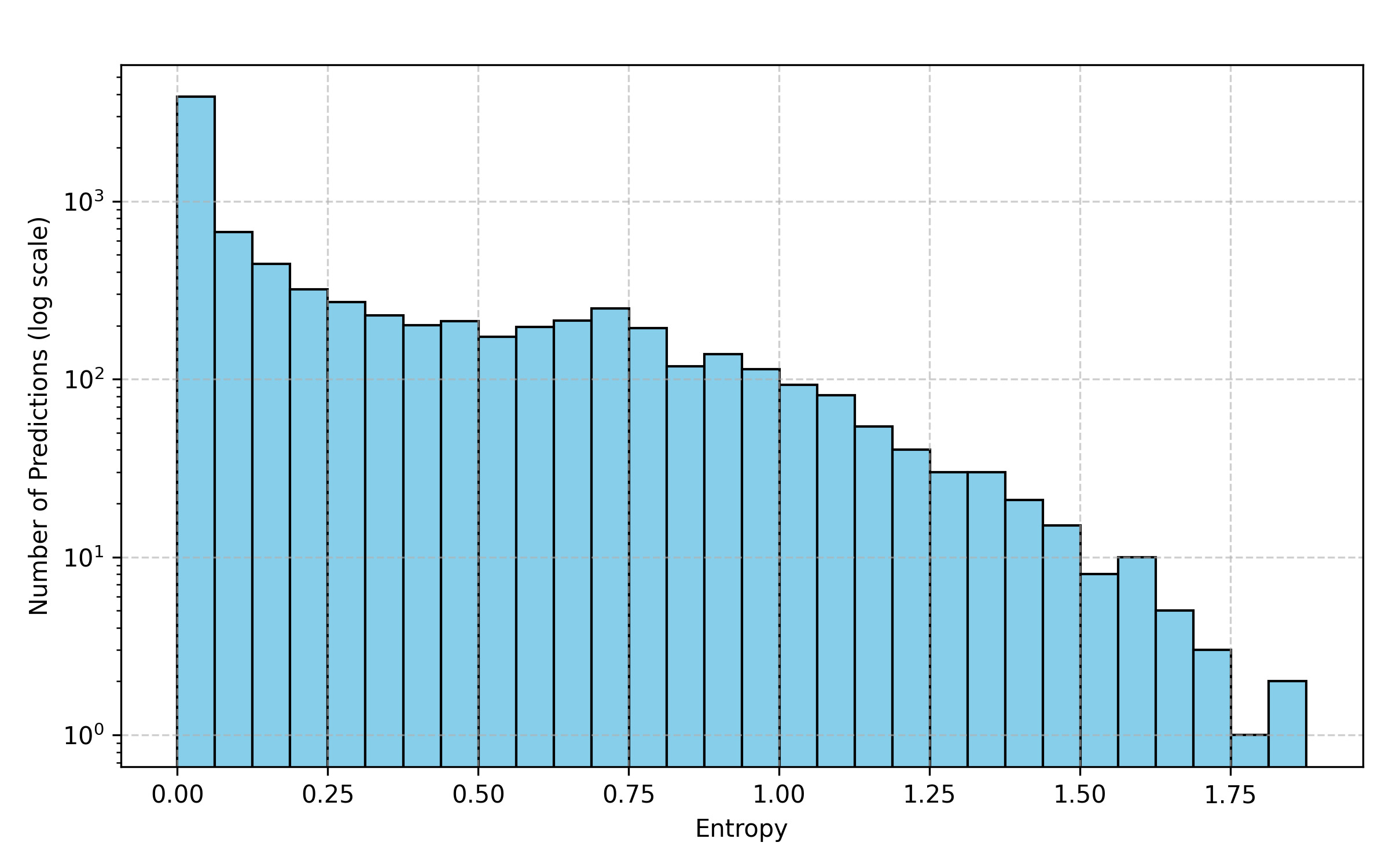}
    \caption{Predictive Entropy for H-CNN VGG16 \& GoogLeNet}
    \label{fig:41}
\end{figure}


\begin{figure}[tbh!]
    \centering
    \includegraphics[width=0.45\textwidth]{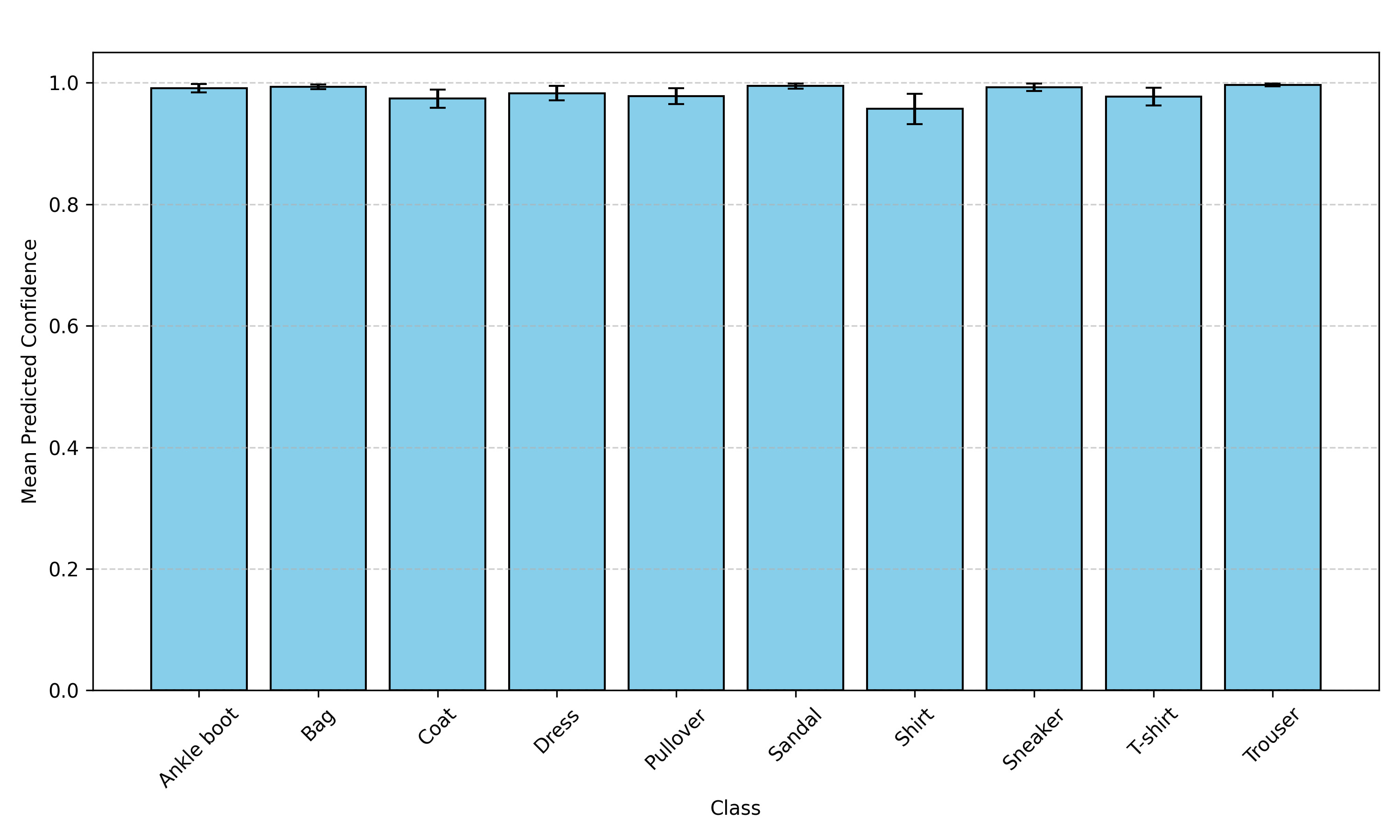}
      \includegraphics[width=0.45\textwidth]{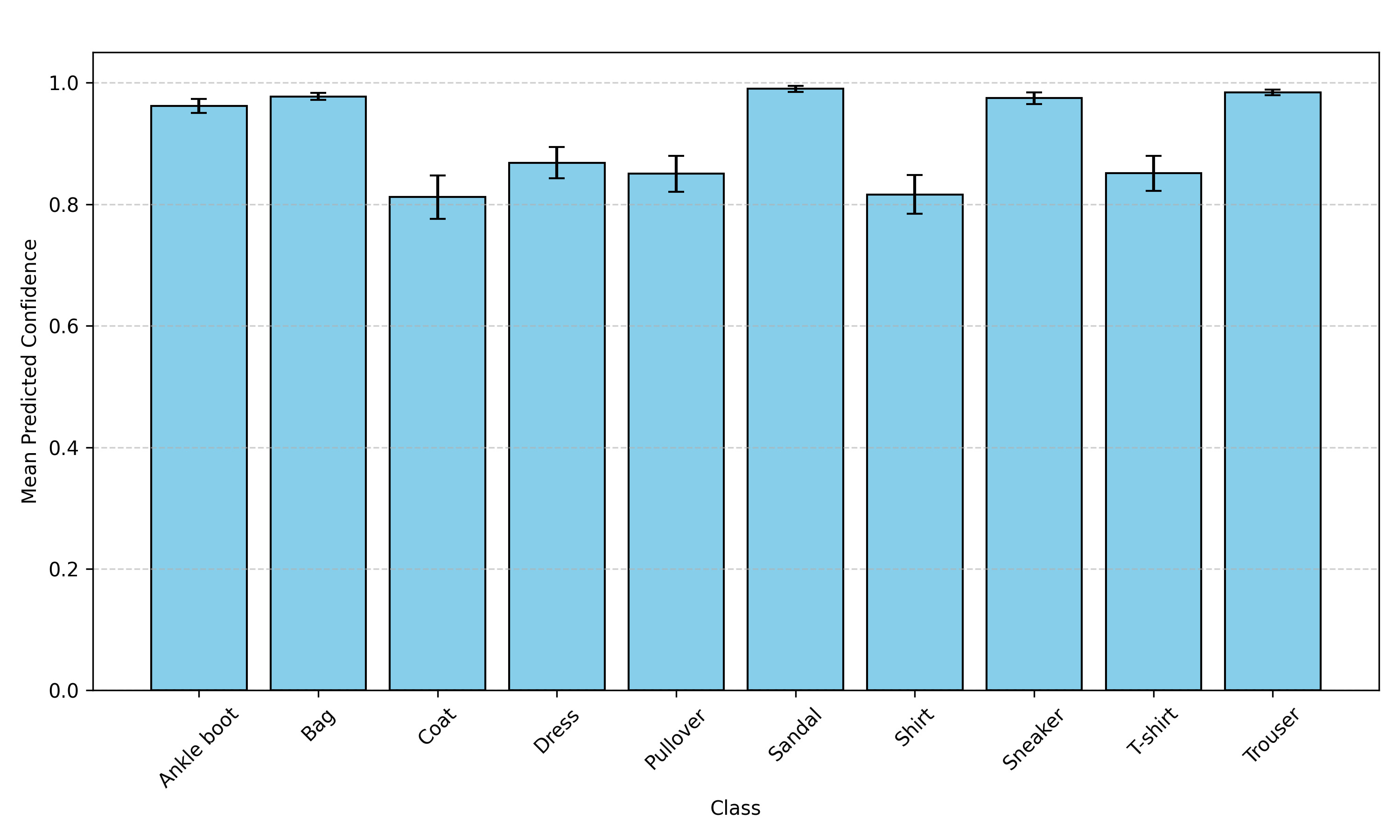}
    \caption{Classwise Confidence Interval for H-CNN VGG16 \& GoogLeNet}
    \label{fig:43}
\end{figure}


\begin{figure}[tbh!]
    \centering
    \includegraphics[width=0.8\linewidth]{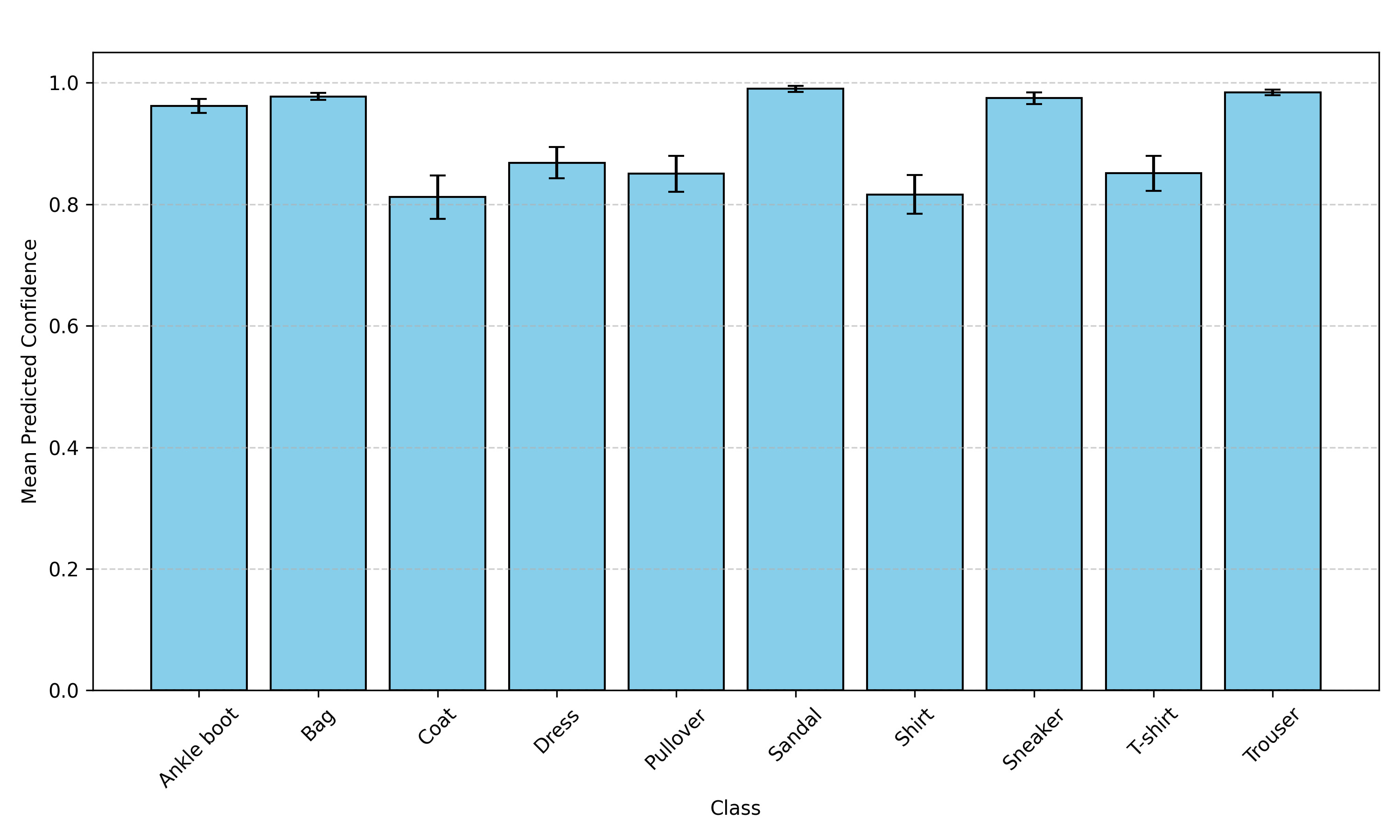}
    \caption{Classwise Confidence Interval GoogLeNet}
    \label{fig:44}
\end{figure}

\begin{table}[h]
\centering
\caption{Sparsity Comparison Across Thresholds for Baseline H-CNN VGG16 and GoogLeNet}
\renewcommand{\arraystretch}{1.15}
\setlength{\tabcolsep}{2pt}
\scriptsize
\begin{tabular}{
    l 
    >{\raggedleft\arraybackslash}p{1.7cm} 
    >{\raggedleft\arraybackslash}p{1.7cm} 
    >{\raggedleft\arraybackslash}p{1.5cm} 
    >{\raggedleft\arraybackslash}p{1.5cm} 
    >{\raggedleft\arraybackslash}p{1.7cm} 
    >{\raggedleft\arraybackslash}p{1.7cm}}
\toprule
\textbf{Weight Range} & 
\multicolumn{2}{c}{\textbf{Number of Weights}} & 
\multicolumn{2}{c}{\textbf{Percentage of Total (\%)}} & 
\multicolumn{2}{c}{\textbf{Cumulative (\%)}} \\
\cmidrule(lr){2-3} \cmidrule(lr){4-5} \cmidrule(lr){6-7}
 & VGG16 & GoogLeNet & VGG16 & GoogLeNet & VGG16 & GoogLeNet \\
\midrule
$<$ 0.00001             & 33,419      & 1,546       & 0.04   & 0.03   & 0.04   & 0.03   \\
0.00001--0.00005        & 133,327     & 5,819       & 0.15   & 0.08   & 0.18   & 0.11   \\
0.00005--0.0001         & 164,021     & 6,024       & 0.18   & 0.10   & 0.37   & 0.21   \\
0.0001--0.0005          & 1,299,677   & 47,958      & 1.44   & 0.80   & 1.81   & 1.01   \\
0.0005--0.001           & 1,610,641   & 59.815      & 1.78   & 1.00   & 3.59   & 2.01   \\
0.001--0.005            & 12,855,606  & 477,097     & 14.23  & 7.98   & 17.82  & 9.99  \\
$\geq$ 0.005            & 74,218,529  & 5,380,271   & 82.18  & 90.01  & 100.00 & 100.00 \\
\bottomrule
\end{tabular}

\label{tab:sparsity_table}
\end{table}


\end{document}